\documentclass[lettersize,journal]{IEEEtran}
\usepackage{amsmath,amsfonts}
\usepackage{algorithmic}
\usepackage{algorithm}
\usepackage{array}
\usepackage{subfigure}
\usepackage{textcomp}
\usepackage{stfloats}
\usepackage{url}
\usepackage{verbatim}
\usepackage{graphicx}
\usepackage{cite}
\usepackage{graphicx}
\usepackage{multirow}
\usepackage{booktabs}
\usepackage{amsfonts}
\usepackage{amsmath}
\usepackage{tikz}
\usepackage[edges]{forest}
\usepackage{color,xcolor}
\usepackage{hyperref}
\usepackage[utf8]{inputenc}
\usepackage{makecell}

\usepackage{standalone}
\usepackage{hyperref}
\usepackage{xr}
\usepackage{tikz}
\usetikzlibrary{shapes.geometric, arrows}

\usepackage[square,sort,comma,numbers]{natbib}

\begin{document}

\title{How to Bridge the Gap between Modalities: Survey on Multimodal Large Language Model}



\author{
    \IEEEauthorblockN{Shezheng Song$^\dagger$, Xiaopeng Li$^\dagger$, Shasha Li$^*$, Shan Zhao$^*$, Jie Yu} \\
    \IEEEauthorblockN{Jun Ma, Xiaoguang Mao, Weimin Zhang, Meng Wang} \\
    \thanks{This work was partly supported by the Hunan Provincial Natural Science Foundation Project (No.2022JJ30668, 2022JJ30046); partly by the science and technology innovation program of Hunan province under grant No. 2021GK2001. This work was supported by the National Natural Science Foundation of China (No.72188101, No.62302144). (Corresponding author: Shasha Li, Shan Zhao)}
    \thanks{Shezheng Song, Xiaopeng Li, Shasha Li, Jie Yu, Jun Ma, Xiaoguang Mao, Weiming Zhang are with the College of Computer, National University of Defense Technology, Changsha 410073, China (e-mail: \{ssz614, shashali, xiaopengli, yj, majun, xgmao\}@nudt.edu.cn, wmzhang104@139.com).}
    \thanks{Shan Zhao, Meng Wang is with the School of Computer and Information Engineering, Hefei University of Technology, Hefei 230009, China (e-mail: zhaoshan@hfut.edu.cn; eric.mengwang@gmail.com).}
    \thanks{$^\dagger$ These authors contributed equally to this work}
    \thanks{$^*$ corresponding author}
}

\markboth{IEEE TRANSACTIONS ON KNOWLEDGE AND DATA ENGINEERING, VOL. XX, NO. XX, DECEMBER 2023}%
{Shell \MakeLowercase{\textit{et al.}}: A Sample Article Using IEEEtran.cls for IEEE Journals}


\maketitle

\begin{abstract}
We explore Multimodal Large Language Models (MLLMs), which integrate LLMs like GPT-4 to handle multimodal data, including text, images, audio, and more. MLLMs demonstrate capabilities such as generating image captions and answering image-based questions, bridging the gap towards real-world human-computer interactions and hinting at a potential pathway to artificial general intelligence.
However, MLLMs still face challenges in addressing the semantic gap in multimodal data, which may lead to erroneous outputs, posing potential risks to society. Selecting the appropriate modality alignment method is crucial, as improper methods might require more parameters without significant performance improvements. This paper aims to explore modality alignment methods for LLMs and their current capabilities. Implementing effective modality alignment can help LLMs address environmental issues and enhance accessibility.
The study surveys existing modality alignment methods for MLLMs, categorizing them into four groups: (1) Multimodal Converter, which transforms data into a format that LLMs can understand; (2) Multimodal Perceiver, which improves how LLMs percieve different types of data; (3) Tool Learning, which leverages external tools to convert data into a common format, usually text; and (4) Data-Driven Method, which teaches LLMs to understand specific data types within datasets.
\end{abstract}



\section{Introduction}
Large Language Models (LLMs) are typically pretrained on vast amounts of data and consist of an enormous number of parameters. These models not only demonstrate significantly enhanced performance across a wide range of tasks but also exhibit emerging capabilities that smaller models lack.
LLMs \cite{LLMSurvey} have garnered substantial attention within the AI community due to their remarkable ability to comprehend, reason with, and generate human language. These models \cite{Vicuna, llama, llama_adpater, llama_adpterv2,peng2023instruction, xu2023baize} exhibit strong language comprehension and logical reasoning abilities.
Despite these significant advancements, the aforementioned LLMs still face challenges when it comes to handling multimodal information, such as tasks involving visual understanding.
    In fact, our perception of the world is enriched by multiple senses, such as sight, sound, and touch \cite{DTCA}. With the rise of social media, people now express their thoughts through information from various modalities. In real-world scenarios, the ability to process and understand multimodal data is essential. This capability, which allows models to comprehend information from different modalities, is the core function of Multimodal Large Language Models (MLLMs).
%
%
Researchers are increasingly considering how to build MLLMs to address a wide range of multimodal tasks. Training MLLMs from scratch to match the NLP capabilities of well-established language models, such as Llama2 \cite{llama2}, is challenging. Therefore, investigating methods to train MLLMs starting from a pre-existing LLM, and exploring ways to bridge the gap between modalities, is an area that requires significant attention in research.
    
  The rapid advancement of MLLMs has significantly enhanced their ability to tackle a wide range of multimodal challenges in real-world applications. However, the semantic gap between modalities remains inadequately addressed. When this gap is not properly managed, it can give rise to several issues:
\textbf{(1) Erroneous generation}, including hallucinations, which could pose significant risks and potential harm to society.
\textbf{(2) Improper modality alignment approaches}, which may require more parameters with minimal performance improvement, leading to high computational costs and inefficiencies.
Therefore, selecting an appropriate modality alignment approach is crucial. This survey aims to explore the modality alignment methods designed for MLLMs and assess the current state of the field. Implementing effective modality alignment can enable MLLMs to address environmental challenges, improve accessibility, and foster inclusivity in deployment.
 \textbf{It is worth noting that the method for transferring the capabilities of LLMs to multimodal scenarios remains unclear.} While  \citet{zhiyuansurvey} focuses on incorporating multimodal information into LLM fine-tuning techniques, such as instruction learning or chain-of-thought, there has been limited attention paid to investigating the differences between modalities within the data.
On the other hand, considerable efforts have been made to align LLMs with human behavior and values. However, the fundamental question of ``alignment with what?'' remains inadequately addressed. To this end, \citet{yao2023instructions} and \citet{shen2023alignment} propose surveys on the alignment objectives of LLMs. Yet, these efforts primarily focus on aligning models to ensure that their behavior is consistent with human values.
As both LLMs and MLLMs are still in the early stages of development, existing approaches to multimodal alignment vary widely, and the research direction remains exploratory. Therefore, \textbf{there is a need to organize the diverse existing research methods related to multimodal information alignment in MLLMs.}
    
    \definecolor{paired-light-blue}{RGB}{198, 219, 239}
\definecolor{paired-dark-blue}{RGB}{49, 130, 188}
\definecolor{paired-light-orange}{RGB}{251, 208, 162}
\definecolor{paired-dark-orange}{RGB}{230, 85, 12}
\definecolor{paired-light-green}{RGB}{199, 233, 193}
\definecolor{paired-dark-green}{RGB}{49, 163, 83}
\definecolor{paired-light-purple}{RGB}{218, 218, 235}
\definecolor{paired-dark-purple}{RGB}{117, 107, 176}
\definecolor{paired-light-gray}{RGB}{217, 217, 217}
\definecolor{paired-dark-gray}{RGB}{99, 99, 99}
\definecolor{paired-light-pink}{RGB}{222, 158, 214}
\definecolor{paired-dark-pink}{RGB}{123, 65, 115}
\definecolor{paired-light-red}{RGB}{231, 150, 156}
\definecolor{paired-dark-red}{RGB}{131, 60, 56}
\definecolor{paired-light-yellow}{RGB}{231, 204, 149}
\definecolor{paired-dark-yellow}{RGB}{141, 109, 49}
\tikzset{%
    parent/.style =          {align=center,text width=1cm,rounded corners=3pt, line width=0.3mm, fill=gray!10,draw=gray!80},
    child/.style =           {align=center,text width=2.3cm,rounded corners=3pt, fill=blue!10,draw=blue!80,line width=0.3mm},
    grandchild/.style =      {align=center,text width=2cm,rounded corners=3pt},
    greatgrandchild/.style = {align=center,text width=1.5cm,rounded corners=3pt},
    greatgrandchild2/.style = {align=center,text width=1.5cm,rounded corners=3pt},    
    referenceblock/.style =  {align=center,text width=1.5cm,rounded corners=2pt},
    top_class/.style =           {align=center,text width=2cm,rounded corners=3pt, fill=paired-light-gray!50,draw=paired-dark-gray!65,line width=0.3mm},
    generation/.style =           {align=center,text width=2cm,rounded corners=3pt, fill= paired-light-green!50,draw=paired-dark-green!75,line width=0.3mm}, 
    generation_wide/.style =           {align=center,text width=2.5cm,rounded corners=3pt, fill= paired-light-green!50,draw=paired-dark-green!75,line width=0.3mm}, 
    generation_more/.style =           {align=center,text width=4cm,rounded corners=3pt, fill= paired-light-green!50,draw=paired-dark-green!75,line width=0.3mm},   
    generation_work/.style =           {align=center,text width=11.0cm,rounded corners=3pt, fill= paired-light-green!50,draw= cyan!0,line width=0.3mm},
    encoder/.style =           {align=center,text width=2cm,rounded corners=3pt, fill=paired-light-orange!50,draw=paired-dark-orange!65,line width=0.3mm},  
    encoder_more/.style =           {align=center,text width=4cm,rounded corners=3pt, fill=paired-light-orange!50,draw=paired-dark-orange!65,line width=0.3mm}, 
    encoder_work/.style =           {align=center,text width=11.0cm,rounded corners=3pt, fill=paired-light-orange!50,draw=red!0,line width=0.3mm},    
    gpa/.style =           {align=center,text width=2cm,rounded corners=3pt, fill=paired-light-blue!50,draw=paired-dark-blue!65,line width=0.3mm},
    gpa_wide/.style =           {align=center,text width=3cm,rounded corners=3pt, fill=paired-light-blue!50,draw=paired-dark-blue!65,line width=0.3mm},   
    gpa_work/.style =           {align=center, text width=10.0cm,rounded corners=3pt, fill=paired-light-blue!50,draw=blue!0,line width=0.3mm},
    data/.style =           {align=center,text width=2cm,rounded corners=3pt, fill=paired-light-blue!50,draw=paired-dark-blue!65,line width=0.3mm},
    data_wide/.style =           {align=center,text width=3cm,rounded corners=3pt, fill=paired-light-blue!50,draw=paired-dark-blue!65,line width=0.3mm},   
    data_work/.style =           {align=center, text width=4.5cm,rounded corners=3pt, fill=paired-light-blue!50,draw=blue!0,line width=0.3mm},  
    model/.style =           {align=center,text width=2cm,rounded corners=3pt, fill=paired-light-orange!50,draw=paired-dark-orange!65,line width=0.3mm},  
    model_more/.style =           {align=center,text width=4cm,rounded corners=3pt, fill=paired-light-orange!50,draw=paired-dark-orange!65,line width=0.3mm}, 
    model_work/.style =           {align=center,text width=4.5cm,rounded corners=3pt, fill=paired-light-orange!50,draw=red!0,line width=0.3mm},    
    pretraining/.style =           {align=center,text width=2cm,rounded corners=3pt, fill= paired-light-green!50,draw=paired-dark-green!75,line width=0.3mm}, 
    pretraining_wide/.style =           {align=center,text width=2.5cm,rounded corners=3pt, fill= paired-light-green!50,draw=paired-dark-green!75,line width=0.3mm}, 
    pretraining_more/.style =           {align=center,text width=4cm,rounded corners=3pt, fill= paired-light-green!50,draw=paired-dark-green!75,line width=0.3mm},   
    pretraining_work/.style =           {align=center,text width=4.5cm,rounded corners=3pt, fill= paired-light-green!50,draw= cyan!0,line width=0.3mm},      
    finetuning/.style =           {align=center,text width=2cm,rounded corners=3pt, fill= paired-light-purple!50,draw=paired-dark-purple!75,line width=0.3mm},   
    finetuning_work/.style =           {align=center,text width=4.5cm,rounded corners=3pt, fill= paired-light-purple!50,draw= orange!0,line width=0.3mm},        
    inference/.style =           {align=center,text width=2cm,rounded corners=3pt, fill= paired-light-red!35,draw=paired-light-red!90,line width=0.3mm},           
    inference_more/.style =           {align=center,text width=2cm,rounded corners=3pt, fill=paired-light-red!35,draw=paired-dark-orange!65,line width=0.3mm},
    inference_work/.style =           {align=center,text width=11.0cm,rounded corners=3pt, fill=paired-light-red!35,draw=red!0,line width=0.3mm},         
}
\begin{figure*}
\centering
\scriptsize
\hspace*{-30pt}
    \begin{forest}
    for tree={
            forked edges,
            grow'=0,
            draw,
            rounded corners,
            node options={align=center,},
            text width=2.7cm,
            s sep=6pt,
            calign=edge midpoint,
        },
    [MLLM, fill=gray!45, parent
        [\textbf{Multimodal Convertor (\S \ref{sec:Multimodal Converter})},encoder
            [Feature Projector (\S \ref{subsec:Feature projector}) ,fill=red!45,encoder
                [
                Llava \cite{llava}; ML-MFSL \cite{ML_MFSL}; FROMAGe \cite{FROMAGe}; VideoLLM \cite{VideoLLM}; Videochat \cite{Videochat}; DreamLLM \cite{DreamLLM}; CogVLM \cite{CogVLM}; PointLLM \cite{pointllm}; METALM \cite{hao2022metalm}; KOSMOS1 \cite{huang2024kosmos1}; KOSMOS2 \cite{peng2023kosmos2}; MiniGPT-v2 \cite{chen2023minigptv2}; CogAgent \cite{cogagent}; NExT-Chat \cite{zhang2023NExT-Chat}; DetGPT \cite{pi2023detgpt}; LayoutGPT \cite{feng2024layoutgpt}; RT-2 \cite{brohan2023rt2}; PointCLIPV2 \cite{zhu2023pointclip}; CaFo \cite{zhang2023cafo}; StableLlava \cite{li2023stablellava}; VisCPM \cite{VisCPM}; MoAI \cite{lee2024moai}; 
                LLaVA-Interactive \cite{LLaVA-Interactive}; Git \cite{Git}; BLIVA \cite{BLIVA}; NExT-GPT \cite{NExT-GPT}; Lynx \cite{Lynx}; Fuyu \cite{fuyu-8b}; LAMM \cite{lamm}; GLaMM \cite{rasheed2024glamm}, encoder_work
                ]
            ]
            [Scaling Up (\S \ref{subsec:Scaling up}),fill=red!45,encoder
                [
                    VILA \cite{lin2024vila}; Qwen-VL~\cite{qwen_vl}; PaLI~\cite{chen2022pali}; PaLI-X \cite{chen2023paliX}; PaLI-3 \cite{chen2023pali3} InternVL \cite{chen2024internvl}; InternVL1.5 \cite{chen2024InternVL15}, encoder_work
                ]
            ]
            [Adapter-based Adjustment (\S \ref{subsec:Adapter-based adjustment}),fill=red!45,encoder
                [
                     Prophet~\cite{prophet}; LLaMa adapter~\cite{llama_adpater}; LLaMa adapter2~\cite{llama_adpterv2}; Voxposer \cite{Voxposer}; Qwen-VL \cite{qwen_vl}; LION \cite{chen2024lion}; LISA \cite{lai2024lisa}; LaVIN \cite{luo2023LAVIN}; Lynx \cite{Lynx}; InternLM-XComposer2 \cite{dong2024InterLM-XComposer2}, encoder_work
                ]
            ]
        ]
        [\textbf{Multimodal Perceiver (\S \ref{sec:Multimodal Perceivers})},for tree={ generation} 
            [AE Perceiver (\S \ref{subsec:VAE Perceiver}),fill=red!45,generation
                [
                    LQAE \cite{LQAE}; SPAE \cite{SPAE}; BEiT \cite{BEIT}; UniCode \cite{UniCode},generation_work
                ]
            ]
            [Q-former Perceiver (\S \ref{subsec:Q-former Perceiver}),fill=red!45,generation
                [
                    BLIP-2 \cite{blip2}; Video-Llama \cite{Video-llama}; BLIVA \cite{BLIVA}; VPG-C \cite{VPG-C}; MMICL \cite{MMICL}; Mini-GPT4 \cite{Minigpt-4}; Mini-GPT5 \cite{MiniGPT-5}; InterLM-XComposer\cite{InternLM-XComposer}; Sparkles~\cite{Sparkles}; GAVIE \cite{GAVIE}; InstructBlip \cite{dai2024instructblip}; LION \cite{chen2024lion}; MA-LMM \cite{he2024mallm} ,generation_work
                ]
            ]
            [Customization Perceiver (\S \ref{subsec:Customization Perceiver}),fill=red!45,generation
                [
                    Flamingo \cite{Flamingo}; OpenFlamingo \cite{OpenFlamingo}; MultiModal-GPT \cite{MultiModal-GPT}; MACAW-LLM \cite{Macaw-LLM}; VisionLLM \cite{visionllm};  PandaGPT \cite{PandaGPT}; Otter \cite{otter}; SEEM \cite{SEEM}; SPHINX \cite{SPHINX}; mPLUG-Owl \cite{mPLUG-Owl}; mPLUG-Owl2 \cite{mPLUG-Owl2}; TimeChat \cite{ren2024timechat}; Mini-Gemini \cite{li2024minigemini}; ASM \cite{wang2023asm}; Vary \cite{wei2023varyscalingvisionvocabulary}; Monkey \cite{li2024monkey}; LLaMA-VID \cite{li2023llamavidimageworth2}; LanguageBind \cite{zhu2023languagebind}; FERRET \cite{you2023ferret}; FERRET-UI \cite{you2024ferret}; SEAL \cite{wu2024v}
                    ,generation_work
                ]
            ]
        ]
        [\textbf{Tool Learning (\S \ref{sec:Tools Assistance})},for tree={ gpa}
            [Natural Language-Assisted (\S \ref{subsec:Natural language assisted}),fill=red!45,gpa_wide
                 [ChatCaptioner~\cite{ChatCaptioner}; MM-REACT~\cite{MM-REACT}; IdealGPT~\cite{IdealGPT}; SMs~\cite{SMs}; AVIS \cite{AVIS}; LLaVA-Plus \cite{LLaVA-Plus}; HuggingGPT~\cite{HuggingGPT}; InternGPT~\cite{InternGPT}; Chameleon~\cite{Chameleon}; CAT~\cite{CAT}; MindAgent \cite{MindAgent}; ControlLLM \cite{liu2023controlllm},gpa_work]
                    ]
            [Code-Assisted (\S \ref{subsec:Code assisted}),fill=red!45,gpa_wide
                    [VISPROG~\cite{VISPROG}; ViperGPT~\cite{ViperGPT}; VoxPoser~\cite{Voxposer},gpa_work
                ]
            ]
            [Combined Code and Natural Language-Assisted (\S \ref{subsec:Both code and natural language assisted}),fill=red!45,gpa_wide
                    [AssistGPT \cite{AssistGPT}; TaskMatrix.AI~\cite{TaskMatrix.AI}; CLOVA \cite{gao2024clova},gpa_work
                ]
            ]
        ]
        [\textbf{Data Driven (\S \ref{sec:Data-driven MLLMs})},for tree={ inference}
            [Enhanced Comprehension (\S \ref{subsec:Enhanced Image Comprehension}),fill=red!45,inference_more
                [
                    ShareGPT4V \cite{ShareGPT4V}; ALLAVA \cite{chen2024allava}; Mini-Gemini \cite{li2024minigemini}; Llava \cite{llava}; SightBeyondText \cite{tu2023sight}; DetGPT \cite{pi2023detgpt}; GAVIE \cite{GAVIE}; Llava1.5 \cite{liu2024Llava1.5}; Osprey \cite{yuan2024osprey}; Otter-HD \cite{OtterHD}; InternLM-XComposer2-4KHD \cite{dong2024InternLM-XComposer2-4KHD}; InternVL1.5 \cite{chen2024InternVL15}; MiniGPT-v2 \cite{chen2023minigptv2}; CogAgent \cite{cogagent}; InterLM-XComposer2 \cite{dong2024InterLM-XComposer2}; LLaVAR \cite{LLaVAR}; VisCPM \cite{VisCPM}; Muffin \cite{muffin}; ASMv2 \cite{wang2024asmv2}
                    ,inference_work 
                ]
            ]
            [Spatial Comprehension (\S \ref{subsec:Spatial Comprehension}),fill=red!45,inference_more
                [
                    Shikra \cite{shikra}; GPT4ROI \cite{GPT4ROI}; KOSMOS2 \cite{peng2023kosmos2}; GLaMM \cite{GLaMM}; ViP-LLaVA \cite{cai2024vipllava}; Lenna \cite{wei2023lenna}; CogVLM \cite{CogVLM}; LION \cite{chen2024lion}; MiniGPT-v2 \cite{chen2023minigptv2}; NExT-Chat \cite{zhang2023NExT-Chat}; InstructDet \cite{dang2023instructdet}; DetGPT \cite{pi2023detgpt}; PVIT \cite{chen2023pvit}; VisionLLM \cite{visionllm}; ASM \cite{wang2023asm}; Osprey \cite{yuan2024osprey}
                    , inference_work
                ]
            ]
            [Complex Modalities (\S \ref{subsec:Complex Modalities}),fill=red!45,inference_more
                [
                   LAMM \cite{lamm}; PointLLM \cite{pointllm}; PointCLIPV2 \cite{zhu2023pointclip}; RemoteSensing ChatGPT \cite{guo2024RemoteSensingChatGPT}; RSGPT \cite{hu2023rsgpt}; H2RSVLM \cite{pang2024h2rsvlm}; LHRS-Bot \cite{muhtar2024LHRS-Bot}; SkyEyeGPT \cite{zhan2024skyeyegpt}; mPLUG-PaperOwl \cite{hu2023mPLUG-PaperOwl}; VELMA \cite{schumann2024velma}; RT-2 \cite{brohan2023rt2}; CogAgent \cite{cogagent}
                   ,inference_work
                ]
            ]
            [Any Modalities (\S \ref{sec:Any Modalities}) ,fill=red!45,inference_more
                [
                    TaskMatrixAI \cite{TaskMatrix.AI}; ImageBind \cite{Imagebind}; PandaGPT \cite{PandaGPT}; OneLLM \cite{OneLLM}; AnyGPT \cite{zhan2024anygpt}; NExT-GPT \cite{NExT-GPT}; MACAW \cite{Macaw-LLM}
                    ,inference_work
                ]  
            ]
            [Domain Specific (\ref{sec:Domain specific}) ,fill=red!45,inference_more
                [
                    Llava-Med \cite{llavamed}; CancerGPT \cite{CancerGPT}; PMC-Llama \cite{wu2024PMC-LLaMA}; BiomedGPT \cite{zhang2023biomedgpt}; Qilin-Med-VL \cite{liu2023Qilin-Med-VL}
                    ,inference_work
                ]  
            ]
        ]
    ]
    \end{forest}
    \caption{The taxonomy of existing approaches for bridging modality gap in Multimodal Large Language Models (MLLMs).}
    \label{fig:Taxonomy}
\end{figure*}
    
    MLLMs refer to LLMs that can perceive and understand multimodal data \cite{zhiyuansurvey}. Exploring how unimodal LLMs can adapt to multimodal data and how vision models can be effectively integrated with unimodal LLMs is of significant value. To this end, we have summarized existing methods for bridging this gap, with the hope that this work can serve as a reference for future researchers in designing and training MLLMs.
Specifically, as shown in Fig. \ref{fig:Taxonomy}, we categorize MLLMs into three structural types and one data type, each addressing modality differences from a distinct perspective.
    \textit{(1) Multimodal Converter}. The converter transforms multimodal information into objects that can be understood or learned by LLMs, leveraging the models' capabilities to learn from these transformed objects. It typically employs a linear projector to align multimodal data with the LLM, ensuring consistent input to both the encoder and the LLM. This non-selective expression facilitates a comprehensive representation of multimodal data, enhancing the LLM ability to acquire complex capabilities, such as referential understanding of images and enhanced image comprehension.
\textit{(2) Multimodal Perceiver}. These methods focus on designing multimodal perceivers to interface with LLMs, primarily aiming to improve the perceptual capabilities of the LLM with respect to multimodal information. The Perceiver model selectively processes multimodal data by transforming visual tokens into features that are compatible with textual tokens. This design enables the model to focus on relevant information, alleviating the challenges of multimodal alignment within the LLM.
\textit{(3) Tool Learning}. These methods utilize external tools to compensate for the limitations of LLMs, which can only understand natural language. By leveraging prompt engineering, LLMs are able to use tools that convert multimodal inputs into a unified modality, typically text, thereby achieving multimodal alignment.
\textit{(4) Data-Driven}. The data-driven approach aims to enhance LLMs' capabilities by training them on domain-specific datasets. For example, training on a point cloud dataset allows the model to understand point clouds. Many researchers adopt this approach by gathering or creating specialized datasets to fine-tune MLLMs, improving their ability to process multimodal data. To maintain comprehensive image perception, data-driven MLLMs often adopt a multimodal converter architecture, with modality perception capabilities derived from the constructed datasets.
    
    Several existing works have systematically discussed the development of MLLMs from the perspectives of data and models \cite{yin2023survey,wu2023multimodal,caffagni2024r,bai2024survey}. Unlike them, we provide a systematic discussion of MLLMs from the perspective of modality alignment. This survey aims to provide a reference for researchers in understanding how MLLMs perceive and process multimodal information, facilitating the selection or design of MLLMs architectures. We provide a comprehensive summary of existing model architectures and trace their development. Additionally, we offer insights into the datasets created or utilized in data-driven MLLMs. Researchers can leverage our analysis to either develop new datasets tailored to specific applications or select existing datasets to achieve targeted capabilities.
       
    Our primary contributions are listed as follows.
    \begin{itemize}
        \item We highlight the significance of methods to bridge the gaps between modalities in MLLMs and provide the first comprehensive survey for giving insight into multimodal information alignment method.
        \item We encompass four methods of bridging the modality gap: multimodal converter, multimodal perceiver, tool learning, and data-driven method, presenting a definition for each of them and tracing their evolution paths.
        \item With the clarification of different methods for the alignment of multimodal information in MLLMs, we discuss the main challenges and future research directions.
    \end{itemize}

\section{Overview}
\label{sec:OverView}
   We categorize recent MLLMs into four groups based on their approach to handling multimodal features: (1) Multimodal Converter, which transforms data into a format that LLMs can understand; (2) Multimodal Perceiver, which enhances how LLMs perceive different types of data; (3) Tool Learning, which leverages tools to convert data into a unified format, typically text; and (4) Data-Driven Methods, which teach LLMs to understand specific types of data in datasets. Existing MLLMs primarily focus on understanding multimodal information such as images. Therefore, this paper concentrates on different methods of image understanding within MLLMs. In addition, we have also reviewed and summarized multimodal large models that involve other modalities, such as audio and video, as discussed in sections including \ref{sec:Any Modalities}.
    
    Based on our classification, we divide existing MLLMs methods for bridging modality gaps into four categories.
We begin with a detailed introduction to the Multimodal Converter, which transforms diverse modalities into a unified object, enabling LLMs to comprehend multimodal features. Next, we introduce MLLMs that leverage the Multimodal Perceiver to process multimodal features, focusing on innovative mechanisms for multimodal perception that allow LLMs to better understand multimodal information.
We also discuss Tool-Learning methods, where LLMs accomplish multimodal tasks by learning to invoke various tools.
Additionally, the Data-Driven method discusses how researchers are adopting strategies to collect or construct domain-specific data to enhance MLLMs ability to understand complex multimodal information in specialized domains.
Finally, we provide a summary and explore potential future directions for MLLMs in bridging modality gaps.

Additionally, we conduct statistical analysis on various aspects of existing MLLMs. As illustrated in Fig. \ref{fig:time-line} of the Appendix, there has been a growing interest in MLLMs over time. According to the data we collected (refer to Tables \ref{tab:StatisticsConvertor}, \ref{tab:StatisticsPerceiver}, \ref{tab:ToolsAssistants}, and \ref{tab:StatisticsDataDriven} in the Appendix), Vicuna \cite{Vicuna} is the preferred foundational LLM for fine-tuning and training. Moreover, in the domain of tool learning methods, GPT4 \cite{GPT4} is the core model most commonly used in studies.
Regarding multimodal information perception, image modalities often rely on encoders such as CLIP \cite{CLIP}, ViT \cite{ViT}, and their variants \cite{ViTG} to process visual data. The current focus in most MLLMs research lies in image perception and facilitating synergies between image and text modalities. The primary supported modalities are text, images, and audio.
In terms of training approaches, researchers tend to favor single-stage training methods, such as single-stage inference and instruction tuning, due to their lower costs. However, models like Flamingo \cite{Flamingo} and GiT \cite{Git} use a two-stage pretraining followed by tuning, while others like FROMAGe \cite{FROMAGe} and ML-MFSL \cite{ML_MFSL} focus exclusively on training the linear layers between input data and LLMs.

Experiment design also plays a crucial role in the performance and practical value of MLLMs. We identify three main experiment types: \textbf{1)P+F (Pretraining and Finetuning)}: Pretraining on large-scale unannotated data to build general knowledge, followed by fine-tuning on specific tasks to optimize performance for domain applications; \textbf{2) IT (Instruction Tuning)}: Using diverse instruction-output pairs to improve the model ability to follow natural language instructions, enhancing generalization in zero-shot and few-shot scenarios;
    \textbf{3) Reasoning without changing model weights}: Utilizing pretrained models directly via prompts or contextual information, offering efficiency and flexibility without additional training.
    We have summarized the experiment types of the 154 MLLMs discussed in this work and presented them in Tables \ref{tab:StatisticsConvertor}, \ref{tab:StatisticsPerceiver}, \ref{tab:ToolsAssistants}, and \ref{tab:StatisticsDataDriven}. This summary highlights the diversity of experimental designs in MLLMs research and provides a framework for understanding their applicability and impact.
    
    \begin{figure}[htbp]
        \centering
        \subfigure[Converter (\ref{sec:Multimodal Converter})]{\includegraphics[height=.21\textwidth]{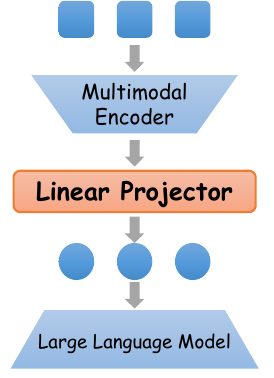}\label{fig:mm-converter}}
        \subfigure[Perceiver (\ref{sec:Multimodal Perceivers})]{\includegraphics[height=.21\textwidth]{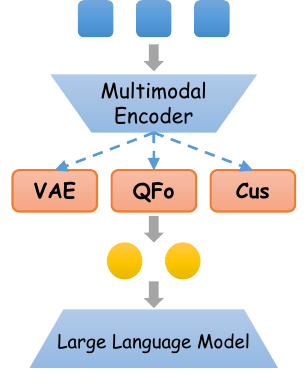}\label{fig:mm-perceiver}}
        \subfigure[Tool Learning (\ref{sec:Tools Assistance})]{\includegraphics[height=.21\textwidth]{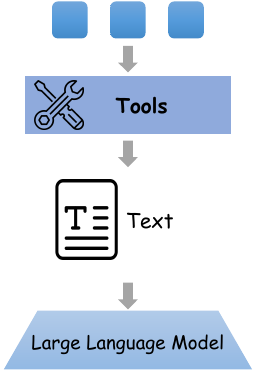}\label{fig:tool-learning}}
        \caption{Structures of Multimodal Perceiver, Multimodal Converter, and Tool Learning. The Converter and Perceiver both transform multimodal inputs into representations within the vector space that LLMs can understand. The difference between the two lies in the transformation process. The former typically involves simply adding a learnable transformation module, usually a linear projector, between the multimodal encoder and the LLMs. Converter often performs a one-to-one transformation, making it more suitable for preserving spatial-related information (as discussed in Section \ref{subsec:Spatial Comprehension} regarding referential capability).
The latter adds a complex perception module between the multimodal encoder and the LLMs. This perception module typically achieves multimodal alignment through a self-attention mechanism, leading to changes in both the features themselves and the number of features, making it more suitable for inputs such as high-resolution or high-noise images.
The main difference between tool learning and the previous two methods is that it achieves multimodal transformation through prompt engineering, requiring no additional training. Moreover, LLMs can utilize tools to perform various multimodal-related tasks, making this approach more versatile.}
    \end{figure}

\section{Multimodal Converter}
\label{sec:Multimodal Converter}
\begin{figure}[htbp]
        \centering
        \subfigure[Feature Projector (\ref{subsec:Feature projector})]{\includegraphics[height=.2\textwidth]{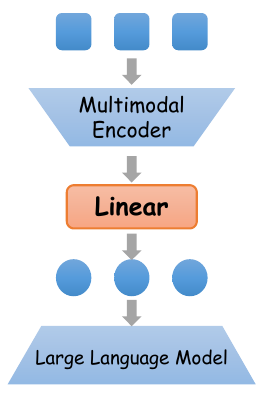}\label{fig:Feature projector}}
        \subfigure[Scaling Up Encoder (\ref{subsec:Scaling up})]{\includegraphics[height=.2\textwidth]{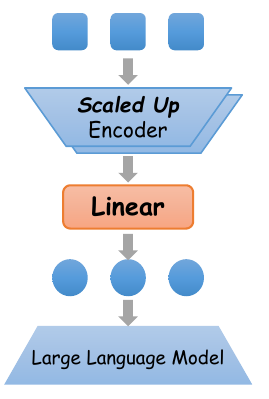}\label{fig:Scaling Up Encoder}}
        \subfigure[Adapter-based Adjustment (\ref{subsec:Adapter-based adjustment})]{\includegraphics[height=.2\textwidth]{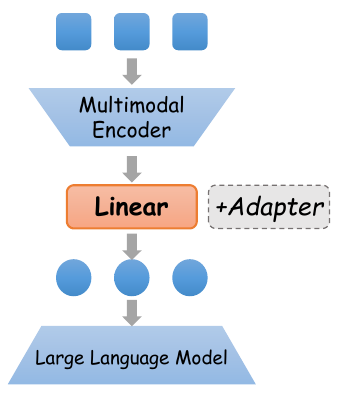}\label{fig:Adapter-based Adjustment}}
        \caption{Structures of feature projector, scaling up encoder, adapter-based adjustment of Converter.}
        \label{fig:1convertor_type}
    \end{figure}

    Given the remarkable capabilities of LLMs, the most straightforward approach to tackling multimodal tasks is to directly input multimodal features into LLMs, allowing it to learn and comprehend these multimodal features. However, since LLMs are primarily trained and learned on generic text, an inevitable semantic gap exists when dealing with multimodal features. Directly injecting these features may lead to severe hallucinations and the generation of answers that deviate from the facts. Consequently, contemporary researchers typically strive to effectively map multimodal features, such as image features, into a feature space that aligns with language, aiming to enhance the performance of MLLMs. We call this type of work as Multimodal Converter, of which structure is shown in Fig. \ref{fig:mm-converter}. Specifically, there are three types of architectures within the Converter, as illustrated in Fig. \ref{fig:1convertor_type}.
    The details of MLLMs involved in the Multimodal Convertor is listed in Table \ref{tab:StatisticsConvertor} in the Appendix. Here, TP (Training Pattern) represents the training methodologies of the MLLMs, encompassing Pretraining, Finetuning, and Reasoning. Modalities refer to the types of modalities supported by the MLLMs.

    In the Converter category, there are three sub-classifications: Feature Projector, Scaling Up, and Adapter-based Adjustment. These approaches all perform one-to-one mapping of features and train linear layers to achieve multimodal feature alignment. However, each method emphasizes different aspects of feature processing.
    1) Feature Projector: These approaches encode features using standard encoders like ViT, followed by linear mapping to achieve alignment.
    2) Scaling Up: These approaches focus on improving image encoders, utilizing large-scale encoders such as ViT-22B or InternViT-6B to enhance image understanding capabilities.
    3) Adapter-based Adjustment: These approaches introduce adapters to perform more fine-grained adjustments to features, particularly in relation to additional modalities.
    
    \subsection{Feature Projector}
    \label{subsec:Feature projector}
    This approach focuses on how to enhance the utilization of images through the simplest weight mapping methods.
    The direct mapping method leverages conventional image encoding methodologies to capture and process visual data.  
    Although this simple approach enables the model to understand images, the issues of hallucination and misalignment of information still represent a direction that urgently requires attention.
    

    In contemporary artificial intelligence research, particularly in the field of image and video understanding, several innovative methods \cite{TKDE1, TKDE2} have been proposed and explored. For instance, the KOSMOS1 \cite{huang2024kosmos1}, KOSMOS2 \cite{peng2023kosmos2}, FROMAGe \cite{FROMAGe}, DreamLLM \cite{DreamLLM}, VisionLLM \cite{visionllm}, Fuyu \cite{fuyu-8b} and NeXT-Chat \cite{zhang2023NExT-Chat} combine visual embeddings of images directly with textual data as input for LLM. This approach represents a straightforward attempt to find a deeper mutual understanding between textual and image data.
    Several methods transform projected features to address the issue of excessively long embeddings in high-resolution images. For instance, MiniGPT2 \cite{chen2023minigptv2} compresses image embeddings by mapping adjacent tokens into a single token, while CogAgent \cite{cogagent} focuses on enabling cross-attention between high-resolution and low-resolution image features.
    Simultaneously, the VideoLLM \cite{VideoLLM} extends this method to video data by encoding each video frame in chronological order and integrating these encoded features with textual data, which could provide a richer context for dynamic video information. Moreover, the VideoChat \cite{Videochat}  employs a more advanced approach by generating textual descriptions for each video frame per second and merging them with video embeddings to form input for LLM. These methods effectively integrate visual video features and text, offering a more comprehensive data input and semantic understanding of videos.
    However, these methods share a common issue: \textit{an over-reliance on the learning capabilities of LLM}. Directly inputting encoded features as vectors presents a challenge when dealing with complex multimodal data. Despite this, these innovative approaches undoubtedly offer new research directions and possibilities in the field of MLLMs.

    Researchers also focus on the hallucination issues brought about by this approach \cite{Hallucination}.  \citet{llava} suggest that this hallucination might arise due to the misalignment of image and text features in the same semantic space. Therefore, to further enhance LLMs' understanding of image features, LLaVA \cite{llava} employs a projection matrix to link the universal visual encoder and the large language model.
    Typically, a two-stage adjustment process is conducted \cite{llava}: First, the LLM parameters are frozen, and the projection matrix is trained to achieve feature alignment between different modalities, resulting in a visual tokenizer for the frozen LLM. Second, end-to-end fine-tuning is performed, freezing only the visual encoder parameters, while both the projection matrix and the large model are involved in training. LLaVA is then extended to LLaVA-Interactive \cite{LLaVA-Interactive} which consists of LLaVA, SEEM \cite{SEEM}, and  GLIGEN \cite{GLIGEN}, thereby enhancing its multimodal capabilities.  
    However, CogVLM \cite{CogVLM} claims that compelling alignment may not always be appropriate. For instance, some details present in the text might not correspond to any information in the image, leading to potential misalignments and hallucinations if strictly enforced. COGVLM \cite{CogVLM} diverges from the conventional approach by independently processing text and image inputs without enforcing direct alignment, employing a new query, key, and value matrix and MLP layer with the text features.

    

    Most data-driven MLLMs ( described in detail in \S \ref{sec:Data-driven MLLMs}) adopt the projector approach. 
    On one hand, the projector is the simplest method, and researchers typically use various data to pretrain the projector. Through the pretraining, the projector learns to map input features into a form that LLMs can effectively process, enabling it to handle complex and diverse data inputs, including medical images~\cite{llavamed}, point cloud images~\cite{pointllm}, GUI image\cite{cogagent}, IMU and fMRI brain activity \cite{OneLLM} and others. For different image formats, some work \cite{llavamed, shikra, lamm} propose constructing specialized datasets to enhance the model understanding of specific types of images. These capabilities are derived from meticulously constructed data.
    On the other hand, the architecture of the projector allows MLLMs to achieve better learning outcomes in certain capabilities, such as Referring Expression Comprehension (REC). Structures like Q-former and resampler face more difficulties than linear projectors when learning REC tasks because they may lose some information \cite{honeybee} and relying solely on constructed data is insufficient. Therefore, most current MLLMs with REC capabilities opt for a linear projector as their model architecture. The equivalent transformation of the linear projector retains all contextual information of visual features, allowing MLLMs to understand regional images.
    \subsection{Scaling Up}
    \label{subsec:Scaling up}
    Traditional MLLMs mainly rely on the powerful capabilities of large language models, but some studies have noticed the disparity in parameter scales between language models and multimodal encoders \cite{chen2024internvl}. Qwen-VL \cite{qwen_vl} uses a large ViT-G \cite{Radford2021CLIP} model with 1.8B parameters to handle image analysis. However, Qwen-VL does not fully consider the impact of the multimodal encoder parameters scale on the overall LLM performance. Subsequent research has paid more attention to this issue. For instance, PALI \cite{chen2022pali} points out that the current image encoder parameter size is too small, limiting the potential of the text component. Therefore, balancing the parameters between the language and vision components becomes crucial. However, retraining a ViT with a massive number of parameters is too costly, so scaling up the parameter size of ViT and other image components is considered to improve overall performance.
    To address this, PALI designs a ViT-e with 4B parameters, which is larger than the previous ViT-G with 1.8B parameters. Experiments have shown that with the language model parameters unchanged, a larger multimodal encoder can effectively enhance the performance of MLLMs. PaLI-X \cite{chen2023paliX} extends this idea further by increasing the size of ViT to ViT-22B, resulting in significant performance improvements. However, InternVL \cite{chen2024internvl} suggests that simply increasing the size of ViT is not the best approach. Therefore, they design InternViT with 6B parameters. Although it has fewer parameters than ViT-22B, it includes an alignment module that aligns the representation of the scaled-up vision encoder with the LLM. By using Qllama as middleware, it smooths the interaction between visual features and textual features.
    \subsection{Adapter-based Adjustment}
    \label{subsec:Adapter-based adjustment}
       In the multimodal understanding learning process of the LLM, not all parameters need to be adjusted. The key is to maintain the linguistic capabilities of the LLM while adjusting the parameters related to multimodality. Therefore, adapter design focuses on adjusting parameters related to additional modalities. Adapters are designed to be user-friendly and minimally impactful to the main LLM parameters and performance.
        Some studies \cite{prophet, llama_adpater, llama_adpterv2, Voxposer, qwen_vl, chen2024lion} do not directly utilize the encoded image information as part of the prompt; instead, they use additional adapters, to initially process and learn from the images. The learned outcomes are then used as inputs to LLM.  
        The adapter-based method involves employing the features encoded from images as a form of adjustment information to influence the original text generation process. 
        For instance, Prophet \cite{prophet} aims to train a model capable of generating heuristic answers, thereby further enhancing LLMs' understanding of task-specific knowledge for various downstream tasks.
        Llama-adapter \cite{llama_adpater} insert lightweight adapters with learnable prompts into $L$ out of $N$ transformer layers of Llama. It uses images to generate multimodal prompts for producing responses conditioned on vision-language inputs, addressing more challenging generative tasks with multimodal comprehension.
        Moreover, instead of using a method that fuses adapter information at each layer, Llama-adapter2 \cite{llama_adpterv2} opts to inject encoded visual tokens and adaptive prompts into different Transformer layers, preserving the independence of information.
        Qwen-VL \cite{qwen_vl} utilizes an adapter to compress the long image feature sequences output by the visual encoder, generating a fixed-length image feature representation, retaining crucial spatial location information through the integration of 2D absolute positional encoding.
        InternLM-XComposer2 \cite{dong2024InterLM-XComposer2} introduces a partial Low-Rank adaptation to convert visual tokens into tokens that can be understood by LLMs.
        LION \cite{chen2024lion} further enhances MLLMs with specialized adapters, treating each as an expert for distinct tasks. Parallel integration and a smart routing module allow dynamic adjustment and information fusion from these adapters, catering to both image and region-level vision-language tasks.

\section{Multimodal Perceiver} 
\label{sec:Multimodal Perceivers}
    
    Similarly, depending on the language understanding of LLMs, the LLMs leveraging multimodal perceiver aims to minimize the semantic gap between images and text by introducing a special multimodal perception module. The perception module is a multimodal perceiver that bridges the gap between text modality and other modalities by transforming multimodal features into multimodal tokens that are consistent with the embedded representation space of LLMs. This process is abstracted in Fig .\ref{fig:mm-perceiver}.
    More sophisticated (but expensive) schemes to connect the image and language representations can also be considered, such as gated cross-attention in Flamingo \cite{Flamingo} and Q-former in BLIP-2 \cite{blip2}, or other vision encoders such as SAM  \cite{SAM} that provide object-level features \cite{llava}. We categorize this type of work into three subcategories based on the main composition patterns of the perceivers:  i) AE Perceiver: These methods proposes using a autoencoder\cite{zhai2018autoencoder} to convert multimodal inputs into representations that LLMs can understand, thereby equipping MLLMs with perception capabilities and ultimately achieving modality alignment; ii) Q-former Perceiver: These methods leverage Q-former to convert multimodal inputs into tokens that
    LLMs can understand; and iii) Customization Perceiver: These method involves non-fixed pattern perceivers. They customizes different perceivers for different requirements to equip MLLMs with multimodal perception capabilities. The statistics of these work are shown in Table \ref{tab:StatisticsPerceiver} in the Appendix.
    \subsection{AE Perceiver}
    \label{subsec:VAE Perceiver}
     This type of method proposes using a autoencoder\cite{zhai2018autoencoder} to convert multimodal inputs into representations that LLMs can understand, thereby equipping MLLMs with perception capabilities and ultimately achieving modality alignment. They offer an approach to modality alignment by using vector quantizers (typically a codebook) to map multimodal inputs into the token space of a LLM. Variational autoencoder (VAE) \cite{VAE} is a strong generative model which is able to reconstruct an image by sampling from prior distributions. It is then developed to discrete VAE (dVAE) \cite{dVAE} and Vector Quantised-VAE (VQ-VAE) \cite{VQVAE} as Visual-Prompt Generator (VPG). They map images to embeddings through a codebook that is similar to word embeddings, providing them with a natural advantage in cross-modal alignment. Consequently, researchers have started employing Variational Autoencoders (VAEs) as multimodal perceivers to align non-text modalities and text modalities \cite{BEIT,LQAE,SPAE,UniCode}. Concretely, BEiT \cite{BEIT} first use dVAE as image tokenizer to transform image to visual tokens. During pre-trainig, BEiT aligns vectors of masked image patches and the visual tokens. LQAE \cite{LQAE} directly uses the pre-trained BERT \cite{BERT} word embeddings as the codebook to connect images and text. This enables LQAE to align text-image in an unsupervised manner. Semantic Pyramid AutoEncoder (SPAE) \cite{SPAE} further develops the VQ-VAE, which is LLM-agnostic and encodes an image into a pyramid of lexical tokens capturing multi-scale representations. In experiment, the SPAE consistently shows higher classification accuracy than the LQAE on mini-ImageNet \cite{vinyals2016matching} few-shot classification benchmark. This inspires us that the hierarchical codebook of VAE can capture semantic information at various levels \cite{He_2017_ICCV}, thus providing richer semantic features for downstream decoders.  The introduction of visual codebooks in these works has resulted in MLLMs containing both visual and textual codebooks, which poses challenges to the training of MLLMs. To address this, UniCode \cite{UniCode} proposes using a unified codebook to integrate different modalities. It enhances MLLMs' multimodal perception capabilities while also improves their multimodal generation abilities. UniCode employs exponential moving average (EMA) \cite{lee2022autoregressive} to smoothly update the codebook, ensuring alignment between the codebook and the visual encoder.
    
    \subsection{Q-former Perceiver}
    \label{subsec:Q-former Perceiver}
        LLMs possess rich pre-trained knowledge; however, they are limited to processing text modality and cannot accommodate other modalities. This limitation constrains their capability to handle multimodal information. Thus researchers have begun exploring approaches to unify multimodal information within a common feature space \cite{blip2}, enabling LLMs to process multimodal tasks cohesively. To this end, Q-former is introduced in BLIP-2 \cite{blip2} as a type of VPG, which achieves multimodal alignment by using an attention mechanism and training on three tasks, consisting of two submodules: the image Transformer and the text Transformer, both sharing two self-attention layers. The image Transformer inputted by a set of learnable query embeddings and image features encoded by an image encoder. These representations interact through cross-attention, and the learnable queries also interact with the input text through shared self-attention layers. The Q-former provides a lightweight solution for multimodal alignment by converting multimodal inputs into tokens that LLMs can understand, even when the multimodal encoder and LLMs are frozen. BLIP-2 Q-former is trained in the first stage with three objectives: 1) image-text contrastive learning, 2) image-grounded txt generation, and 3) iImage-text matching. This equips the Q-former with the ability to fuse image-text modalities. To enable LLMs (e.g., OPT \cite{Opt} and FlanT5 \cite{FlanT5}) to process the fused features from Q-former, BLIP-2 employs a fully connected layer to linearly project the query embeddings outputted from Q-former to LLMs' embeddings. Then the fully connected layer and Q-former are fine-tuned during the second training stage. Following this two-stage training process, BLIP-2 exhibits state-of-the-art performance in various vision-language tasks. With a mere 188 million parameters, BLIP-2 Q-former bridges the gap between image and text modalities, laying the foundation for numerous subsequent MLLMs research endeavors.
        
        After the Q-former of BLIP-2, many works employ Q-former as multimodal perceiver to align various modalities \cite{Video-llama,BLIVA,Minigpt-4,VPG-C,MMICL,InternLM-XComposer,Sparkles,GAVIE,chen2024lion,Macaw-LLM}. For instance, MiniGPT-4 \cite{Minigpt-4} employs a pre-trained EVA-CLIP \cite{EVA-CLIP} and incorporates the Q-former as its visual modality perceiver. The visual modality perceiver is connected to LLMs, exemplified by Vicuna \cite{Vicuna}, through a linear projection layer which serves as the sole trainable module in the two-stage training process of MiniGPT-4. After that, MiniGPT-5 \cite{MiniGPT-5} is proposed which equips the miniGPT-4 with Stable Diffusion 2 \cite{SD2}, consequently enhancing the ability of image generation. Visual Prompt Generator Complete module (VPG-C) \cite{VPG-C} is extended from Q-former which is a generic and lightweight module and designed to capture finer visual details by intercepting the intermediate representations of the LLMs, enhancing the LLMs' comprehension of demonstrative instructions. MMICL \cite{MMICL} utilizes Q-former as VPG to handle images in text-image interleaved inputs, enabling users to provide free-form text and image inputs. Through training on the constructed Multimodal In-Context Learning dataset, MMICL achieves remarkable performance in various general vision-language tasks. BLIVA \cite{BLIVA} simultaneously feeds inputs text embeddings, learned query embeddings from Q-former, and encoded patch embeddings from vision encoder into LLMs, improving performance in various multimodal tasks. The endeavor of BLIVA indicates that the mixture of various multimodal features proves beneficial for LLMs. Except for visual modality, Q-former is also able to transform audio into queries. Video-Llama \cite{Video-llama} utilizes the pre-trained Imagebind \cite{Imagebind} as audio encoder and map the audio features to the embedding space of the LLMs through Q-former and a linear layer. Besides, Q-former also inspires researchers to explore other sophisticated multimodal perceivers. InternLM-XComposer \cite{InternLM-XComposer} is an advanced image-text comprehension and composition MLLMs, of which perceive sampler as a Q-former like module. The perceive sampler is a BERT$_{\text{base}}$ \cite{BERT} model which takes the refined 64 image embeddings as inputs.

    \subsection{Customization Perceiver}
    \label{subsec:Customization Perceiver}
    In addition to the AE and Q-former as multimodal perceivers mentioned earlier, researchers have explored alternative multimodal perceivers to achieve alignment across modalities \cite{Flamingo,Macaw-LLM,visionllm,PandaGPT,SEEM}. This series of works customizes multimodal perceivers for the requirements of multimodal tasks in different scenarios, offering a task-specific way to multimodal alignment. On the one hand, researchers have designed multimodal prompt generators for the MLLMs. For instance, VisionLLM \cite{visionllm} has devised a language-guided image tokenizer that generates visual prompts under instruction. MACAW-LLM \cite{Macaw-LLM} has formulated an alignment module to map image, audio, and video features into LLMs' embedding space. Unlike the typical two-phase training, MACAW-LLM achieves multimodal tasks following human instructions through a single-stage instruction fine-tuning. However, MACAW-LLM lacks quantitative evaluation results. SEEM \cite{SEEM} maps five types different prompts into joint visual-semantic space by design a new prompting scheme, equipping strong generalization capability. TimeChat \cite{ren2024timechat} is equipped with a time-aware frame encoder and a sliding video Q-former as a multimodal prompt generator. This allows TimeChat to extract both visual and timestamp features from video frames and combine the frame features with the timestamp information. ASM \cite{wang2023asm} is equipped with a location-aware image tokenizer, composed of a 12-layer Transformer decoder, designed to extract image features with location information. SPHINX \cite{SPHINX} integrates features from different networks, pre-training paradigms, and information granularities to capture robust visual prompts. After training on the mixed visual features and language instructions, it demonstrates strong multimodal understanding capabilities across a wide range of applications. Mini-Gemini \cite{li2024minigemini} utilizes dual vision encoders to leverage both high- and low-resolution visual features for image-text understanding and generation. mPLUG-Owl \cite{mPLUG-Owl} and mPLUG-Owl2 \cite{mPLUG-Owl2} utilize a visual abstractor module as a generator for visual prompts. The visual abstractor module employs cross-attention between encoded visual features and learnable queries. In contrast to mPLUG-Owl, mPLUG-Owl2 is newly added a Modality-Adaptive Module which project various modalities into a shared semantic space, thereby achieving modality coordination. Vary \cite{wei2023varyscalingvisionvocabulary} proposes to fuse the vocabulary of MLLMs with a new one, thereby achieving more fine-grained visual perception. Monkey \cite{li2024monkey} utilizes a shared resampler to capture both local and global features of an image, enabling it to achieve high resolution image understanding. Flamingo \cite{Flamingo} designs a perceiver resampler module to connect the vision encoder and LLMs. The perceiver resampler architecture resembles the n-layer decoder of Transformers, using flattened visual features and a fixed number of learnable latent queries as input. Within the resampler, latent queries act as queries, while keys and values are formed by concatenating these queries with visual features. The output dimensions match the learnable latent queries, similar to the Q-former multimodal Perceiver, but specifically tailored for transforming multiple visual modalities (e.g., video frames) into fixed visual tokens. This design enhances Flamingo vision-text alignment efficiency, occurring in the GATED XATTN-DENSE layers. Trained on interleaved image-text, image-text pairs, and video-text pairs datasets, Flamingo achieves state-of-the-art performance in image and video understanding tasks with few-shot learning. Building on Flamingo, open-source efforts \cite{OpenFlamingo} and extensions of its architecture have introduced new capabilities with varied datasets \cite{MultiModal-GPT,otter}, as discussed in Section \ref{sec:Data-driven MLLMs}.
    
    On the other hand, researchers have been inspired by the human ability to associate a series of sensory experiences with images or languages. Following CLIP \cite{CLIP} contrastive learning approach that aligns different modalities and images, Imagebind \cite{Imagebind} maps encoded features of modalities such as images, text, audio, thermal images, depth images, and Inertial Measurement Unit (IMU) to a fixed-size embedding space. By aligning the visual modality with other non-visual modalities, Imagebind achieves natural modality alignment and new emergent alignments. The new emergent alignments signify that once ImageBind aligns images with text and audio, its embedding space internally establishes alignment between text and audio. In experimental evaluations, Imagebind has exhibited strong zero-shot performance across various multimodal retrieval and classification tasks. After that, PandaGPT \cite{PandaGPT} employs ImageBind as a multimodal Perceiver, mapping multimodal inputs into the vocabulary embedding space of LLMs (i.e. Vicuna \cite{Vicuna}). Interaction examples provided in PandaGPT demonstrate its capability to receive various modal inputs simultaneously and engage in human-like communication. Unlike ImageBind, LanguageBind \cite{zhu2023languagebind} binds different modalities with language by mapping them to a unified embedding space, thereby aligning the modalities.

\section{Tool Learning}  
\label{sec:Tools Assistance}
    Based on the profound capacity of humans to adeptly employ tools to solve a diverse array of problems, many contemporary research endeavors seek to endow LLMs with the capability to utilize various tools \cite{Toolformer,HuggingGPT,TaskMatrix.AI,IdealGPT,InternGPT,MM-REACT,AssistGPT}, such as foundation models and APIs. In the field of MLLMs, LLMs are encouraged to harness tools for the conversion of different modalities into a unified modality, predominantly text, and ultimately to finish multimodal tasks. Vanilla LLMs are inherently equipped to process textual inputs exclusively. Consequently, researchers have sought to guide LLMs in acquiring tool-usage skills through the construction of textual formats, including natural language, code, and structured text. For instance, TaskMatrix.AI \cite{TaskMatrix.AI} maintains an API platform that provides LLMs with an API library, enabling them to access and leverage external services, databases, and interfaces. These APIs can grant LLMs abilities beyond language, such as multimodal understanding/generation and persistent storage, greatly expanding LLMs’ capabilities and applications. This series of works offers an approach to multimodal alignment by leveraging external tools, enabling the model to acquire capabilities it previously lacked. Based on whether LLMs learn to use tools through natural language or code, this series of efforts can be categorized into three primary approaches: i) Natural language-assisted: these methods enable LLMs to learn how to use and interact with tools through natural language; ii) Code-assisted: these methods focus on helping LLMs understand tasks and generate code to call the tools; and iii) Combined code and natural language-assisted: these method combine the flexibility of natural language with the clarity of code, enabling LLMs to use tools effectively. This collection of research endeavors expands the functional capabilities of LLMs through tool integration, paving the way for future developments in Artificial General Intelligence (AGI). Nonetheless, these initiatives are contingent upon LLMs' In-context Learning (ICL) capacity, implying that LLMs need to attain a certain threshold of model parameters to effectively acquire tool-usage proficiency. The statistics for these works are shown in Table \ref{tab:ToolsAssistants} in the Appendix.
    \subsection{Natural Language Assisted}
    \label{subsec:Natural language assisted}
        Benefiting from the ICL capabilities, LLMs can accomplish a wide range of tasks by utilizing appropriate prompts without any parameter adjustments. Researchers have initiated the construction of tailored natural language prompts to facilitate LLMs in acquiring the usage of various tools for performing multimodal tasks \cite{ChatCaptioner,MM-REACT,HuggingGPT,TaskMatrix.AI,HuggingGPT,Chameleon}. On the one hand, LLMs serve as coordinators, employing natural language to interact with diverse tools. On the other hand, LLMs act as controllers, utilizing natural language for planning and controlling the use of various tools. 
    \subsubsection{LLMs as Coordinators}
    \label{subsec:LLMs as coordinators}
\begin{figure}[tbp]
            \centering
            \includegraphics[width=\linewidth]{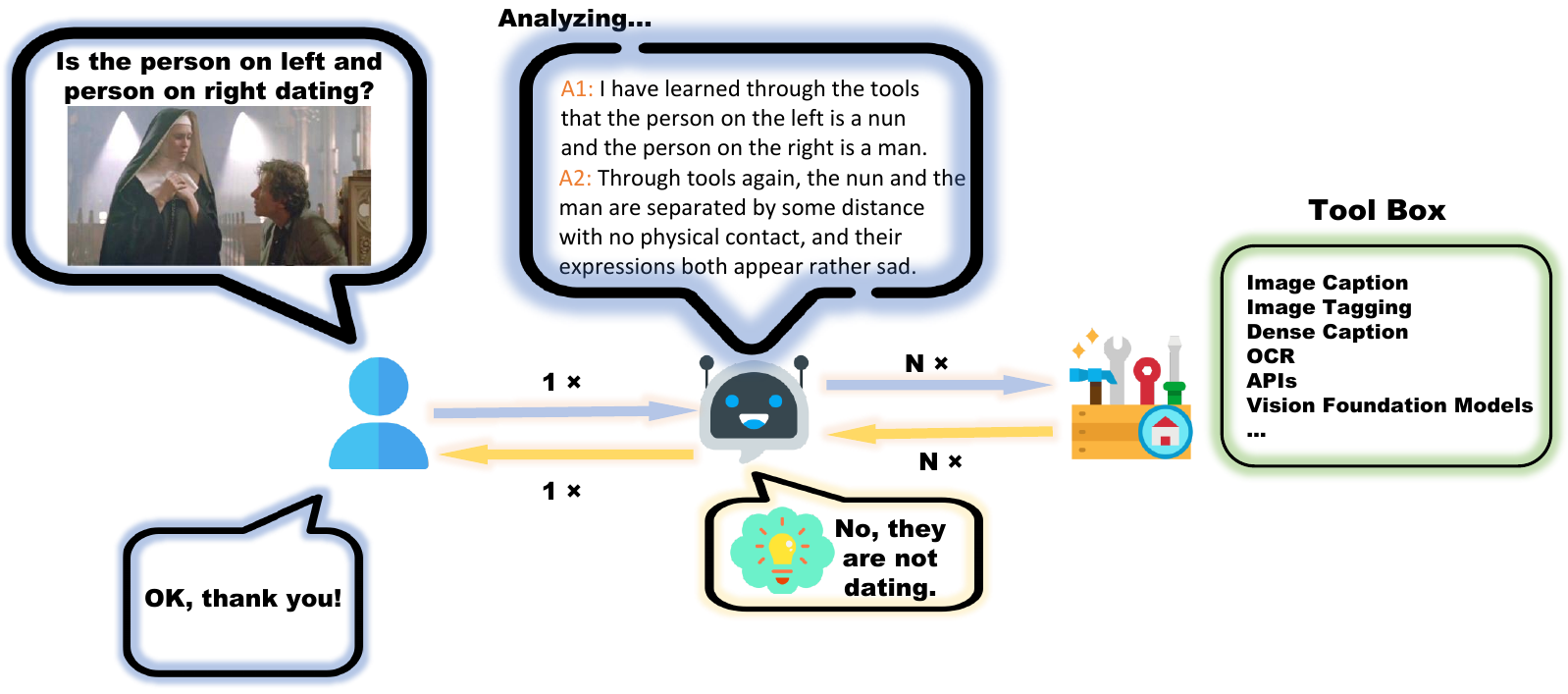}
            \caption{LLMs as coordinators. The example comes from IdealGPT \cite{IdealGPT}.}
            \label{fig:llms_as_oordinators}
        \end{figure}

        LLMs, serving as coordinators, coordinate the utilization of various tools to execute user instructions. LLMs are responsible for direct user interaction, receiving and comprehending user requests, and then invoking tools based on those requests to either obtain their feedback or directly return results to users. If LLMs invoke tools, they subsequently make decisions on whether further tool usage is required or if the results can be directly presented to users. This process is depicted in Fig. \ref{fig:llms_as_oordinators}. An exemplary instance of this approach is demonstrated by MM-REACT \cite{MM-REACT}, which employs ChatGPT as an LLMs. ChatGPT is designated to employ specific watchwords in its responses to indicate the necessity of tool usage, thereby determining the next course of action. Similarly, ChatCaptioner \cite{ChatCaptioner}, and IdealGPT \cite{IdealGPT} also utilize ChatGPT as their LLMs. LLaVA-Plus \cite{LLaVA-Plus} extends the tool utilization capabilities of LLaVA \cite{llava} and maintains a tool repository, enabling LLaVA-Plus to invoke tools on the fly to address multimodal tasks. AVIS \cite{AVIS} leverages PALM 540B \cite{Palm} as its LLM Planner, which is responsible for planning tool usage, and LLM Reasoner for analyzing tool-generated results and making decisions for the next steps. AVIS \cite{AVIS} also introduces a working memory to retain records of user interactions, making it more akin to a personal assistant. ControlLLM \cite{liu2023controlllm} introduces Thoughts-on-Graph for task planning, allowing it to seek optimal solutions for subtasks on a graph. The aforementioned methods \cite{TKDE3, TKDE4} utilize prompt engineering to guide large models in completing tasks without making any adjustments to the parameters of the large models. However, there are also approaches that involve tuning the parameters of large models. One such method is CoVLM \cite{CogVLM}, which adds special tokens to train LLMs to communicate with some expert models for solving vision-language tasks.
    \subsubsection{LLMs as Controllers}
    \label{subsec:LLMs as controllers}
        The distinction between LLMs as controllers and LLMs as coordinators lies in the fact that the former does not re-invoke tools within a single user instruction. LLMs as controllers comprehend user intent and, upon invoking tools, directly respond to the user based on the results obtained from the tools. In other words, in Fig. \ref{fig:llms_as_oordinators}, there are only 1x communications between LLMs and tools. Typical works in this category include HugginGPT \cite{HuggingGPT} and InternGPT \cite{InternGPT}. HugginGPT dissects user input into four distinct steps: 1) Task Planning, 2) Model Selection, 3) Task Execution, and 4) Response Generation. Each of these crucial steps is executed by LLMs using predefined prompt templates, thus making full use of the powerful ICL abilities of LLMs. Similarly, InternGPT employs a similar approach to process user input, but it accommodates more diverse input modalities. Specifically, InternGPT input is not limited to text and images; users can interact with it through gestures such as clicking.  In this regard, CAT \cite{CAT} introduces three interaction modes—Trajectory, Bounding Box, and Points—for users to mark areas of interest within images, thus enhancing the playability of MLLMs. In terms of versatility, Chameleon \cite{Chameleon} presents a plug-and-play compositional reasoning framework with LLMs as core controllers. This framework equips LLMs with tool-usage capabilities to undertake a wide range of tasks. Furthermore, researchers have begun to explore LLMs as controllers for multi-agent systems in gaming interactions, as exemplified by the work on MindAgent \cite{MindAgent}. This underscores the significant potential of LLMs as controllers.
    \begin{figure}[tbp]
            \centering
            \includegraphics[width=\linewidth]{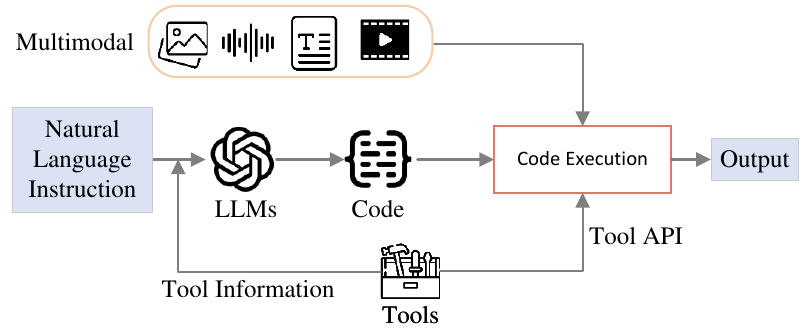}
            \caption{Workflow of code assisted.}
            \label{fig:code_assited}
        \end{figure}
    \subsection{Code Assisted}
    \label{subsec:Code assisted}
        Although LLMs can assist in multimodal tasks by using natural language, their interpretability is poor and they are not suitable for high-precision scenarios due to the inherent complexity and ambiguity of natural language. In contrast, code has better interpretability and can express tasks more precisely. Therefore, some research has started to utilize the code generation capability of LLMs to accomplish multimodal tasks. These methods first use LLMs to generate code based on tool information that meets the requirements of natural language instructions, and then call the tools to process multimodal tasks and obtain the final result. The workflow of this method is shown in Figure \ref{fig:code_assited}. VISPROG \cite{VISPROG} combines symbolic and neurotic approaches, using LLMs to transform natural language into Python-like code, decomposing tasks described in natural language into step-by-step code. This allows users to see intermediate results and provides good interpretability. Later, researchers proposed ViperGPT \cite{ViperGPT}, which converts user natural language instructions into more complex Python code to accomplish tasks step by step. Furthermore, researchers have also used this approach to manipulate robots \cite{Voxposer}. For instance, Voxposer \cite{Voxposer} allows LLMs and VLMs to interact to understand the surrounding environment, reason about feasible actions, identify potential constraints, and ultimately generate code to manipulate robot actions to complete tasks.
    \subsection{Both Code and Natural Language Assisted}
    \label{subsec:Both code and natural language assisted}
        Considering the convenience of natural language and the precision of code, researchers have proposed AssistGPT \cite{AssistGPT}, which employs a sophisticated reasoning approach that combines natural language and code. This approach, called Plan, Execute, Inspect, and Learn (PEIL), consists of four components in AssistGPT: Planner, Executor, Inspector, and Learner. Under the framework of PEIL, AssistGPT is capable of seamlessly utilizing various tools to accomplish multimodal tasks. Specifically, AssistGPT utilizes GPT4 as the Planner to generate a specific path for task completion, producing interleaved natural language and code outputs. The Executor component executes various tools based on the generated code from the Planner. The Inspector manages visual inputs from the user and outputs from the Executor, assisting the Planner in the planning process. The Learner evaluates the execution process of the entire task and records successful trials as in-context examples. Similarly, TaskMatrix.AI \cite{TaskMatrix.AI} employs the Multimodal Conversational Foundation Model as the Planner to interact with users, understand their intentions, and generate code for task execution. In terms of tool integration, AssistGPT standardizes various tools into a common interface, i.e., [Module$\_$Name](``text$\_$query'', ``visual$\_$index''), which facilitates the generation of calling code by the Planner. However, this approach lacks scalability and maintainability. TaskMatrix.AI maintains an API platform that allows API developers to publish and maintain APIs, providing greater scalability. In addition to simply using tools, some researchers have considered continuous learning of tools. CLOVA \cite{gao2024clova} introduces two additional stages—reflection and learning—to ensure it can promptly update tools that need to be updated and adapt to changes of updated tools.
\section{Data-driven MLLMs}
\label{sec:Data-driven MLLMs}
    

    The rapid development of LLM heavily relies on training with extensive datasets. Traditional large models, including multimodal models, are primarily built upon generic datasets, often sourced from unannotated data available on the internet \cite{falcon_refine}. While these datasets cover a wide array of fields, these models lean more towards universal capabilities. However, when faced with more complex multimodal information, such as medical images \cite{llavamed} or the structure of biological molecules \cite{CancerGPT}, the data related to these modalities is few. Consequently, the models lack training and cognition for these specific domains, leading to poor performance, or even an inability to comprehend these modalities.
    \citet{Emergent_Abilities} indicate that with the increase in training parameters and data quantity, models show predictable performance improvements, higher sample utility, and even unpredictable abilities. These unpredictable capabilities are less common in smaller models but have appeared in LLMs. Thus, more and more researchers are adopting a data-driven strategy, collecting or constructing domain-specific data, and using it as the basis for training and fine-tuning LLMs to provide them with additional understanding capabilities for multimodal information.
    Experiments have proven that without changing the model structure, altering the emphasis on content in the training data and the proportion of domain-specific data can endow large models with different capabilities \cite{Llava-med, shikra, lamm, LLaVAR, GAVIE, GPT4ROI}. 

    The information pertaining to MLLMs involved in the data-driven section is listed in Table \ref{tab:StatisticsDataDriven} in the Appendix. Here, AP (Alignment Pattern) denotes the classification of how multimodal information is perceived in MLLMs, specifically including 1) Convertor, 2) Perceiver, and 3) Tools Assistant, along with their subcategories. TP (Training Pattern) represents the training methodologies of the MLLMs, encompassing Pretraining, Finetuning, and Reasoning. Modalities refer to the types of modalities supported by the MLLMs.

    Within data-driven MLLMs, there are various sub-classifications, each designed to endow MLLMs with multimodal perception capabilities through data-driven methods. However, these sub-classifications focus on different aspects:
    i) Enhance Image Comprehension: Enhancing model capabilities or mitigating hallucination issues by providing more detailed textual descriptions, higher-resolution images, or interlacing images and text.
    ii) Spatial Comprehension: Developing datasets that include spatial reference information, aiding in the comprehension of user-specified regions.
    iii) Complex Modalities: Constructing datasets related to complex modalities such as point clouds, remote sensing images, and schematic diagrams from research papers.
    iv) Any Modalities: Expanding multimodal perception capabilities beyond images to other modalities, such as audio, fMRI, depth, thermal, IMU data, etc.
    v) Domain-Specific: Perceiving domain-specific data such as medical images and biomolecular images.
    \subsection{Enhanced Image Comprehension}
    \label{subsec:Enhanced Image Comprehension}
        The capabilities acquired from the additional created data encompass not only the understanding of other modalities but also the further enhancement of comprehension of image data.
        Some researchers focus on creating different forms of image representations, such as complex text-image interleaved formats, higher quality image data, and higher resolution images, to enhance the multimodal understanding of MLLMs.

        \begin{figure}[htbp]
            \centering
            \includegraphics[width=\linewidth]{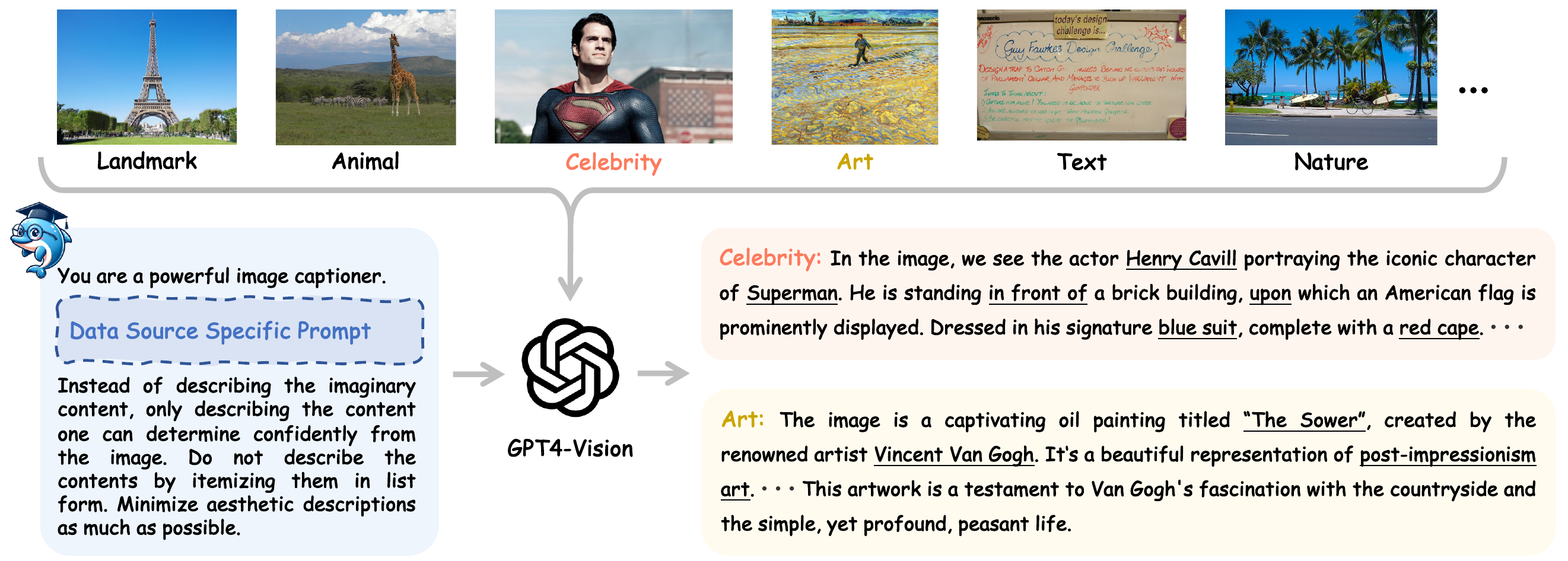}
            \caption{The comparison of rough and detailed image captions from ShareGPT4V \cite{ShareGPT4V}}
            \label{fig:ShareGPT4V}
        \end{figure}

        \subsubsection{Detailed Textual Description}
\begin{figure}[htbp]
            \centering
            \includegraphics[width=.7\linewidth]{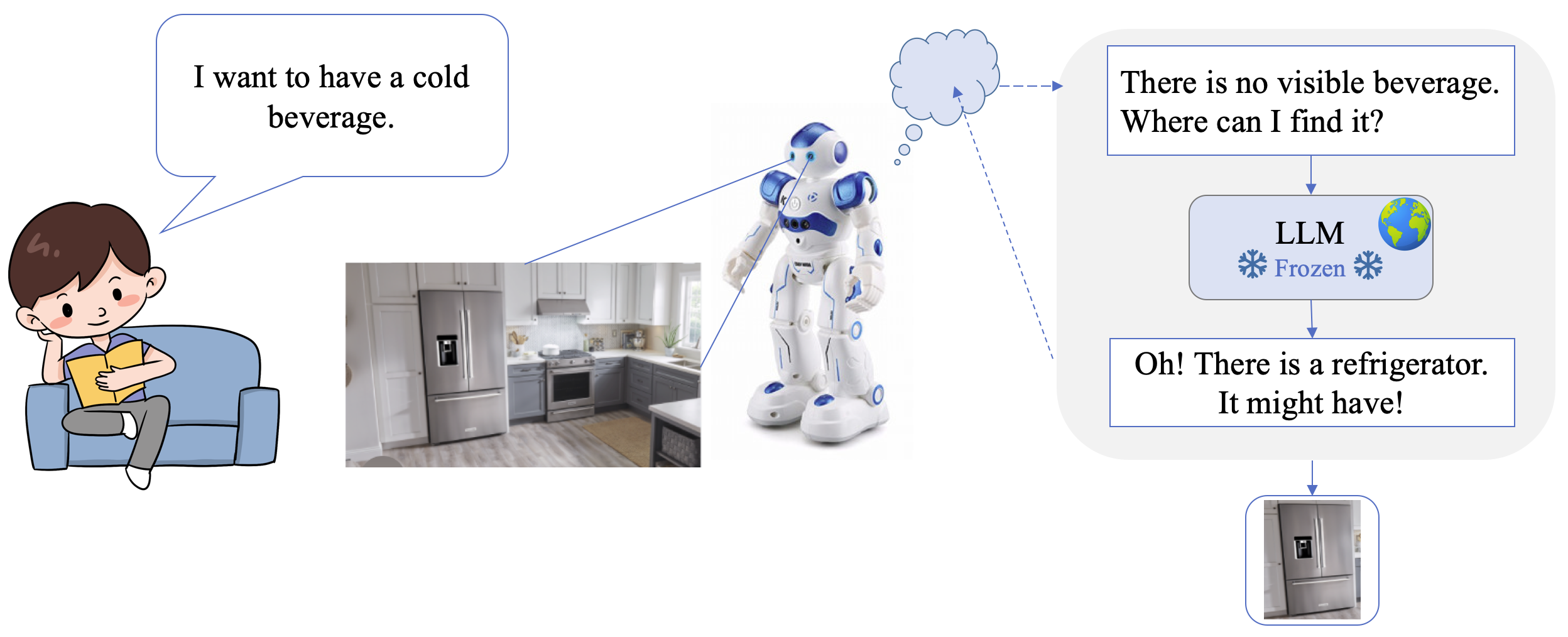}
            \caption{Reasoning-based object detection of DetGPT \cite{pi2023detgpt}.}
            \label{fig:reason ability}
        \end{figure}

        As shown in Fig. \ref{fig:ShareGPT4V}, studies like ShareGPT4V \cite{ShareGPT4V}, ALLaVA \cite{chen2024allava}, and MiniGemini \cite{li2024minigemini} emphasize using detailed textual descriptions to train models for better alignment and multimodal capabilities. Traditional text-image pairs with brief titles often fail to capture detailed correspondences, prompting these works to use GPT-4 for annotating image details, enhancing precise image understanding and reasoning-based generation. Conversely, LLaVA \cite{llava} constructs high-quality instruction-following datasets by replacing images with detailed textual descriptions, relying solely on GPT-4 text understanding.
        \citet{tu2023sight} further demonstrates that high-quality visual-text data improves a model authenticity and ethics, even without explicitly engineering prompts. This is attributed to the quality of visual-text instructions, which align models more closely with human behavior and cognition. High-quality multimodal data also enhances reasoning abilities. For instance, DetGPT \cite{pi2023detgpt} uses GPT to describe image details and build answers, while LISA \cite{lai2024lisa} annotates image-instruction-mask datasets such as ADE20K \cite{zhou2017ADE20K}. Together, these studies (see Fig. \ref{fig:reason ability}) illustrate the critical role of high-quality data in advancing reasoning capabilities in multimodal models.

        \subsubsection{Detailed Image Resolution}
        On the other hand, the high quality of images is reflected in the details of the images themselves, specifically their resolution. For example, models like Osprey \cite{yuan2024osprey}, OtterHD \cite{OtterHD}, InternLM-XComposer2-4KHD \cite{dong2024InternLM-XComposer2-4KHD}, and InterVL1.5 \cite{chen2024InternVL15} use high-resolution images to enhance their performance. Osprey \cite{yuan2024osprey} employs pixel-level instruction tuning, providing the model with a more detailed understanding of images. InternLM-XComposer2-4KHD further recognizes that while the model parses high-resolution image details, it also needs to understand the overall view at low resolution. InternLM-XComposer2-4KHD uses a twin visual encoder mechanism, similar to the design in MiniGemini \cite{li2024minigemini}, where one encoder processes high-resolution images and the other handles low-resolution images. Through this dual-view mechanism, the model can capture both global and local information simultaneously. InterVL1.5 \cite{chen2024InternVL15} uses dynamic resolution technology, enabling the model to support image processing up to 4K resolution.
        Additionally, to address the issue of long embedding vectors caused by high-resolution images, MiniGPTv2 \cite{chen2023minigptv2} and CogAgent \cite{cogagent} adopt a downsampling approach. This technique effectively shortens the vector length by mapping multiple image tokens into a single token, optimizing the model ability to handle high-resolution images.
        
        \subsubsection{Intense Visual-Text Interaction}
        
        In addition, enhancing the interaction between images and text is an important method for data construction. By designing input information in a cross-modal form and increasing the complexity of instructions, the model ability to understand and process image content can be significantly improved. For example, InterLM-XComposer2 \cite{dong2024InterLM-XComposer2} and LLaVAR \cite{LLaVAR} are two representative models in this field. InterLM-XComposer2 \cite{dong2024InterLM-XComposer2} enhances the model understanding of image structure and content by introducing diverse image outlines, detailed textual specifications, and reference images and integrating these various forms of data with text.  These intersecting data allow the model to capture the semantic information of images more comprehensively, thereby improving its ability to parse complex scenes.
        LLaVAR \cite{LLaVAR} takes a different approach by collecting text-rich images to train the model and enhance its ability to recognize and understand text in images. LLaVAR has demonstrated excellent performance in processing images with abundant textual information.
        By enhancing the interaction between images and text, these models can not only more accurately understand the textual information within images but also better handle the complex relationships between images and text.

        \begin{figure}[htbp]
            \centering
            \includegraphics[width=\linewidth]{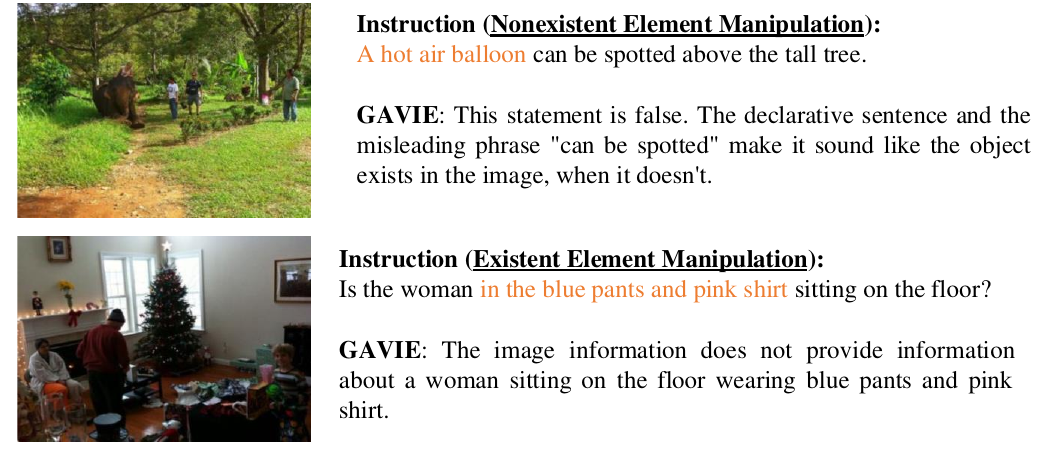}
            \caption{Negative instruction of LRV-instruction dataset from GAVIE \cite{GAVIE}.}
            \label{fig:4_GAVIE_LRVData}
        \end{figure}
        
        \subsubsection{Addressing Hallucination with Detailed Information}
        The hallucination problem, widely studied in the academic community, is addressed by GAVIE \cite{GAVIE}, which attributes it to insufficient dataset diversity and poor-quality web data. To mitigate this, GAVIE introduces the LRV-instruction dataset with 16 vision-related tasks. LRV-instruction includes two types of instructions: 1) Positive instructions (traditional) and 2) Negative instructions (entirely false, involving non-existent targets or insufficient prior information). As shown in Fig. \ref{fig:4_GAVIE_LRVData}, negative instructions aim to train models to assess the correctness of premises, helping them distinguish valid from invalid inputs.
        LLaVA1.5 \cite{liu2024Llava1.5} further demonstrates that increasing input image resolution significantly reduces hallucinations. Higher-resolution images provide more visual detail, improving the model ability to process and interpret content accurately. This finding underscores the importance of balancing detailed data annotations with enhanced information processing capabilities to reduce errors caused by unclear or low-quality images.
        
    \subsection{Spatial Comprehension}
    \label{subsec:Spatial Comprehension}
        
        \begin{figure}[htbp]
            \centering
            \includegraphics[width=\linewidth]{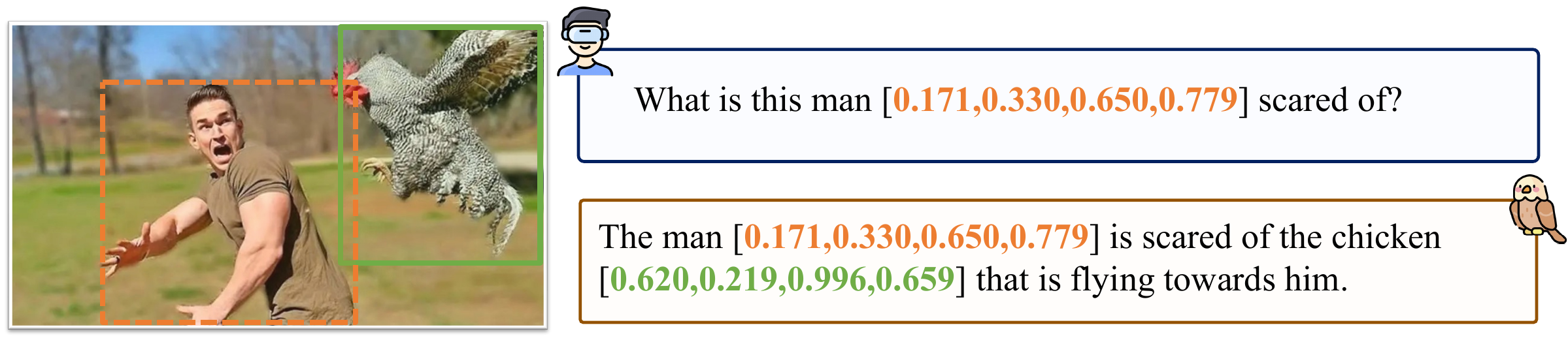}
            \caption{Example of spatial comprehension from Shikra \cite{shikra}.}
            \label{fig:4_shikra_example}
        \end{figure}

        The spatial comprehension ability of MLLMs is important, encompassing the understanding of specific areas within images. This capability spans multiple tasks, including Grounded Captioning \cite{chen2023minigptv2}, Image Captioning, Visual Question Answering (VQA), Referring Expression Generation (REG) \cite{chen2023minigptv2}, Referring Expression Comprehension (REC) \cite{wei2023lenna}, and Referring Object Detection (ROD) \cite{instructGPT, dang2023instructdet}, among others.

        An increasing number of studies \cite{zhang2023NExT-Chat, dang2023instructdet, pi2023detgpt, chen2023pvit, visionllm} are focusing on the spatial comprehension capabilities of MLLMs. For example, as shown in Fig. \ref{fig:4_shikra_example}, Shikra \cite{shikra} constructs Referential Dialogue (RD), enabling the model to extract and understand designated areas within an image. The GPT4RoI \cite{GPT4ROI} enhances the understanding of spatial instructions by converting bounding boxes into spatial instruction formats and using these instructions with language embeddings in an end-to-end method. KOSMOS-2 \cite{peng2023kosmos2} strengthens the grounding ability for rectangular areas within images by integrating the GRIT dataset. GLaMM \cite{GLaMM}, based on GPT-4, automatically creates the annotated dataset GranD, further advancing spatial comprehension.
        Contrasting with other models that focus on understanding input ROIs, Lenna \cite{wei2023lenna} focuses on the REC task, emphasizing the generation of Region of Interest (ROI). CogVLM \cite{CogVLM} constructs a visual anchoring dataset of 40 million images, uses spaCy for part-of-speech tagging, and employs GLIPv2 \cite{zhang2022glipv2} to predict bounding boxes, showing the potential for transferring the capabilities of small models to larger ones. MiniGPTv2 \cite{chen2023minigptv2} showcases its spatial understanding abilities. The ViP-LLaVA \cite{cai2024vipllava} further extends the model spatial capabilities by accommodating diverse visual prompts, such as arrows and circles, in addition to conventional ROI rectangles, utilizing a region-specific dataset constructed with GPT-4V.
        In fact, the data-driven approach, while giving models new modality understanding abilities, can also impact the original capabilities of the models, reducing their understanding of images at a granular level. Therefore, models such as LION \cite{chen2024lion} employ an innovative supervised fine-tuning method to augment the comprehension of region-level REC tasks while preserving their proficiency in original image-level tasks.

    \subsection{Complex Modalities}
    \label{subsec:Complex Modalities}
        MLLMs have shown tremendous potential in handling various information sources such as vision and language. However, they still face significant challenges when dealing with complex modalities such as point cloud data, remote sensing images, paper model diagrams, tables, etc.
    

        \subsubsection{Point Cloud}
        Point cloud data refers to a set of vectors in a three-dimensional coordinate system, which may include color information (RGB) or reflection intensity information (Intensity). Due to their high dimensionality and unstructured nature, point cloud data present higher demands on the MLLMs' understanding and processing capabilities. To enable large language models (LLMs) to understand these diverse modalities, researchers aim to train and align models using point cloud data.
        For example, LAMM \cite{lamm}, based on Vicuna-13B \cite{Vicuna} and CLIP \cite{CLIP}, designs new point cloud datasets and benchmarks to transform visual tasks into instruction-response pairs, significantly enhancing the model ability to extend to new modalities. Additionally, PointLLM \cite{pointllm}, utilizing Point-BERT \cite{yu2021pointbert}, collects 660K simple pairs and 70K complex pairs of point-text instructions, achieving two-stage alignment. The first stage aligns the latent spaces, and the second stage fine-tunes the model with instructions. PointCLIP-V2 \cite{zhu2023pointclip} leverages the powerful expressive capabilities of LLMs to improve performance on 3D depth maps and point clouds.

        \subsubsection{Remote Sensing}

        Remote sensing images are pictures of the Earth surface or atmosphere captured by satellites or airplanes using remote sensing technology. Analyzing these images also benefits from data-driven methods. Remote-Sensing ChatGPT \cite{guo2024RemoteSensingChatGPT} creates visual cues for remote sensing images using various tools \cite{he2016resnet, redmon2016yolov5}, helping the model understand the image content. High-quality (manually labeled) datasets significantly improve the ability to analyze remote sensing images. RSGPT \cite{hu2023rsgpt} and H2RSVLM \cite{pang2024h2rsvlm} are fine-tuned on high-quality datasets RSICap and HqDC-1.4M, respectively, greatly enhancing model performance. H2RSVLM \cite{pang2024h2rsvlm}, in particular, builds a large RSSA dataset to boost the self-awareness of MLLMs, improving model honesty and reducing hallucinations by including unanswerable questions in typical remote sensing visual question-answering tasks.
        Additionally, some works explore extra strategies or architectures to enhance the capability to analyze remote sensing images. For instance, LHRS-Bot \cite{muhtar2024LHRS-Bot}, built on the LHRS-Align and LHRS-Instruct datasets, uses a novel multi-level visual-language alignment strategy and curriculum learning method. SkyEyeGPT \cite{zhan2024skyeyegpt} projects remote sensing visual features into the language domain via an alignment layer, then combines them with task-specific instructions in an LLM-based remote sensing decoder to predict answers for open-ended remote sensing tasks.
    
        \subsubsection{Customised}
        In perceiving other complex modalities, mPLUG-PaperOwl \cite{hu2023mPLUG-PaperOwl} focuses on extracting and interpreting charts from scientific research papers. 
        For visual navigation images, VELMA \cite{schumann2024velma} extracts landmark information and combines it with CLIP to determine locations, converting this location data into text input for LLMs, thereby enhancing the decision-making capabilities. RT-2 \cite{brohan2023rt2}, operating as a visual-language-action model, interprets robot actions as natural language components and undergoes training and inference based on PaLI-X \cite{chen2023paliX} and PaLM-E \cite{PaLM-E} frameworks.

    \subsection{Any Modalities}
    \label{sec:Any Modalities}

    Individuals perceive and exchange information through various modalities, including vision, language, sound, and touch. This section reviews the current progress in training MLLMs on Any-Modality datasets, equipping them with the ability to align across various modalities (Text, Image, Video, etc.)

    \subsubsection{Multimodal Integration and Decision-Making Based on API Perception}
    A direct approach is using API perception and integrated decision-making to handle any modality data. TaskMatrixAI \cite{TaskMatrix.AI} perceives various modalities (including Text, Image, Audio, Video, Location, Code, Game, etc.) using tens of thousands of APIs. Its architecture includes an API platform, selector, and executor, which perceive specific modalities through APIs and convert them into feature tensors for decision-making in a core LLM like GPT-4 \cite{GPT4}. This method achieves multimodal alignment without constructing and training datasets by leveraging existing APIs.

    \subsubsection{Multimodal Learning Based on Alignment Concepts}
    
    To further enhance the understanding of any modality, researchers propose alignment-based methods by training models on datasets that include various modalities. For example, ImageBind \cite{Imagebind} achieves multimodal alignment through image-centered paired data, showing a general capability across modalities and introducing the concept of "\textit{alignment emergence}," where the embedding space aligns the relationships between different modalities after aligning images with other modalities. Based on ImageBind, PandaGPT \cite{PandaGPT} perceives any modality, converting data from audio, video, and images into feature tensors, which are then linearly transformed and input into the Vicuna \cite{Vicuna}. OneLLM \cite{OneLLM} expands the range of compatible modalities to eight types, including images, audio, video, point clouds, depth/normal maps, IMU, and fMRI brain activity. Each modality is tokenized into features using independent tokenizers, followed by a unified encoder and projector. OneLLM \cite{OneLLM} adopts a progressive alignment approach, initially training on image-text data and gradually aligning other modalities with the language model by increasing data complexity, facilitating model training. Unlike traditional two-step fine-tuning and alignment methods, MACAW \cite{Macaw-LLM} uses an alignment module for multimodal alignment. MACAW trains a multimodal feature learning module that can be naturally inserted into LLM input.

    \subsubsection{Constructing Interleaved Any-Modality Datasets}
    To achieve multimodal alignment, researchers explore data construction, proposing various datasets that include interleaved any-modality data. AnyGPT \cite{zhan2024anygpt} introduces the first large-scale any-modality instruction dataset, AnyInstruct-108k \cite{zhan2024anygpt}, which contains multi-turn dialogues interleaved with information from various modalities. The training data, multimodal interleaved instruction data, enhances the model ability to process complex multimodal data. Similarly, NeXT-GPT \cite{NExT-GPT} connects LLMs with multimodal adapters and different diffusion decoders to perceive and generate any combination of text, images, videos, and audio. NeXT-GPT introduces modality switch instruction tuning and manually curates a high-quality dataset, MosIT, that is formally interleaved with any modality data. NeXT-GPT performs alignment on both the encoding and decoding ends, including LLM-centered alignment on the encoding end and instruction-following alignment on the decoding end.

    \subsection{Domain Specific}
    \label{sec:Domain specific}
    In the field of medical imaging, MLLMs are facing challenges in modality understanding due to differences in specialized knowledge. For instance, although medical and biomolecular images are both image data in form, the domain specialized knowledge they contain makes it difficult for general MLLMs to understand deeply. To overcome this, researchers explore ways to adapt these models to specialized fields.
    \subsubsection{Adaptation in Medical Image}
    \citet{llavamed} adapts LLaVA \cite{llava} to the field of medical imaging by collecting and organizing medical image data. 
    The development process of LLaVA-Med \cite{llavamed} first aligns biomedical vocabulary through illustrative data. Then, it uses instruction-following data generated by GPT-4 to gradually master the semantics of open dialogue contexts. This process simulates the way people gradually acquire biomedical knowledge, enabling the model to understand medical images deeply.
    Additionally, the proposal of the Qilin-Med-VL \cite{liu2023Qilin-Med-VL} marks significant MLLMs progress in the Chinese medical field. This model consists of a pre-trained Vision Transformer (ViT) \cite{ViT} and a foundational Llama-7B \cite{llama}. Through two-stage training involving feature alignment and instruction tuning, Qilin-Med-VL significantly improves its ability to analyze medical data. The release of the ChiMed-VL dataset \cite{liu2023Qilin-Med-VL}, which provides over 1M image-text pairs, further enriches Chinese medical multimodal data resources.
    PMC-LLaMA \cite{wu2024PMC-LLaMA}, an open-source medical language model, surpasses ChatGPT in multiple medical question-answering benchmarks through medical knowledge injection and medical-specific instruction tuning. 
    MedAgents \cite{tang2023medagents} further introduces the agent mechanism \cite{wu2023autogenagent}, more aligned with the decision-making processes in the medical field. The framework conducts multi-round discussions through role-playing agents, providing an innovative multidisciplinary collaboration solution. 
    
    \subsubsection{Innovative Applications in Biomolecular Image}
    In biomolecular imaging, CancerGPT \cite{CancerGPT} leverages the DrugComb Portal \cite{zagidullin2019drugcomb} database to construct biomolecular attribute tables, transforming this data into descriptions for binary prediction tasks, such as assessing drug synergy in cell lines. This approach provides new perspectives for biomolecular image analysis and tools for drug discovery. BiomedGPT \cite{zhang2023biomedgpt}, a unified biomedical generative pre-trained transformer, uses the Transformer architecture with multi-head attention, achieving 16 state-of-the-art results across five tasks, including surpassing GPT-4 \cite{GPT4} in radiology evaluations and outperforming Med-PaLM M \cite{tu2024Med-PaLMM} in breast cancer diagnosis and medical visual question answering.
    
    The advancement of domain-specific MLLMs highlights their growing potential. From LLaVA-Med \cite{llavamed} medical image understanding to CancerGPT \cite{CancerGPT} biomolecular applications, and the development of models like PMC-LLaMA \cite{wu2024PMC-LLaMA} and BiomedGPT \cite{zhang2023biomedgpt}, these MLLMs continue to adapt and enhance their domain-specific knowledge capabilities.

    \subsection{Datasets Statistics}
    \label{sec:Datasets Statistics}



    We have summarized and analyzed the datasets \textbf{proposed} or \textbf{utilized} by various MLLMs in \S \ref{sec:Data-driven MLLMs}, and the statistics are shown in Table \ref{tab:NewlyProposedDataset} and \ref{tab:ExistedDataset} in the Appendix.

    As shown in Table \ref{tab:NewlyProposedDataset} in the Appendix, we present the new datasets proposed in MLLMs. The information summarized includes:
    (1) The name of the MLLMs. 
    (2) The name of the proposed dataset. 
    (3) The training pattern (TP), whether it follows the Pretrain+Finetuning (P+F) method or the Instruction Tuning (IT) method.
    (4) The role of the dataset in the respective training stage.
    (5) The size and content of the dataset.
    Besides, the detailed methods of dataset collection, characteristics, and descriptions are provided in Appendix \ref{App:Datasets Collection and Description}. 
    This review of newly proposed datasets aims to offer subsequent researchers insights on how to collect and construct datasets that enhance specific capabilities of models.
    
    As shown in Table \ref{tab:ExistedDataset} in the Appendix, we list the datasets utilized by MLLMs, which are pre-existing datasets employed by these models. This review of datasets utilized in data-driven MLLMs aims to provide subsequent researchers with information on previously existing datasets suitable for training MLLMs.
    
    Table \ref{tab:NewlyProposedDataset} in the Appendix indicates that the P+F model is the primary training approach adopted by data-driven MLLMs. The datasets used for pretraining typically contain around 1M data, such as the 11M data in the GranD \cite{GLaMM} and the 40M data in the Visual Grounding Dataset \cite{CogVLM}. Some datasets, such as Wukong \cite{VisCPM} and Laion-COCO \cite{VisCPM}, even contain hundreds of millions of data. In contrast, the datasets used for finetuning are generally smaller in quantity but higher in quality, often involving extensive human annotation. For example, the LLaVA-instruct-158K \cite{llava} comprises 158,000 unique language-image instruction-following samples, and the RSICap \cite{hu2023rsgpt} includes 2,585 human-annotated images.
    These datasets typically utilize existing images and texts as source data, and employ GPT-4 \cite{GPT4} for instruction design and question-answer pair creation, forming datasets for pretraining or finetuning. This raises a challenge for future research (\S \ref{sec:Future directions and Challenges}): whether GPT-4 itself limits the performance of MLLMs trained on GPT-4 annotated data, and how GPT-4 influences the tendencies, ethics, and behaviors of these MLLMs through the data annotation process.
\section{Future directions and Challenges}
In this section, we discuss the future directions and challenges of MLLMs. The relationship between future directions and MLLMs' lifecycle is presented in the Appendix \ref{apdx:Overview of Future Directions}.
\label{sec:Future directions and Challenges}

    \subsection{A More Refined Way of Bridging}
    
        \subsubsection{More Sophisticated Multimodal Perceiver}\label{sec:More Sophisticated Multimodal Perceiver}
        The current landscape of multimodal perceivers is predominantly influenced by Q-former or Q-former-like architectures. However, it is not definitive that Q-former represents the optimal multimodal perceiver. Although Q-former and its variants have demonstrated robust modal alignment capabilities, they exhibit certain limitations. For instance, they rely on high-quality image-text pairs, and the use of only 32 learnable tokens for representing multimodal features may lead to insufficient expressive capacity. It is worth exploring whether simply increasing the learnable tokens of the Q-former can enhance its representational capacity, thereby improving its performance on multimodal tasks.
        
        Therefore, the future calls for more sophisticated multimodal perceivers. Exploring how to achieve multimodal information alignment and enable the free expression of multimodal features on diverse, unlabeled multimodal data is one of the promising directions.
        
        \subsubsection{Perceiver Adaptive to LLMs}
\label{sec:Perceiver Adaptive to LLMs}
        In traditional multimodal perceiver models, the focus is on narrowing the semantic gap between different modalities of data, such as images and text. These models employ specialized mechanisms to effectively reduce noise elements in images, thereby aligning more accurately with textual data. In contrast, within MLLMs, the multimodal perceivers place greater emphasis on compatibility and adaptability with the input multimodal data. For instance, they may interpret images through parsed instructions or integrate information from images into reasoning processes, like chain-of-thought(CoT), enabling a more profound understanding and processing of multimodal data.
        
        \subsubsection{Unifying Multimodalities}
        \label{sec:Unifying Multimodalities}
        In the future, it may be unnecessary to employ an intermediary adapter like a perceiver. Similar to biological systems, external information could be directly encoded through specific sensory organs into the brain. Multimodal information, post-encoding by dedicated encoders, might be fed directly into a unified model, eliminating the need for intricate alignment processes. Despite Transformer models demonstrating the capability to encode multi-modalities \cite{ViT}, the encoded multimodal features still necessitate alignment with the embedding space of LLMs through fully connected layers. Current model designs are yet incapable of simultaneously handling multimodal features, warranting exploration into a unified model for different modalities.
        
        However, inherent gaps exist between diverse modalities; for instance, text modality exists discretely, while visual modality often exists continuously. The primary challenge in exploring a unified modality model lies in enabling the model to simultaneously process these disparate modalities.
\subsection{Multimodal Data}
\subsubsection{Construction of High-quality Multimodal Data}\label{sec:Construction of High-quality Multimodal Data}
    The quality of the dataset directly determines the performance of a model, and under an equivalent model architecture, higher-quality training data leads to superior model performance \cite{gunasekar2023textbooks}. Despite the construction of numerous multimodal datasets, the assurance of their quality remains a challenge. First, the internet is now filled with multimodal data from platforms like YouTube, Twitter, and TikTok, but much of it remains underutilized. Exploring how to extract high-quality multimodal training data from the internet is a worthwhile direction. 
  \subsubsection{Exploring the Effectiveness of synthetic data}  \label{sec:Exploring the Effectiveness of synthetic data}
    The synthesis of language data \cite{xu2024magpie,chen2024allava,llava,NExT-GPT,zhan2024anygpt} has already been practiced in both LLMs and MLLMs. However, to what extent these synthesis data is helpful is still unknown. Some works like Llava \cite{llava}, Next-GPT utilize GPT-4 to construct textual question-answer pairs. Intuitively, the upper limit of a model trained on data generated by GPT-4 is itself. Whether it is possible to break this limit is also a question worth investigating in both LLMs and MLLMs \cite{xu2024magpie,wang2022self}. 
\subsubsection{Utilization of Unaligned and Incomplete Multimodal Data}
\label{sec:Unaligned and Incomplete Multimodal Data}
Recent advancements in multimodal learning have focused on addressing challenges from unaligned and incomplete data. Studies such as \cite{xu2024reliable, mai2024meta} highlight the importance of handling conflictive information across modalities using evidential learning to assess reliability and improve decision-making. Similarly, \cite{zhao2021telecomnet} demonstrates weakly-supervised methods for leveraging incomplete or noisy data, while works like \cite{hu2023cross, hu2022unsupervised} propose tailored approaches and benchmarks for partially aligned datasets. Together, these efforts emphasize the need to manage multimodal data inconsistencies effectively.

Future research will likely integrate these approaches, focusing on adaptive methods to handle low-quality or incomplete data and generalized benchmarks for diverse multimodal scenarios. As noted in \cite{zhang2024multimodal, xu2019adversarial}, addressing the scarcity of high-quality datasets will drive innovations in weakly-supervised and self-supervised techniques, ensuring robust and scalable multimodal systems for practical applications.
\subsection{More Comprehensive Benchmarks}\label{sec:More Comprehensive Benchmarks}
    Benchmarks serve as the primary ways to assess model quality, and their development significantly influences the advancement of models. For instance, in the domain of LLMs, numerous benchmarks \cite{guo2023evaluating} have propelled progress. In the realm of MLLMs, several benchmarks \cite{MMBench,haydarov2023affective,MME} exist, primarily focusing on the performance of MLLMs in downstream tasks. However, these benchmarks overlook the hallucinations \cite{wang2023evaluation,bang2023multitask}, biases, and security issues inherent in MLLMs. Therefore, future benchmarks should comprehensively evaluate MLLMs, addressing not only downstream tasks but also considering hallucinations, biases, and security concerns. Considering the limitations of automatic evaluation, such as the lack of flexibility, MLLMs might generate content that differs from the labels but, from a human evaluation perspective, this content may be similar to or even better than the labeled content. Crowdsourcing voting \cite{chiang2024chatbotarenaopenplatform} could be also considered as a method to compare and evaluate the quality
 of different MLLMs.
\subsection{Multimodal Agents}\label{sec:Multimodal Agents}
    The multimodal agent represents a precursor to the future development of robots \cite{li2023multimodal}, capable of reacting and acting based on multimodal information. In Section \ref{sec:Tools Assistance}, we reviewed current research on LLMs utilizing tools to accomplish multimodal tasks. These studies serve as pioneering work for the development of multimodal agents, enhancing the tool utilization capabilities of LLMs. Enabling LLMs to analyze and respond to multimodal environments is a crucial pathway toward achieving Artificial General Intelligence (AGI) in the future. In the future, first, it is essential to enhance the capabilities of multimodal agents. Currently, multimodal agent can accomplish simple multimodal tasks with the help of tools, but they are still not able to handle complex multimodal tasks effectively, such as automatically generating voiceovers for videos \cite{yang2024synchronized}. Second, it is necessary to expand the boundaries of multimodal agents' capabilities. At present, multimodal agents mainly process modalities like text, images, videos, and audio. However, the ability to understand dynamic data such as sensor data and operating system data is limited \cite{kim2024health}, which is a crucial factor in determining whether multimodal agents can have broader applications in the future.

\subsection{Green MLLMs}\label{sec:Green MLLMs}
    On the one hand, while MLLMs exhibit robust capabilities in multimodal generation and comprehension, the majority of MLLMs often necessitate a two-stage training process, leading to significant energy consumption during their training. Similar to LLMs, MLLMs need to explore environmentally friendly, energy-efficient, and sustainable development approaches  \cite{zhou2023opportunities}. For instance, leveraging parameter-efficient fine-tuning for efficient tuning of foundation models \cite{llama_adpater, llama_adpterv2}, utilizing editing techniques to rectify factual errors in MLLMs \cite{cheng2023can}, and collaborating with Knowledge Graph to enhance MLLMs' reasoning ability and timeliness in a sustainable way \cite{KeLiang1,KeLiang2}. Effectively harnessing these techniques to update MLLMs efficiently, achieving results comparable to vanilla fine-tuning, poses a challenging task. On the other hand, MLLMs are trending toward having increasingly larger parameters, which means more inference overhead in practical use. In the future, it is worth exploring ways to compress MLLMs through techniques such as quantization, pruning, and distillation \cite{wang2024model}, to reduce inference overhead without compromising their performance.

\subsection{Domain-specific MLLMs}\label{sec:Domain-specific MLLMs}
Currently, there is a considerable and continually growing number of MLLMs \cite{zhiyuansurvey}. While the original intent of MLLMs is to serve humanity, a substantial portion of the research remains focused on the models themselves. The practical application of MLLMs is still limited, with the latest example being GPT-4, which serves as a paradigm for a human AI assistant. In the future, it would be worthwhile to explore domain-specific MLLMs, applying them in various areas such as personal AI assistants, education, medicine, autonomous driving, agriculture, environmental protection, retail, manufacturing, and others. When applying MLLMs in a specific domain, it is essential to first understand the domain needs, gather domain data, and then design the model architecture accordingly, using this data to train the domain-specific MLLMs.

\subsection{Safety of MLLMs}
\label{sec:Safety of MLLMs}
    Security has been a persistent concern in the development of information technology \cite{du2021combating, yi2019incremental, xu2024reliable, han2022trusted}. For MLLMs, key security issues include hallucinations, privacy, ethics, and biases. Hallucinations, common in LLMs, occur when models fabricate answers to unknown facts. Current strategies, such as Reinforcement Learning with Human Feedback (RLHF) \cite{RLHF}, help mitigate false outputs but are costly, highlighting the need for more efficient approaches \cite{yin2023woodpecker, liu2023mitigating}. Privacy concerns stem from the vast datasets required to train MLLMs, which may contain sensitive information, necessitating methods to prevent data leakage \cite{Privacy}. Ethical and bias issues arise from societal biases in training data, which models may perpetuate if not properly filtered \cite{zhuo2023exploring}, requiring solutions at both the data and model levels.
    Trustworthy multimodal learning offers a promising direction to enhance MLLMs with interpretability, fairness, privacy, and robustness \cite{xu2024reliable, han2022trusted}. Future work should focus on improving interpretability to ensure reliable outputs and addressing fairness, privacy, and robustness during training to create unbiased and secure models.

\section{Conclusion}

The paper comprehensively discusses the development and application of Multimodal MLLMs in modality alignment, especially how they integrate LLMs to process data containing various modalities like text and vision. These models demonstrate new abilities such as generating image narratives and visual question answering (VQA), indicating a move towards more realistic human-computer interactions and artificial general intelligence. However, MLLMs still face challenges in processing the semantic gap in multimodal data, which may lead to incorrect generation and pose risks to society. Therefore, selecting appropriate modality alignment methods is crucial to avoid resource wastage and enhance performance.
Thus, we categorize and analyze existing modality alignment methods, highlighting their distinct features and research directions in the early stages of MLLMs development. These methods include (1) Multimodal Converters, (2) Multimodal Perceivers, (3) Tool learning, and (4) Data-Driven methods, each addressing differences between modalities from various angles. Overall, this paper provides valuable insights into understanding and improving MLLMs in processing multimodal data and lays a foundation for future research directions.

\bibliography{LATEX}
\bibliographystyle{IEEEtranN}
\clearpage

\section{Acknowledgment}
This work was partly supported by the Hunan Provincial Natural Science Foundation Projects (No. 2022JJ30668 and No. 2022JJ30046), and also partly supported by the National Key R\&D Program of China (No. 2024YFB4506200).
The science and technology innovation Program of Hunan Province: 2024RC1048.
This work was supported by the National Natural Science Foundation of   China (No.62302144, No.72188101) and the Fundamental Research Funds  For the Central Universities(JZ2024HGTB0251).

\appendix
\subsection{Dataset Description}
\label{App:Datasets Collection and Description}
        \begin{figure*}[tbp]
            \centering
            \includegraphics[width=1\linewidth]{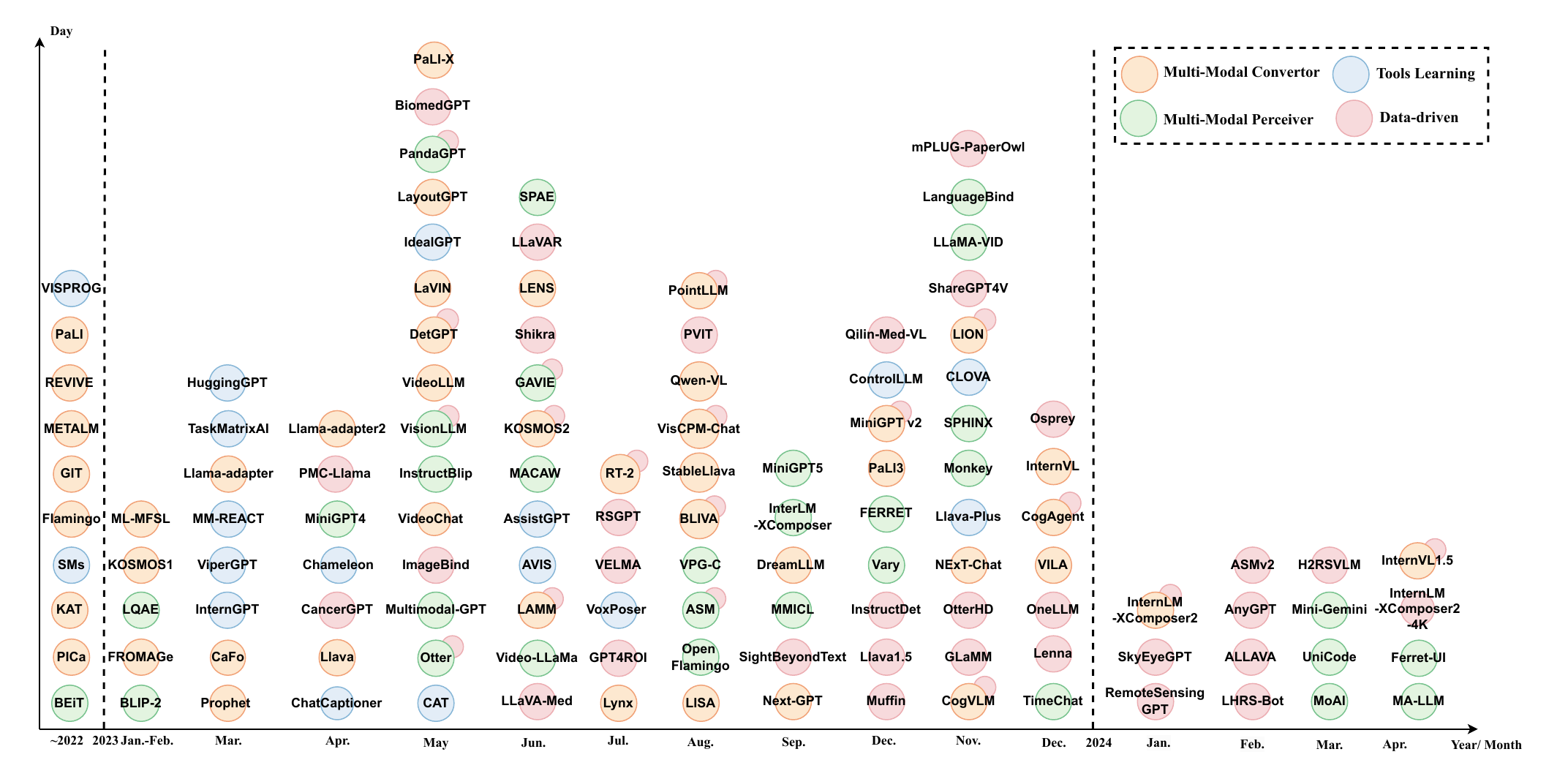}
            \caption{
                Timeline of MLLMs.
                As shown in the figure, an increasing number of researchers have turned their attention to the development of MLLMs over time.
            }
            \label{fig:time-line}
        \end{figure*}

\begin{table*}[htbp]
\centering
  \caption{The Statistics of MLLMs of Convertor Part (\S \ref{sec:Multimodal Converter}). (AP: Alignment Pattern, TP: Training Pattern. P denotes Pretraining, F denotes Finetuning, IT denotes Instruction Tuning, and R denotes Reasoning without changing model weights.)}
\resizebox{0.95\textwidth}{!}{
\begin{tabular}{|p{2.5cm}|p{4cm}|p{3cm}|p{4cm}|p{0.7cm}|p{1.3cm}|p{3cm}|l|}
\toprule
\multicolumn{1}{|c}{\textbf{MLLM}} & \multicolumn{1}{|c}{\textbf{First affiliation}}
& \multicolumn{1}{|c}{\textbf{LLM Backbone}}                                                                
& \multicolumn{1}{|c}{\textbf{Multimodal Encoder}}                       
& \multicolumn{1}{|c}{\textbf{TP}} 
& \multicolumn{1}{|c}{\textbf{Paper}} 
& \multicolumn{1}{|c}{\textbf{Modalities}}
& \multicolumn{1}{|c|}{\textbf{Time}}       \\ 
\midrule
\multicolumn{8}{c}{\textbf{1 Feature Projector}} \\
\midrule
\textbf{LLaVA} \cite{llava}                    & University of Wisconsin–Madison            & Vicuna                            & CLIP ViT-L/14                                                                & P+F         & NeurIPS23                      & Text, Image             & 2023.04.17    \\ \hline
\textbf{ML-MFSL} \cite{ML_MFSL}                 & University of Amsterdam                    & GPT2                              & CLIP                                                                         & IT          & NeurIPS22                      & Text, Image             & 2023.02.28    \\ \hline
\textbf{FROMAGE} \cite{FROMAGe}                 & Carnegie Mellon University                 & OPT                               & CLIP ViT-L/14                                                                & IT          & ICML23                         & Text, Image             & 2023.01.31    \\ \hline
\textbf{VideoLLM} \cite{VideoLLM}               & Nanjing University                         & GPT2, T5, OPT                     & LSTR, Testra, ASFormer, SS-TCN, MS-TCN, Ego4D, VSGN, InternVideo, Moment DETR & IT          & CVPR23                         & Text, Image, Video      & 2023.05.22    \\ \hline
\textbf{VideoChat} \cite{Videochat}               & Shanghai AI Laboratory                     & BLIP2, StableVicuna               & ViT-G                                                                        & P+F         & CVPR24                         & Text, Video             & 2023.05.10    \\ \hline
\textbf{DreamLLM} \cite{DreamLLM}                & Xi’an Jiaotong University                  & Vicuna                            & CLIP-Large, Stable Diffusion                                                  & P+F         & ICLR24                         & Text, Image             & 2023.09.20    \\ \hline
\textbf{CogVLM} \cite{CogVLM}                  & Tsinghua University                        & Vicuna1.5-7B                      & EVA2-CLIP-E                                                                  & P+F         & arXiv                          & Text, Image             & 2023.11.06    \\ \hline
\textbf{PointLLM} \cite{pointllm}                & The Chinese University of Hong Kong        & Vicuna7B, Vicuna 13B              & Point-BERT                                                                   & IT          & arXiv                          & Text, Point Cloud Image & 2023.08.31    \\ \hline
\textbf{METALM} \cite{hao2022metalm}                  & Microsoft Research                         & /                                 & Non-causal Transformer                                                       & P+F         & arXiv                          & Text, Image             & 2022.06.13    \\ \hline
\textbf{KOSMOS1} \cite{huang2024kosmos1}                  & Microsoft Research                         & MATALM                            & CLIP ViT-L/14                                                                & IT          & NeurIPS23                      & Text, Image             & 2023.02.27    \\ \hline
\textbf{KOSMOS2} \cite{peng2023kosmos2}                  & Microsoft Research                         & MATALM                            & CLIP ViT-L/14                                                                & IT          & ICLR24                         & Text, Image             & 2023.06.26    \\ \hline
\textbf{MiniGPT-v2} \cite{chen2023minigptv2}              & KAUST                                      & LLaMA-2 (7B)                      & EVA                                                                          & P+F         & arXiv                          & Text, Image             & 2023.10.14    \\ \hline
\textbf{CogAgent} \cite{cogagent}                & Tsinghua University                        & CogVLM17B                         & EVA2-CLIPE                                                                   & P+F         & arXiv                          & Text, Image, GUI Image  & 2023.12.14    \\ \hline
\textbf{NExT-Chat}  \cite{zhang2023NExT-Chat}              & National University of Singapore           & Vicuna-1.5                        & CLIP ViT-L/14                                                                & P+F         & ICML24                         & Text, Image             & 2023.11.08    \\ \hline
\textbf{DetGPT}    \cite{pi2023detgpt}               & HKUST                                      & Vicuna                            & BLIP-2                                                                       & P+F         & EMNLP23                        & Text, Image             & 2023.05.23    \\ \hline
\textbf{LayoutGPT} \cite{feng2024layoutgpt}               & University of California, Santa Barbara    & Codex, GPT3.5, GPT3.5-chat, GPT4 & GLIGEN                                                                       & R           & NeurIPS23                      & Text, Image, Layout     & 2023.05.24    \\ \hline
\textbf{RT-2} \cite{brohan2023rt2}                    & Google DeepMind                            & PaLI-X, PaLI-E                    & ViT-22B and ViT-4B                                                           & P+F         & PMLR24                         & Text, Image             & 2023.07.28    \\ \hline
\textbf{PointCLIPV2} \cite{zhu2023pointclip}             & City University of Hong Kong               & GPT3                             & CLIP                                                                         & R           & ICCV23                         & Text, Point Cloud Image & 2022.11.21    \\ \hline
\textbf{CaFo} \cite{zhang2023cafo}                   & Shenzhen Institutes of Advanced Technology & GPT3                             & ResNet-50, DALL-E                                                            & R           & CVPR23                         & Text, Image             & 2023.03.03    \\ \hline
\textbf{StableLlava} \cite{li2023stablellava}             & University of Technology Sydney            & LLaVA                             & CLIP-ViT-L/14                                                                & IT          & arXiv                          & Text, Image             & 2023.08.20    \\
\hline
\textbf{MoAI} \cite{lee2024moai}  &KAIST& InternLM-7B&CLIP-L/14 &I&arXiv  &Text, Image, Video, Audio&2024.03.12\\
\hline
\textbf{VisCPM-Chat} \cite{VisCPM}              & Tingshua University                        & Vicuna-13B                        & BEiT-3                                                                       & P+F         & ICLR24                         & Text, Image             & 2023.08.23    \\ 
\midrule
\multicolumn{8}{c}{\textbf{2 Scaling Up}}     \\ 
\midrule
\textbf{VILA} \cite{lin2024vila}                    & NVIDIA                                     & Vicuna-1.5-7B                     & CLIP-L/CLIP-R                                                                & P+F         &  CVPR24 & Text, Image             & 2023.12.12    \\ \hline
\textbf{Qwen-VL} \cite{qwen_vl}                 & Alibaba Group                              & Llama                             & ViT-bigG                                                                     & P+F         & arXiv                          & Text, Image             & 2023.08.24    \\ \hline
\textbf{PaLI} \cite{chen2022pali}                   & Google Research                            & mT5-Large, mT5-XXL                & ViT-G or ViT-e                                                               & P+F         & ICLR23                         & Text, Image             & 2022.09.14    \\ \hline
\textbf{PaLI-X}  \cite{chen2023paliX}                  & Google Research                            & mT5-Large, mT5-XXL                & ViT-22B                                                                      & IT          & arXiv                          & Text, Image             & 2023.05.29    \\ \hline
\textbf{PaLI3} \cite{chen2023pali3}                  & Google Research                            & UL2-3B                            & SigLIP-2B                                                                    & P+F         & arXiv                          & Text, Image             & 2023.10.13    \\ \hline
\textbf{InternVL} \cite{chen2024internvl}              & Shanghai AI Laboratory                     & Vicuna13B                         & InternViT-6B                                                                 & IT          & CVPR24                         & Text, Image             & 2023.12.21    \\ \hline
\textbf{InternVL 1.5} \cite{chen2024InternVL15}            & Shanghai AI Laboratory                     & InternLM2-20B                     & InternViT-6B                                                                 & P+F         & arXiv                          & Text, Image             & 2024.04.25    \\ 
\midrule
\multicolumn{8}{c}{\textbf{3 Adapter-based Adjustment}}\\
\midrule                               
\textbf{Prophet} \cite{prophet}                 & Hangzhou Dianzi University                 & GPT3                              & CLIP                                                                         & IT          & CVPR23                         & Text, Image             & 2023.03.03    \\ \hline
\textbf{Llama-adapter} \cite{llama_adpater}           & Shanghai AI Laboratory                     & Llama                             & ViT                                                                          & IT          & ICLR24                         & Text, Image             & 2023.03.28    \\ \hline
\textbf{Llama-adapter-v2} \cite{llama_adpterv2}         & Shanghai AI Laboratory                     & Llama                             & CLIP, BLIP, DocVQA                                                           & IT          & arxiv                          & Text, Image             & 2023.04.28    \\ \hline
\textbf{VoxPoser} \cite{Voxposer}                & Stanford University                        & GPT4                             & OWL-ViT, XMEM                                                                & IT          & arxiv                          & Text, Image, Video      & 2023.07.12    \\ \hline
\textbf{Qwen-VL} \cite{qwen_vl}                 & Alibaba Group                              & Llama                             & ViT-bigG                                                                     & P+F         & arXiv                          & Text, Image             & 2023.08.24    \\ \hline
\textbf{LION} \cite{chen2024lion}                    & Harbin Institute of Technology             & FlanT5-XL(3B), FlanT5-XXL(11B)    & ViT-G/14, RAM-14M                                                            & IT          & CVPR24                         & Text, Image             & 2023.11.20    \\ \hline
\textbf{LISA} \cite{lai2024lisa}                   & The Chinese University of Hong Kong        & LLaVA-7B-v1-1, LLaVA-13B-v1-1     & SAM(Segment Anything Model)                                                  & IT          & CVPR24                         & Text, Image             & 2023.08.01    \\ \hline
\textbf{LaVIN}  \cite{luo2023LAVIN}                  & Xiamen University                          & LLaMA7B, LLaMA-13B                & ViT-L/14                                                                     & IT          & NeurIPS23                      & Text, Image             & 2023.05.24    \\ \hline
\textbf{Lynx} \cite{Lynx}                    & ByteDance Research                         & LLaMA-7B, Vicuna-7B               & EVA-1B                                                                       & P+F         & arXiv                          & Text, Image, Video      & 2023.07.05    \\ 
\hline
\textbf{InternLM-XComposer2 } \cite{dong2024InterLM-XComposer2}  &Shanghai AI Lab& InternLM2-7B&CLIP ViT-L-14 &P+F&arXiv  &Text, Image, Outlines&2024.01.29\\
\bottomrule

\end{tabular}}
\label{tab:StatisticsConvertor}
\end{table*}

\begin{table*}[htbp]
\centering
  \caption{
    The Statistics of MLLMs of Perceiver Part (\S \ref{sec:Multimodal Perceivers}). (TP: Training Pattern. P denotes Pretraining, F denotes Finetuning, IT denotes Instruction Tuning, and R denotes Reasoning without changing model weights.)  
}
\resizebox{0.95\textwidth}{!}{
\begin{tabular}{|p{2.5cm}|p{4cm}|p{3cm}|p{4cm}|p{0.7cm}|p{1.3cm}|p{3cm}|l|}
\toprule
\multicolumn{1}{|c}{\textbf{MLLM}} & \multicolumn{1}{|c}{\textbf{First affiliation}}
& \multicolumn{1}{|c}{\textbf{LLM Backbone}}                                                                
& \multicolumn{1}{|c}{\textbf{Multimodal Encoder}}                       
& \multicolumn{1}{|c}{\textbf{TP}} 
& \multicolumn{1}{|c}{\textbf{Paper}} 
& \multicolumn{1}{|c}{\textbf{Modalities}}
& \multicolumn{1}{|c|}{\textbf{Time}}       \\ 
\midrule
\multicolumn{8}{c}{\textbf{1 AE}} \\
\midrule
\textbf{BEiT} \cite{BEIT}        &  Harbin Institute of Technology  & BERT & ViT&   P+F    &    ICLR22             &     Text, Image         &  2021.06.15 \\ \hline
\textbf{SPAE} \cite{SPAE}        &  Google Research    & GPT 3.5, PaLM 2  & CLIP(ViT-L/14)&   F    &    NeurIPS23             &     Text, Image         &  2023.06.30 \\ \hline
\textbf{LQAE} \cite{SPAE}        &  UC Berkeley     &       GPT 3.5                      &     ViT-base                                                         &  F     &      NeurIPS23           &  Text, Image     &  2023.02.02 \\ \hline
\textbf{UniCode} \cite{UniCode} &BAAI&Vicuna-7B&ViT&P+I&arXiv&Text, Image&2024.03.14\\
\midrule
\multicolumn{8}{c}{\textbf{2 Q-Former}}\\
\midrule                               
\textbf{BLIP2} \cite{blip2}                 & Salesforce Research  &OPT, FlanT5 & CLIP(ViT-L/14), EVA-CLIP(ViT-g/14)  &P & ICML23 &Text, Image &2023.01.30 \\ \hline
\textbf{MiniGPT4} \cite{Minigpt-4}                 & KAUST  &Llama-2, Vicuna & EVA-CLIP(ViT-g/14)  &P+F& ICLR24 &Text, Image &2023.04.20 \\ \hline
\textbf{Video-Llama} \cite{Video-llama}  & DAMO Academy  &Llama, Vicuna & ViT, ImageBind  &P+F& EMNLP23 &Text, Image, Video, Audio &2023.06.05 \\ 
\hline
\textbf{GAVIE} \cite{GAVIE}  & University of Maryland &MiniGPT4/mPLUG-Owl & ViT  &I& ICLR24 &Text, Image &2023.06.26 \\ 
\hline
\textbf{VPG-C} \cite{VPG-C}  & Zhejiang University &Vicuna-7B, Llama2-7B & EVA-CLIP(ViT-g/14)  &F& ICLR24 &Text, Image &2023.08.08 \\ 
\hline
\textbf{BLIVA} \cite{BLIVA}  & UC San Diego &Vicuna-7B & EVA-CLIP(ViT-g/14)  &P+F& AAAI24 &Text, Image, Video  &2023.08.19 \\ 
\hline
\textbf{MMICL} \cite{MMICL}  & Peking University &FLANT5 & EVA-CLIP(ViT-g/14)  &P+F& ICLR24 &Text, Image  &2023.09.14 \\ 
\hline
\textbf{Sparkles} \cite{Sparkles}  & Sun Yat-sen University &Vicuna & EVA-ViT  &I& ICLR24 &Text, Image  &2023.08.31 \\ 
\hline
\textbf{InterLM-XComposer} \cite{InternLM-XComposer}  & Shanghai AI Lab &InternLM & EVA-CLIP(ViT-g/14)  &P+F& arXiv &Text, Image  &2023.09.26 \\ 
\hline
\textbf{MiniGPT4} \cite{Minigpt-4}  &KAUST&Llama2, Vicuna-13B/7B& EVA-CLIP(ViT-g/14) &P+F& ICLR24 &Text, Image  &2023.04.23 \\ 
\hline
\textbf{MiniGPT5} \cite{MiniGPT-5}  &University of California, Santa Cruz&Llama2, Vicuna & ViT &P+F& arXiv &Text, Image  &2023.09.26 \\ 
\hline
\textbf{InstructBlip} \cite{dai2024instructblip}  &Salesforce Research&FlanT5, Vicuna & ViT-g/14 &I& NeurIPS23 &Text, Image  &2023.05.11 \\ 
\hline
\textbf{LION} \cite{chen2024lion}  &Harbin Institute of Technology, Shenzhen&FlanT5& ViT-G/14, RAM-14M &I& CVPR24 &Text, Image  &2023.11.20 \\ 
\hline
\textbf{MA-LMM} \cite{he2024mallm}  &University of Maryland&Vicuna& ViT-G/14 &F& CVPR24 &Text, Image  &2024.04.08 \\ 
\midrule
\multicolumn{8}{c}{\textbf{3 Customization}}\\
\midrule     
\textbf{Flamingo} \cite{Flamingo}  &Deepmind   & Transformer&NFNet &P+F &NeurIPS22  &Text, Image &2022.04.29 \\
\hline
\textbf{OpenFlamingo} \cite{OpenFlamingo}  &University of Washington   & RedPajama, MPT& CLIP ViT-L/14 &P &arXiv  &Text, Image &2023.08.02 \\
\hline
\textbf{Multimodal-GPT} \cite{MultiModal-GPT}  &Shanghai AI Lab   & Llama7B, OpenFlamingo&CLIP  &I&arXiv  &Text, Image &2023.05.08 \\
\hline
\textbf{VisionLLM} \cite{visionllm}  &Shanghai AI Lab   & Alpaca-7B&ResNet and InternImage-H  &R&NeurIPS23  &Text, Image &2023.05.18 \\
\hline
\textbf{Otter} \cite{otter}  &Nanyang Technological University  & OpenFlamingo&CLIP ViT-L/14 &I&ICLR24  &Text, Image &2023.05.05 \\
\hline
\textbf{PandaGPT} \cite{PandaGPT}  &University of Cambridge  & Vicuna&ImageBind &I&ACL23  &Text, Image, Video, Audio, depth, thermal, and inertial measurement units (IMU) &2023.05.25 \\
\hline
\textbf{MACAW-LLM} \cite{Macaw-LLM}  &Tencent AI Lab & Llama-7b&CLIP, WHISPER &I&arXiv  &Text, Image, Video, Audio&2023.06.15\\
\hline
\textbf{SEEM} \cite{SEEM}  &University of Wisconsin-Madison & UniCL, Florence&FocalT, DaViT &F&NeurIPS23  & Text, Points,
Boxes, Scribbles, Region&2023.04.13\\
\hline
\textbf{mPLUG-Owl} \cite{mPLUG-Owl}  &Alibaba &LLaMA-7B &ViT-L/14 &P+I&arXiv  &Text, Image &2023.04.27\\
\hline
\textbf{mPLUG-Owl2} \cite{mPLUG-Owl2}  &Alibaba &LLaMA-2-7B &ViT-L/14 &P+I&  CVPR24&Text, Image &2023.11.07\\
\hline
\textbf{TimeChat} \cite{ren2024timechat}  &Peking University& LLaMA-2 (7B)&ViT-G/14, Video Q-former &I&CVPR24  &Text, Image, Video&2023.12.04\\
\hline
\textbf{Mini-Gemini} \cite{li2024minigemini}  &The Chinese University of Hong Kong& Mixtral-8×7B, Hermes-2-Yi-34B&ViT, ConvNeXt &P+F&arXiv  &Text, Image&2024.03.27\\
\hline
\textbf{ASM} \cite{wang2023asm}  &Shanghai AI Lab&  Husky-7B&ViT-g/14, RoIAlign&F&arXiv  &Text, Image&2023.08.03\\
\hline
\textbf{Vary} \cite{wei2023varyscalingvisionvocabulary}  &MEGVII Technology & Qwen-7B, Vicuna-7B &SAM ViTDet CLIP-L&P+F&arXiv  &Text, Image &2023.10.11\\
\hline
\textbf{SPHINX} \cite{SPHINX}  &Shanghai AI Lab & LLaMA-2 &CLIP-ViT CLIP-ConveNeXt, DINOv2-ViT, Q-former&P+F&arXiv  &Text, Image &2023.11.13\\
\hline
\textbf{Monkey} \cite{li2024monkey}  &Huazhong University of Science and Technology & QwenVL&Vit-BigG&I&CVPR24  &Text, Image &2023.11.11\\
\hline
\textbf{LLaMA-VID} \cite{li2023llamavidimageworth2}  &CUHK & Llava&EVA-G, QFormer&P+F&ECCV24  &Video, Text &2023.11.28\\
\hline
\textbf{LanguageBind} \cite{zhu2023languagebind}  &Peking University & -&ViTL/14 OpenCLIP&P&ICLR24  &Text, Image, Video, Infrared, Depth, Audio, N-th modality&2023.11.28\\
\hline
\textbf{SEAL} \cite{wu2024v}  &UC San Diego & Vicuna-7B&ViTL/14 OpenCLIP&I&CVPR24  &Text, Image&2023.10.21\\
\hline
\textbf{FERRET} \cite{you2023ferret}  &Columbia University& Vicuna&CLIP-ViT-L/14&F&arXiv  &Text, Image&2023.10.11\\
\hline
\textbf{FERRET-UI} \cite{you2024ferret}  &Apple& Vicuna&CLIP-ViT-L/14&F&arXiv  &Phone UI&2024.04.08\\
\bottomrule
\end{tabular}}

\label{tab:StatisticsPerceiver}
\end{table*}

\begin{table*}[htbp]
\centering
  \caption{
    The Statistics of MLLMs of Tool Learning Part (\S \ref{sec:Tools Assistance}). Reasoning Rounds is divided into single-turn (S) and multi-turn (M). Single-turn means that MLLMs perform a fixed number of reasoning (usually once) when performing a task, while multi-turn indicates an uncertain number of reasoning.
}
\resizebox{0.95\textwidth}{!}{
\begin{tabular}{|p{2.5cm}|p{4cm}|p{3cm}|p{4cm}|p{0.7cm}|p{1.3cm}|p{3cm}|l|}
\toprule
\multicolumn{1}{|c}{\textbf{MLLM}} & \multicolumn{1}{|c}{\textbf{First affiliation}}
& \multicolumn{1}{|c}{\textbf{LLM/VLM Backbone}}                                                                
& \multicolumn{1}{|c}{\textbf{Tool}}                       
& \multicolumn{1}{|c}{\textbf{Reasoning Rounds}} 
& \multicolumn{1}{|c}{\textbf{Paper}} 
& \multicolumn{1}{|c}{\textbf{Modalities}}
& \multicolumn{1}{|c|}{\textbf{Time}}       \\ 
\midrule
\multicolumn{8}{c}{\textbf{1 Natural Language-
Assisted}} \\
\midrule
\textbf{ChatCaptioner} \cite{ChatCaptioner}        &  KAUST    &ChatGPT                    &BLIP-2   &  M     &  arXiv        &   Text, Image           &2023.03.23   \\ \hline
\textbf{MM-REAC} \cite{MM-REACT}   &  Microsoft Azure AI    &     ChatGPT                       & APIs           &  M     &       arXiv          &    Text, Image             & 2023.03.20  \\ \hline
\textbf{IdealGPT} \cite{IdealGPT}        & Columbia University     &    ChatGPT                        & BLIP-2, MiniGPT4, LLaVA     &    M   &    ACL23             &     Text, Image          &2023.05.24   \\ \hline
\textbf{SMs} \cite{SMs}        &  Google     &    GPT3               &    CLIP      &  M     &  ICLR23               &    Text, Image, Audio      &2022.04.01   \\ \hline
\textbf{AVIS} \cite{AVIS}        &   UCLA   &     PALM 540B                       &   PALI, Google Lens, Google Web Search API                  &  M     &   NeruIPS23        &    Text, Image          &2023.06.13   \\ \hline
\textbf{LLaVA-Plus} \cite{LLaVA-Plus}        & Tsinghua University     &     LLaVA                       &  BLIP2, SAM and etc.       &  S  &      arXiv           &    Text, Image          &   2023.11.09\\ \hline
\textbf{HuggingGPT} \cite{HuggingGPT}        &  Zhejiang University    & ChatGPT3.5/4                           &   ToolKit                                               & S      &   NeruIPS23              & Text, Image             &  2023.03.30 \\ \hline
\textbf{InternGPT} \cite{InternGPT}        & Shanghai AI Lab    &  GPT-4, Llama                      & ToolKit                                &  S     &     arXiv            &   Text, Image, Pointing       &  2023.03.09 \\ \hline
\textbf{Chameleon} \cite{Chameleon}        &  UCLA    &  GPT-3.5/GPT-4               &     ToolKit                  &    S   &       NeruIPS23          &     Text, Image         & 2023.04.19  \\ \hline
\textbf{CAT} \cite{CAT}        &  SUSTech &       ChatGPT                     &  SAM, BLIP2  &  S     & arXiv                &  Text, Image, Points, Boxes, Trajectory            &  2023.05.04 \\ \hline
\textbf{ControlLLM} \cite{liu2023controlllm}        &Shanghai AI Lab   &     GPT-3.5                       &   ToolKit      &   M &       arXiv          &       Text, Image, Video, Audio       & 2023.10.26  \\  
\midrule
\multicolumn{8}{c}{\textbf{2 Code-Assisted}}\\
\midrule                               
\textbf{VISPROG} \cite{VISPROG}                 &  Allen Institute for AI  & GPT3& Stable Diffusion, MaskFormer  &S & CVPR23 & Text, Image&2022.11.18\\ \hline
\textbf{ViperGPT} \cite{ViperGPT}                 & Columbia University  & GPT-3.5& GLIP, MiDaS, BLIP-2, X-VLM, Codex  &S & ICCV2023 &Text, Image & 2023.03.14\\ \hline
\textbf{VosPoser} \cite{Voxposer}                 &Stanford University & GPT-4& OWL-ViT, SAM, XMEM &S & PMLR23 &Text, Image, Video & 2023.07.12\\ 
\midrule
\multicolumn{8}{c}{\textbf{3 Combined Code and Natural Language-Assisted}}\\
\midrule  
\textbf{AssistGPT} \cite{AssistGPT}  &National University of Singapore   &GPT-4 &  BLIP2, InstructBLIP, Gounding Dino,  Google OCR  &M &arXiv  &Text, Image, Video &2023.06.14 \\  \hline
\textbf{TaskMatrix.AI} \cite{TaskMatrix.AI}   &Microsoft &  Arbitrary LLMs&APIs & S & arXiv& Text, Image, Video, Audio, Music, Location &2023.03.29  \\ \hline
\textbf{CLOVA} \cite{gao2024clova}                 &  Peking University  & GPT3&  ToolKit &M & CVPR24 & Text, Image&2023.11.18 \\ 
\bottomrule
\end{tabular}}
\label{tab:ToolsAssistants}
\end{table*}

\begin{table*}[htbp]
\centering
  \caption{
    {The Statistics of MLLMs of Data-driven Part (\S \ref{sec:Data-driven MLLMs}) . (AP: Alignment Pattern, TP: Training Pattern. P denotes Pretraining, F denotes Finetuning, IT denotes Instruction Tuning, and R denotes Reasoning without changing model weights.)  }
}
\resizebox{0.95\textwidth}{!}
{ 
\begin{tabular}{|p{2.7cm}|p{5cm}|p{3cm}|p{3cm}|p{2cm}|c|p{1.3cm}|p{3cm}|l|}
\toprule
\multicolumn{1}{|c}{\textbf{MLLM}} & \multicolumn{1}{|c}{\textbf{First Affiliation}}
& \multicolumn{1}{|c}{\textbf{LLM Backbone}}                                                                
& \multicolumn{1}{|c}{\textbf{Multimodal Encoder}}                       
& \multicolumn{1}{|c}{\textbf{AP}}                                             
& \multicolumn{1}{|c}{\textbf{TP}} 
& \multicolumn{1}{|c}{\textbf{Paper}} 
& \multicolumn{1}{|c}{\textbf{Modalities}}
& \multicolumn{1}{|c|}{\textbf{Time}}       \\ 
\midrule
\multicolumn{9}{c}{\textbf{1 Enhanced Image Comprehension}}                                                                                                                                                                                                                                                                                                                                               \\ 
\midrule
\textbf{ShareGPT4V} \cite{ShareGPT4V}              & University of Science and Technology of China       & Vicuna7B, Vicuna13B                                                           & GPT4-vision                              & 1-Projector                                                   & P+F              & arXiv       & Text, Image                                                                     & 2023.11.21 \\ \hline
\textbf{ALLAVA} \cite{chen2024allava}                   & Shenzhen Research Institute of Big Data             & Phi2-2.7B, StableLM2-1.6B, Phi3-mini-128K                                  & CLIP-ViT-L/14@336                        & 1-Projector                                                   & P+F              & arXiv       & Text, Image                                                                     & 2024.02.18 \\ \hline
\textbf{Mini-Gemini} \cite{li2024minigemini}              & The Chinese University of Hong Kong                 & Mixtral-8×7B, Hermes-2-Yi-34B                                                 & ViT, ConvNeXt                            & 2-Customized                                                  & P+F              & arXiv       & Text, Image                                                                     & 2024.03.27 \\ \hline
\textbf{Llava} \cite{llava}                   & University of Wisconsin–Madison                     & Vicuna                                                                        & CLIP ViT-L/14                            & 1-Projector                                                   & P+F              & NeurIPS23   & Text, Image                                                                     & 2023.04.17 \\ \hline
\textbf{SightBeyondText} \cite{tu2023sight}          & University of Chinese Academy of Sciences           & OpenLLaMA, LLaMA, LLaMA2, OpenAlpaca, Vicuna & CLIP ViT-L/14                            & /                                                             & P                & arXiv       & Text, Image                                                                     & 2023.09.13 \\ \hline
\textbf{DetGPT} \cite{pi2023detgpt}                   & HKUST                                               & Vicuna                                                                        & BLIP-2                                   & 1-Projector                                                   & P+F              & EMNLP23     & Text, Image                                                                     & 2023.05.23 \\ \hline
\textbf{GAVIE} \cite{GAVIE}                    & University of Maryland                              & MiniGPT4/mPLUG-Owl                                                            & ViT                                      & 2-QFormer                                                     & IT               & ICLR24      & Text, Image                                                                     & 2023.06.26 \\ \hline
\textbf{Llava1.5} \cite{liu2024Llava1.5}                 & University of Wisconsin–Madison                     & Vicuna v1.5 13B                                                               & CLIP-ViT-L-336px                         & 1-Projector                                                   & P+F              & arXiv       & Text, Image                                                                     & 2023.10.05 \\ \hline
\textbf{Osprey} \cite{yuan2024osprey}                   & Zhejiang University                                 & Vicuna                                                                        & ConvNeXt-L                               & 1-Projector                                                   & P+F              & CVPR24     & Text, Pixel-level image                                                         & 2023.12.25 \\ \hline
\textbf{Otter-HD} \cite{OtterHD}                 & Nanyang Technological University                    & Fuyu-8B                                                                       & ViT                                      & 1-Projector                                                   & IT               & arXiv       & Text, Image                                                                     & 2023.11.07 \\ \hline
\textbf{InternLM4KHD} \cite{dong2024InternLM-XComposer2-4KHD} & Shanghai AI Laboratory                              & InternLM2-7B                                                                  & ViT-Large/14                             & 1-Projector                                                   & P+F              & arXiv       & Text, Image                                                                     & 2024.04.09 \\ \hline
\textbf{InternVL1.5} \cite{chen2024InternVL15}            & Shanghai AI Laboratory                              & InternLM2-20B                                                                 & InternViT-6B                             & 1-Projector                                                   & P+F              & arXiv       & Text, Image                                                                     & 2024.04.25 \\ \hline
\textbf{MiniGPT-v2} \cite{chen2023minigptv2}               & KAUST                                               & LLaMA-2 (7B)                                                                  & EVA                                      & 1-Projector                                                   & P+F              & arXiv       & Text, Image                                                                     & 2023.10.14 \\ \hline
\textbf{CogAgent} \cite{cogagent}                 & Tsinghua University                                 & CogVLM17B                                                                     & EVA2-CLIPE                               & 2-Customized                                                  & P+F              & arXiv       & Text, Image, GUI Image                                                          & 2023.12.14 \\ \hline
\textbf{InternLM-XComposer2} \cite{dong2024InterLM-XComposer2}      & Shanghai Artificial Intelligence Laboratory         & InternLM2-7B                                                                  & CLIP ViT-L-14-336                        & 2-Customized                                                  & P+F              & arXiv       & Text, Image, Outlines                                                           & 2024.01.29 \\ \hline
\textbf{LLaVAR} \cite{LLaVAR}                   & Georgia Tech                                        & Vicuna-13B                                                                    & CLIP-ViT-L/14, CLIP-ViT-L/14-336         & 1-Projector                                                   & IT               & arxiv       & Text, Image                                                                     & 2023.06.29 \\ \hline
\textbf{VisCPM-Chat} \cite{VisCPM}              & Tingshua University                                 & Vicuna-13B                                                                    & BEiT-3                                   & 1-Projector                                                   & P+F              & ICLR24      & Text, Image                                                                     & 2023.08.23 \\ \hline
\textbf{Muffin} \cite{muffin}                   & Tsinghua University                                 & Vicuna-13B                                                                    & BEiT-3                                   & 1-Projector                                                   & P+F              & arXiv       & Text, Image                                                                     & 2023.10.01 \\ \hline
\textbf{ASM v2} \cite{wang2024asmv2}                 & Shanghai AI Laboratory                              & Vicuna-13B                                                                    & CLIP-ViT-L-336px                         & 1-Projector                                                   & P+F              & arXiv       & Text, Image                                                                     & 2024.02.29 \\ 
\midrule
\multicolumn{9}{c}{\textbf{2 Spatial Comprehension}}                                                                                                                                                                                                                                                                                                                                                      \\ 
\midrule
\textbf{Shikra} \cite{shikra}                  & SenseTime Research                                  & Vicuna-7/13B                                                                  & CLIP(ViT-L/14)                           & 1-Projector                                                   & IT               & arxiv       & Text, Image                                                     & 2023.06.27 \\ \hline
\textbf{GPT4ROI} \cite{GPT4ROI}                  & The University of Hong Kong                         & Llama7B                                                                       & CLIP                                     & 1-Projector                                                   & P+F              & arXiv       & Text, Image                                                                     & 2023.07.07 \\ \hline
\textbf{KOSMOS2} \cite{peng2023kosmos2}                  & Microsoft Research                                  & MATALM                                                                        & CLIP ViT-L/14                            & 1-Projector                                                   & IT               & ICLR24      & Text, Image                                                                     & 2023.06.26 \\ \hline
\textbf{GLaMM} \cite{GLaMM}                    & Mohamed bin Zayed University of AI                  & Vicuna7B                                                                  & ViT-H/14 CLIP                            & 1-Projector                                                   & P+F              & arXiv       & Text, Image                                                                     & 2023.11.06 \\ \hline
\textbf{ViP-Llava} \cite{cai2024vipllava}                & University of Wisconsin–Madison                     & Vicuna v1.5                                                                   & CLIP-336px                               & 1-Projector                                                   & P+F              & CVPR24      & Text, Image                                                                     & 2023.12.01 \\ \hline
\textbf{Lenna} \cite{wei2023lenna}                    & Meituan Inc                                         & LLaVA                                                                         & Grounding-DINO                           & 1-Projector                                                   & IT               & arXiv       & Text, Image                                                                     & 2023.12.05 \\ \hline
\textbf{CogVLM} \cite{CogVLM}                   & Tsinghua University                                 & Vicuna1.5-7B                                                                  & EVA2-CLIP-E                              & 1-Projector                                                   & P+F              & arXiv       & Text, Image                                                                     & 2023.11.06 \\ \hline
\textbf{LION} \cite{chen2024lion}                     & Harbin Institute of Technology                      & FlanT5-XL, FlanT5-XXL                                                & ViT-G/14, RAM-14M                        & \begin{tabular}[c]{@{}l@{}}2-Qformer\\ 1-Adapter\end{tabular} & IT               & CVPR24      & Text, Image                                                                     & 2023.11.20 \\ \hline
\textbf{MiniGPT-v2} \cite{chen2023minigptv2}               & KAUST                                               & LLaMA-2 (7B)                                                                  & EVA                                      & 1-Projector                                                   & P+F              & arXiv       & Text, Image                                                                     & 2023.10.14 \\ \hline
\textbf{NExT-Chat} \cite{zhang2023NExT-Chat}                & National University of Singapore                    & Vicuna-1.5                                                                    & CLIP ViT-L/14                            & 1-Projector                                                   & P+F              & ICML24      & Text, Image                                                                     & 2023.11.08 \\ \hline
\textbf{InstructDet} \cite{dang2023instructdet}             & Tongji University                                   & LLaVA(miniGPT4)                                                               & /                                        & 1-Projector                                                   & IT               & ICLR24      & Text, Image                                                                     & 2023.10.08 \\ \hline
\textbf{DetGPT} \cite{IdealGPT}                   & HKUST                                               & Vicuna                                                                        & BLIP-2                                   & 1-Projector                                                   & P+F              & EMNLP23     & Text, Image                                                                     & 2023.05.23 \\ \hline
\textbf{PVIT} \cite{chen2023pvit}                     & Tsinghua University                                 & LLaVA                                                                         & CLIP ViT-L/14, RegionCLIP                & 1-Projector                                                   & P+F              & arXiv       & Text, Image                                                                     & 2023.08.25 \\ \hline
\textbf{VisionLLM} \cite{visionllm}                & Shanghai AI Laboratory                              & Alpaca-7B                                                                     & ResNet,  InternImage-H                   & 2-Customized                                                  & R                & NeurIPS23   & Text, Image                                                                     & 2023.05.18 \\ \hline
\textbf{ASM} \cite{wang2023asm}                      & Shanghai AI Laboratory                              & Husky-7B                                                                      & ViT-g/14,   RoIAlign                     & 2-Customized                                                  & IT               & arXiv       & Text, Image                                                                     & 2023.08.03 \\ \hline
\textbf{Osprey} \cite{yuan2024osprey}                & Zhejiang University                                 & Vicuna                                                                        & ConvNeXt-L                               & 1-Projector                                                   & P+F              & CVPR24     & Text, Pixel-level image                                                         & 2023.12.25 \\ 
\midrule
\multicolumn{9}{c}{\textbf{3 Complex Modalities}}                                                                                                                                                                                                                                                                                                                                                         \\ 
\midrule
\textbf{LAMM} \cite{lamm}                    & Shanghai AI Laboratory                              & Vicuna-13B                                                                    & CLIP(ViT-L/14)              & 1-Projector                                                   & IT               & NeurIPS23   & Text, Point Cloud Image                                                         & 2023.06.11 \\ \hline
\textbf{PointLLM} \cite{pointllm}                & The Chinese University of Hong Kon                  & Vicuna7B, Vicuna 13B                                                          & Point-BERT                               & 1-Projector                                                   & IT               & arXiv       & Text, Point Cloud Image                                                         & 2023.08.31 \\ \hline
\textbf{PointCLIPV2} \cite{zhu2023pointclip}              & City University of Hong Kong                        & GPT-3                                                                         & CLIP                                     & 1-Projector                                                   & P                & ICCV23      & Text, Point Cloud Image                                                         & 2022.11.21 \\ \hline
\textbf{RemoteSensing ChatGPT} \cite{guo2024RemoteSensingChatGPT}    & Wuhan University                                    & ChatGPT                                                                       & BLIP                                     & 3-Coordinators                                                             & R                & IGARSS24    & Text, Remote Sensing Image                                                      & 2024.01.17 \\ \hline
\textbf{RSGPT} \cite{hu2023rsgpt}                    & DAMO Academy, Alibaba Group                         & InstructBLIP                                                                  & EVA-G                                    & 2-qformer                                                     & P+F              & arXiv       & Text, Remote Sensing Image                                                      & 2023.07.28 \\ \hline
\textbf{H2RSVLM} \cite{pang2024h2rsvlm}                  & Wuhan University                                    & LLaVA                                                                         & CLIP-Large                               & 1-Projector                                                   & P+F              & arXiv       & Text, Remote Sensing Image                                                      & 2024.03.29 \\ \hline
\textbf{LHRS-Bot} \cite{muhtar2024LHRS-Bot}                 & Nanjing University                                  & LLaMA2-7B                                                                     & ViT-L/14                                 & 1-Projector                                                   & P+F              & ECCV24      & Text, Remote Sensing Image                                                      & 2024.02.04 \\ \hline
\textbf{SkyEyeGPT} \cite{zhan2024skyeyegpt}             & Northwestern Polytechnical University               & LLaMA2-chat                                                                   & EVA-CLIP                                 & 1-Projector                                                   & IT               & arXiv       & Text, Remote Sensing Image                                                      & 2024.01.18 \\ \hline
\textbf{mPLUG-PaperOwl} \cite{hu2023mPLUG-PaperOwl}           & Alibaba Group                                       & mPLUG-DocOwl                                                                  & ViT-L/14                                 & 1-Projector                                                   & IT               & arXiv       & Text, Image, Table                                                              & 2023.11.30 \\ \hline
\textbf{VELMA} \cite{schumann2024velma}                    & Heidelberg University                               & LLaMa-7b                                                                      & EVA-CLIP                                 & /                                                             & IT               & AAAI24      & Text, Navigation Image                                                          & 2023.07.12 \\ \hline
\textbf{RT-2} \cite{brohan2023rt2}                     & Google DeepMind                                     & PaLI-X and PaLI-E                                                             & ViT-22B,  ViT-4B                         & 1-Projector                                                   & P+F              & PMLR24      & Text, Image, Robots Actions                                                     & 2023.07.28 \\ \hline
\textbf{CogAgent} \cite{cogagent}               & Tsinghua University                                 & CogVLM17B                                                                     & EVA2-CLIPE                               & 2-Customized                                                  & P+F              & arXiv       & Text, Image, GUI Image                                                          & 2023.12.14 \\ 
\midrule
\multicolumn{9}{c}{\textbf{4 Any Modalities}}\\ 
\midrule
\textbf{TaskMatrixAI} \cite{TaskMatrix.AI}            & Microsoft                                           & Millions of APIs                                                              & Tools: Millions of APIs                  & 3-Coordinators                                                & R                & arxiv       & Text, Image, Video, Audio, Music, Location                                      & 2023.03.29 \\ \hline
\textbf{ImageBind} \cite{Imagebind}               & FAIR, Meta AI                                       & OpenCLIP                                                                      & ViT-H 630M                               & 1-Projector                                                   & P                & CVPR23      & Images, text, audio, depth, thermal, and IMU data                               & 2023.05.09 \\ \hline
\textbf{PandaGPT} \cite{PandaGPT}                & University of Cambridge                             & Vicuna                                                                        & ImageBind                                & 2-Customized                                                  & IT               & ACL23       & Text, Image, Video, Audio, depth, thermal, and IMU & 2023.05.25 \\ \hline
\textbf{OneLLM} \cite{OneLLM}                 & The Chinese University of Hong Kong                 & Llama2                                                                        & ViT, Furthest Point Sampling             & 1-Projector                                                   & P+F              & CVPR24      & image, audio, video, point cloud, depth/normal map, IMU and fMRI & 2023.12.06 \\ \hline
\textbf{AnyGPT} \cite{zhan2024anygpt}                 & Fudan University                                    & Llama-2 (7B)                                                                  & SEED tokenizer, SpeechTokenizer, Encodec & 1-Projector                                                   & P+F              & arXiv       & Text, Image, Audio, Music                                                       & 2024.02.19 \\ \hline
\textbf{Next-GPT} \cite{NExT-GPT}                 & National University of Singapore                    & Vicuna                                                                        & ImageBind                                & 1-Projector                                                   & IT               & arXiv       & Text, Image, Video, Audio                                                       & 2023.09.11 \\ \hline
\textbf{MACAW} \cite{Macaw-LLM}                   & Tencent AI Lab                                      & Llama-7b                                                                      & CLIP, Whisper                 & 2-Customized                                                  & IT               & arxiv       & Text, Image, Video, Audio                                                       & 2023.06.15 \\ 
\midrule
\multicolumn{9}{c}{\textbf{5 Domain-Specific}} \\ 
\midrule
\textbf{Llava-Med} \cite{llavamed}               & Microsoft                                           & Vicuna                                                                        & BioMedCLIP                               & 1-Projector                                                   & P+F              & NeurIPS23   & CXR, CT, MRI, gross pathology, histopathology                               & 2023.06.01 \\ \hline
\textbf{CancerGPT} \cite{CancerGPT}                & School of Information University of Texas at Austin & GPT2                                                                          & Not Mentioned                            & 1-Projector                                                   & R                & arXiv       & Text, BioMedicine Image                                                         & 2023.04.18 \\ \hline
\textbf{PMC-Llama} \cite{wu2024PMC-LLaMA}                & Shanghai Jiao Tong University                       & LLaMA                                                                         & Not Mentioned                            & Not Mentioned                                                 & P+F              & arXiv       & Text, BioMedicine Image                                                         & 2023.04.27 \\ \hline
\textbf{BiomedGPT} \cite{zhang2023biomedgpt}                & Lehigh University                                   & BART                                                                          & ResNet-50, ResNet-101                    & 1-Projector                                                   & P+F              & arXiv       & Text, CT, MRI, X-ray                                                   & 2023.05.26 \\ \hline
\textbf{Qilin-Med-VL} \cite{liu2023Qilin-Med-VL}            & Peking University                                   & LLaMA2                                                                        & Clip-ViT-large-patch14-336               & 1-Adapter                                                     & P+F              & arXiv       & Text, BioMedicine Image                                                         & 2023.10.27 \\ 
\bottomrule
\end{tabular}
}
\label{tab:StatisticsDataDriven}
\end{table*}
\begin{table*}[htbp]
\centering
  \caption{
    {The newly proposed datasets of data-driven MLLMs (\S \ref{sec:Datasets Statistics}). More information about the collection methods and description can be found in the Appendix \ref{App:Datasets Collection and Description}. (TP: Training Pattern. P: Pretraining, F: Finetuning, IT: Instruction Tuning.)  }
}
\resizebox{.9\textwidth}{!}
{ 
\begin{tabular}{|m{2.5cm}|c|p{3.5cm}|p{3cm}|p{9cm}|p{5cm}|c|}
    \toprule
    \multicolumn{1}{|c}{\textbf{MLLM}} &
    \multicolumn{1}{|c}{\textbf{TP}} &
    \multicolumn{1}{|c}{\textbf{Modalities}} &
    \multicolumn{1}{|c}{\textbf{Dataset Name}} &
    \multicolumn{1}{|c}{\textbf{Dataset Size}} &
    \multicolumn{1}{|c}{\textbf{Dataset Usage}} &
    \multicolumn{1}{|c|}{\textbf{MoreInfo}}
    \\
    \midrule
    \multicolumn{7}{c}{\textbf{1 Enhanced Image Comprehension \S \ref{subsec:Enhanced Image Comprehension}}} \\
    \midrule
    \textbf{ShareGPT4V} \cite{ShareGPT4V} & P+F   & Text, Image & ShareGPT4V & 1.2M highly descriptive captions & P: ShareGPT4V-PT, F:100K high-quality captions in the ShareGPT4V dataset. & App. \ref{App:ShareGPT4V} \\
    \hline
    \multirow{2}[4]{*}{\textbf{ALLAVA} \cite{chen2024allava} } & \multirow{2}[4]{*}{P+F} & \multirow{2}[4]{*}{Text, Image} & Allava & 1.4M samples & P: ALLaVA-Caption-4V, Visual Instruction Fine-tuning.: ALLaVA-Instruct-4V  & App. \ref{App:Allava} \\
                                                                \cline{4-7}          &       &       & Evol-Intruct-GPT4-Turbo-143K & 143K samples & Pretraining & App. \ref{App:Evol-Intruct-GPT4-Turbo-143K} \\
    \hline
    \textbf{Mini-Gemini} \cite{li2024minigemini} & P+F   & Text, Image & Mini-Gemini & 1.2M image-caption pairs for modality alignment, and 1.5M single or multi-turn dialogues for instruction fine-tuning. Also includes a 13K instruction-following dataset generated by GPT-4 Turbo. & Finetuning & App. \ref{App:Mini-Gemini} \\
    \hline
    \textbf{Llava} \cite{llava} & P+F   & Text, Image & LLaVA-Instruct-158K & 158K unique language-image instruction-following samples. These samples include dialogues, detailed descriptions, and complex reasoning. & Finetuning & App. \ref{App:LLaVA-Instruct-158K} \\
    \hline
    \textbf{DetGPT} \cite{pi2023detgpt} & P+F   & Text, Image & / & consists of 5K images and approximately 30K query-answer pairs. & Finetuning & App. \ref{App: DetGPT Dataset} \\
    \hline
    \textbf{GAVIE} \cite{GAVIE}  & IT    & Text, Image & Large-scale Robust Visual (LRV)-Instruction & 400K visual instructions, covering 16 visual and language tasks, including open-ended instructions and answers. & Instruction Tuning & App. \ref{App: Large-scale Robust Visual (LRV)-Instruction} \\
    \hline
    \textbf{Osprey} \cite{yuan2024osprey} & P+F   & Text, Pixel-level image & Osprey-724K & 724K samples & Finetuning & App. \ref{App: Osprey-724K} \\
    \hline
    \textbf{Otter-HD} \cite{OtterHD} & IT    & Text, Image & MagnifierBench & 283 question-answer pairs, derived from 166 images & Instruction Tuning & App. \ref{App: MagnifierBench} \\
    \hline
    \textbf{InternVL 1.5} \cite{chen2024internvl} & P+F   & Text, Image & High-Quality Bilingual Dataset & /     & Finetuning & App. \ref{App: High-Quality Bilingual Dataset} \\
    \hline
    \textbf{CogAgent} \cite{cogagent} & P+F   & Text, Image, GUI Image & CCS400K (Common Crawl Screenshot 400K) & Contains 400K webpage screenshots and 140M question-answer pairs & P, especially in the enhancement of GUI graphic understanding capabilities. & App. \ref{App: CCS400K (Common Crawl Screenshot 400K)} \\
    \hline
    \textbf{InternLM-XComposer2} \cite{dong2024InterLM-XComposer2} & P+F   & Text, Image, Outlines & Free-form Text-Image Composition & /     & Pretraining & App. \ref{App: Free-form Text-Image Composition} \\
    \hline
    \multirow{2}[4]{*}{\textbf{LLaVAR} \cite{LLaVAR}} & \multirow{2}[4]{*}{P+F} & \multirow{2}[4]{*}{Text, Image} & Noisy Instruction-following Data & 422K samples & P, to improve feature alignment. & App. \ref{App: Noisy Instruction-following Data} \\
                                                            \cline{4-7}          &       &       & GPT-4-based Instruction-following Data & 16K samples & F, to enhance the model's ability to follow complex instructions. & App. \ref{App: GPT-4-based Instruction-following Data} \\
    \hline
    \multirow{3}[2]{*}{\textbf{VisCPM-Chat} \cite{VisCPM}} & \multirow{3}[2]{*}{P+F} & \multirow{3}[2]{*}{Text, Image} & Wukong & 100M image-text pairs & Pretraining & App. \ref{App: Wukong} \\
                                                            \cline{4-7}          &       &       & Laion-COCO & 600M image entries & Pretraining & App. \ref{App: Laion-COCO} \\
                                                            \cline{4-7}          &       &       & Zero  & 20M images and corresponding text descriptions & Pretraining & App. \ref{App: Zero} \\
    \hline
    \textbf{Muffin} \cite{muffin}& P+F   & Text, Image & UniMM-Chat & 117,238 dialogues & Finetuning & App. \ref{App: UniMM-Chat} \\
    \hline
    \textbf{ASM v2} \cite{wang2024asmv2} & P+F   & Text, Image & AS-1B & The dataset includes 1.2B regions in 11M images, covering 3.5M unique semantic tags, as well as 3.3B visual question-answer pairs and 1.2B region captions. & Pretraining & App. \ref{App: AS-1B} \\
    \midrule
    \multicolumn{7}{c}{\textbf{2 Spatial Comprehension \S \ref{subsec:Spatial Comprehension}}} \\
    \midrule
    \textbf{Shikra} \cite{shikra} & IT    & Referential Dialogue, Image & Shikra-RD & 5,922 question-answer pairs & Instruction Tuning & App. \ref{App: Shikra-RD} \\
    \hline
    \textbf{KOSMOS2} \cite{peng2023kosmos2} & P+F   & Text, Image & GRIT(Grounded Image-Text pairs) & Approximately 91M images, 137M text spans, and 114M associated bounding boxes & Pretraining & App. \ref{App: GRIT (Grounded Image-Text pairs)} \\
    \hline
    \textbf{GLaMM} \cite{GLaMM} & P+F   & Text, Image & Grounding-anything Dataset (GranD) & 11M images. The dataset also includes 84M referring expressions, 22M region-based captions, and 11M dense region-based captions & Pretraining & App. \ref{App: Grounding-anything Dataset (GranD)} \\
    \hline
    \textbf{ViP-Llava} \cite{cai2024vipllava} & P+F   & Text, Image & ViP-Bench & 303 image-question pairs & / & App. \ref{App: ViP-Bench} \\
    \hline
    \textbf{Lenna} \cite{wei2023lenna} & IT    & Text, Image & ReasonDet & The training set contains 239 images and 1,326 text segments, and the validation set contains 200 images and 344 text segments & Instruction Tuning & App. \ref{App: ReasonDet} \\
    \hline
    \textbf{CogVLM} \cite{CogVLM} & P+F   & Text, Image & Visual Grounding Dataset & 40M images & Pretraining & App. \ref{App: Visual Grounding Dataset} \\
    \hline
    \textbf{InstructDet} \cite{dang2023instructdet} & IT    & Text, Image & InDET & Includes images from MSCOCO, Flicker, and Objects365, with a total of 120.6K images and 908.4K referred object sets, totaling 3.6M instructions & Instruction Tuning & App. \ref{App: InDET} \\
    \hline
    \textbf{DetGPT} \cite{pi2023detgpt} & P+F   & Text, Image & query-answer instruction dataset & The dataset consists of 5K images and approximately 30K query-answer pairs & Finetuning & App. \ref{App: Query-answer instruction dataset} \\
    \midrule
    \multicolumn{7}{c}{\textbf{3 Complex Modalities \S \ref{subsec:Complex Modalities}}} \\
    \midrule
    \textbf{LAMM} \cite{lamm}  & IT    & Text, Point Cloud Image & Language-Assisted Multi-Modal instruction tuning dataset & 186,098 image-language instruction-response pairs and 10,262 point cloud-language instruction-response pairs & \multicolumn{1}{l|}{IT} & App. \ref{App: Language-Assisted Multi-Modal instruction tuning dataset} \\
    \hline
    \textbf{PointLLM} \cite{pointllm} & IT    & Text, Point Cloud Image & Point-Text Instruction Following Data & 660K simple descriptive instructions and 70K complex instructions, totaling approximately 730K samples & Aligning latent space and IT. & App. \ref{App: Point-Text Instruction Following Data} \\
    \hline
    \textbf{RSGPT} \cite{hu2023rsgpt} & P+F   & Text, Remote Sensing Image & RSICap & 2,585 human-annotated image titles, 100 human-annotated image titles and 936 visual question-answer pairs. & Finetuning & App. \ref{App: RSICap} \\
    \hline
    \multirow{3}[6]{*}{\textbf{H2RSVLM} \cite{pang2024h2rsvlm}} & \multirow{3}[6]{*}{P+F} & \multirow{3}[6]{*}{Text, Remote Sensing Image} & HqDC-1.4M & 1.4M image-caption pairs & Pretraining & App. \ref{App: H2RSVLM} \\
            \cline{4-7}          &       &       & RSSA  & /     & Finetuning & App. \ref{App: H2RSVLM} \\
            \cline{4-7}          &       &       & HqDC-Instruct, RS-Specialized-Instruct & /     & Finetuning & App. \ref{App: H2RSVLM} \\
    \hline
    \multirow{2}[1]{*}{\textbf{LHRS-Bot} \cite{muhtar2024LHRS-Bot}} & \multirow{2}[1]{*}{P+F} & \multirow{2}[1]{*}{Text, Remote Sensing Image} & LHRS-Align & 1.15M meaningful high-quality remote sensing image-text pairs & Pretraining & App. \ref{App: LHRS-Align} \\
            \cline{4-7}          &       &       & LHRS-Instruct & /     & Multi-task Pre-training, F & App. \ref{App: LHRS-Instruct} \\
    \hline
    \textbf{SkyEyeGPT} \cite{zhan2024skyeyegpt} & IT    & Text, Remote Sensing Image & SkyEye-968k & 968k samples & Remote Sensing Image-Text Alignment and Multi-Task Dialogue Fine-tuning. & App. \ref{App: SkyEye-968k} \\
    \hline
    \textbf{mPLUG-PaperOwl} \cite{hu2023mPLUG-PaperOwl} & IT    & Text, Image, Table & M-Paper & Contains 48,688 papers, covering over 15 categories. ( Chart papers:47,583, Chart samples:345,810, Analytical samples:269,758, Outline recommendation samples:84,908) & Instruction Tuning & App. \ref{App: M-Paper} \\
    \midrule
    \multicolumn{7}{c}{\textbf{4 Any-to-any \S \ref{sec:Any Modalities}}} \\
    \midrule
    \textbf{OneLLM} \cite{OneLLM} & P+F   & image, audio, video, point cloud, depth/normal map, IMU and fMRI brain activity & a comprehensive multimodal instruction dataset & Approximately 2M items & Finetuning & App. \ref{App: A comprehensive multimodal instruction dataset} \\
    \hline
    \textbf{AnyGPT} \cite{zhan2024anygpt} & P+F   & Text, Image, Audio, Music & AnyInstruct-108k & 108K multi-turn dialogue samples, including approximately 205K images, 503K audio recordings, and 113K music tracks & Pretraining & App. \ref{App: AnyInstruct-108k} \\
    \hline
    \textbf{Next-GPT} \cite{NExT-GPT} & IT    & Text, Image, Video, Audio & Modality-switching Instruction Tuning (MosIT) & 5K dialogue samples & Instruction Tuning & App. \ref{App: Modality-switching Instruction Tuning (MosIT)}\\
    \hline
    \textbf{MACAW} \cite{Macaw-LLM} & IT    & Text, Image, Video, Audio & MACAW-LLM Instruction Dataset & The MACAW dataset is based on multiple existing datasets: 1.The dataset based on COCO image captions contains approximately 69K examples. 2.The dataset based on Charades and AVSD video captions contains approximately 50K examples. & Instruction Tuning & App. \ref{App: MACAW-LLM Instruction Dataset} \\
    \midrule
    \multicolumn{7}{c}{\textbf{5 Domain-Specific \S \ref{sec:Domain specific}}} \\
    \midrule
    \multirow{2}[1]{*}{\textbf{PMC-Llama} \cite{wu2024PMC-LLaMA}} & \multirow{2}[1]{*}{P+F} & \multirow{2}[1]{*}{Text, BioMedicine Image} & MedC-K & 4.8M biomedical academic papers30K medical textbooks & Data-driven Knowledge Injection. & App. \ref{App: MedC-K} \\
            \cline{4-7}          &       &       & MedC-I & 202M tokens & Finetuning & App. \ref{App: MedC-I} \\
    \hline
    \multirow{3}[3]{*}{\textbf{Qilin-Med-VL} \cite{liu2023Qilin-Med-VL}}  & \multirow{3}[3]{*}{P+F}   & \multirow{3}[3]{*}{Text, BioMedicine Image} & \multirow{3}[3]{*}{ChiMed-VL} & ChiMed-VL contains two subsets: 1. Concept Alignment (ChiMed-VL-Alignment), 580,014 image-text pairs. 2. Instruction Adjustment (ChiMed-VL-Instruction), 469,441 question-answer pairs & Medical Visual-Linguistic Feature Alignment and IT & App. \ref{App: ChiMed-VL} \\
    \bottomrule
    \end{tabular}%
}
\label{tab:NewlyProposedDataset}
\end{table*}
    
\begin{table*}[htbp]
\centering
  \caption{
    {The Datasets used of data-driven MLLMs (\S \ref{sec:Datasets Statistics}). (TP: Training Pattern.)  }
}
\resizebox{0.9\textwidth}{!}
{ 
\begin{tabular}{|m{3cm}|p{1cm}|p{2.5cm}|p{75em}|}
    \toprule
    MLLM  & \textbf{TP} & \multicolumn{1}{c|}{\textbf{Modalities}} & \multicolumn{1}{c|}{\textbf{Dataset Description}} \\
    \midrule
    \multicolumn{4}{c}{\textbf{1 Enhanced Image Comprehension \S \ref{subsec:Enhanced Image Comprehension}}} \\
    \midrule
    \textbf{SightBeyondText} \cite{tu2023sight} & P     & Text, Image & Based on existing datasets such as CC3M, LLaVA \cite{llava}, TruthfulQA, Ethics, etc. \\
    \hline
    \textbf{DetGPT} \cite{pi2023detgpt} & P+F   & Text, Image & Based on existing datasets such as SBU, LAION and Conceptual Caption\\
    \hline
    \textbf{Llava1.5} \cite{liu2024Llava1.5} & P+F   & Text, Image & VQA: Visual Question Answering dataset, used for visual question answering tasks. GQA: Gathering and leveraging knowledge for real-world visual reasoning and compositional question answering. OKVQA: Open Knowledge Visual Question Answering dataset, which requires the model to use external knowledge for answering questions. OCRVQA: Optical Character Recognition Visual Question Answering dataset, involving text reading within images. RefCOCO: A dialogue dataset for region-level understanding. Visual Genome: A dataset containing extensive image annotations. ShareGPT \cite{ShareGPT}: A dataset containing multi-turn dialogues. \\
    \hline
    \textbf{InternLM-XComposer2-4KHD} \cite{dong2024InternLM-XComposer2-4KHD} & P+F   & Text, Image & ShareGPT4V-PT \cite{ShareGPT4V}, COCO, Nocaps, TextCaps \cite{sidorov2020textcaps}, LAION400M, SBU, CC3M - Used for general semantic alignment and world knowledge alignment.\newline{}WanJuan, Flicker, MMC-Inst, RCTW-17, CTW, LSVT, ReCTs, ArT - Used for enhancing visual capabilities.\newline{}DocVQA, ChartQA, InfoVQA, TextVQA, OCRBench - Used for high-resolution understanding tasks. \\
    \hline
    \textbf{InternVL 1.5} \cite{chen2024InternVL15} & P+F   & Text, Image & Laion-EN, Laion-ZH, COYO, GRIT: Used for image captioning tasks. Objects365, All-Seeing \cite{wang2023allseeing}: Used for detection tasks. Wukong-OCR, LaionCOCO-OCR, Common Crawl PDFs: Used for OCR tasks. TextVQA, ChartQA, DocVQA: Used for text reading and document understanding tasks. \\
    \hline
    \textbf{MiniGPT-v2} \cite{chen2023minigptv2} & P+F   & Text, Image\newline{} & LAION: A large-scale image-text paired dataset. CC3M: A dataset used for image captioning. SBU: Another dataset for image captioning. GRIT-20M \cite{peng2023kosmos2}: A dataset used for referring expression comprehension (REC) and referring expression generation (REG). COCO caption: A common dataset for image captioning tasks. Text Captions: A dataset for image captioning. RefCOCO, RefCOCO+, RefCOCOg: Datasets used for referring expression comprehension tasks. GQA, VQA-v2, OCR-VQA, OK-VQA, AOK-VQA: Datasets used for visual question answering (VQA) tasks. Flickr30k: A dataset used to improve the model's ability to generate contextual image descriptions. Unnatural Instructions: A dataset used to help restore the dialogue capabilities of language models. \\
    \hline
    \textbf{InternLM-XComposer2} \cite{dong2024InterLM-XComposer2} & P+F   & Text, Image & General Semantic Alignment: The MLLM utilize datasets such as ShareGPT4V-PT \cite{ShareGPT4V}, COCO, Nocaps, and TextCaps to endow the model with the basic ability to understand image content.\newline{}World Knowledge Alignment: It utilizes the Concept Data \cite{InternLM-XComposer}  to align world knowledge in images with the knowledge already acquired by the LLM.\newline{}Vision Capability Enhancement: Includes abilities such as optical character recognition (OCR), object grounding, and understanding structured images (e.g., charts), using datasets like WanJuan, Flickr, and MMC-Instruction.\\
    \hline
    \textbf{VisCPM-Chat} \cite{VisCPM} & P+F   & Text, Image & COCO, Visual Genome, CC3M, CC12M, Laion2B, Laion-COCO, Wukong, LLaVA-Instruct-150K, UniMM-Chat, and M3IT datasets\\
    \hline
    \textbf{Muffin} \cite{muffin} & P+F   & Text, Image & Visual Genome, COCO, CC3M, CC12M, and LAION-COCO within 180M image-text pairs \\
    \hline
    \textbf{ASM v2} \cite{wang2024asmv2} & P+F   & Text, Image & MiniGPT-4 \cite{Minigpt-4}, LLaVA-150k \cite{llava}, COCO subtitle dataset\\
    \midrule
    \multicolumn{4}{c}{\textbf{2 Spatial Comprehension \S \ref{subsec:Spatial Comprehension}}} \\
    \midrule
    \textbf{Shikra} \cite{shikra} & IT    & Referential Dialogue, Image & LLaVA-Pretraining: A dataset used for model pretraining. Flickr30K Entities: A dataset used for Spotting Captioning tasks. Visual Genome: A dataset used for Grounding Caption tasks. RefCOCO, RefCOCO+, RefCOCOg: Datasets used for Referring Expression Generation (REG) and Referring Expression Comprehension (REC) tasks. VQAv2: A dataset used for Visual Question Answering (VQA) tasks. PointQA-Local/Twice, Visual-7W: Datasets used for PointQA tasks \\
    \hline
    \textbf{GPT4ROI} \cite{GPT4ROI} & P+F   & Text, Image & COCO, RefCOCO, and RefCOCO+. GPT4ROI also utilize the Visual Genome, Flickr30K entities, and VCR datasets, as well as the LLaVA150K dataset by generating bounding boxes using pre-trained detectors. \\
    \hline
    \textbf{LION} \cite{chen2024lion}  & IT    & Text, Image & Vicuna-Instructions: A dataset containing 80 questions covering 9 different task categories, widely used for evaluating the capabilities of large language models (LLMs). The LION model utilizes this dataset for open-ended generation tasks.\newline{}AGIEval: A well-known benchmark that quantifies the reasoning ability of foundational models in human-centered standardized exams, including college entrance exams, math competitions, and bar exams. The paper selects all English multiple-choice questions from AGIEval (8 tasks, 2546 samples) for experiments.\newline{}BIG-Bench Hard (BBH): Comprising a series of challenging tasks designed to assess the capabilities and limitations of large language models. These tasks are ones where previous language models have performed poorly compared to average human evaluators. The paper selects all tasks that could be formatted as multiple-choice questions (23 tasks, 5511 samples). \\
    \hline
    \textbf{NExT-Chat} \cite{zhang2023NExT-Chat} & P+F   & Text, Image\newline{} & Pretraining Phase: Utilized datasets such as Flickr30K Entities, Visual Genome, RefCOCO, RefCOCO+, RefCOCOg, VQAv2, PointQA, Visual7W, VCR, etc.\newline{}Instruction Fine-Tuning Phase: It utilizes datasets such as VQAv2, RefCOCO, Flickr30K, LLaVA-instruct \cite{llava}, VG grounded captioning, VCR, and Shikra-RD \cite{shikra}.\newline{}Segmentation Training Phase: Utilized the segmentation parts of the RefCOCO, RefCOCO+, and RefCOCOg datasets.\newline{}\newline{}Flickr30K Entities: This dataset collects Flickr images and their corresponding descriptions for training models. Visual Genome: Provides a rich collection of image and description pairs for multimodal tasks. RefCOCO, RefCOCO+, RefCOCOg: These datasets focus on visual grounding and region description tasks. VQAv2: A dataset used for visual question answering tasks. PointQA: Contains images and related questions, requiring the model to locate the area where the answer lies. Visual7W: Contains images and "seven W" questions (i.e., questions involving elements such as who, what, where, when, why, how, and which). VCR: A dataset used for visual commonsense reasoning. \\
    \hline
    \textbf{PVIT} \cite{chen2023pvit}  & P+F   & Text, Image\newline{} & Visual Genome: Used for collecting region descriptions. Grounded COCO and Visual Genome datasets, using bounding boxes extracted by MDETR. SBU, using bounding boxes obtained by GLIP. \\
    \hline
    \textbf{VisionLLM} \cite{visionllm} & R     & Text, Image  & COCO2017: This is a widely used dataset for image recognition, detection, and segmentation, containing objects from 80 categories. It is large in volume and commonly used in computer vision tasks. RefCOCO, RefCOCO+, RefCOCOg: These datasets focus on visual grounding tasks, containing rich textual descriptions and corresponding image regions. They collectively contain over 120k referred objects. COCO Caption: This is an image captioning dataset, containing images and their corresponding descriptive texts. LLaVA-Instruct-150K: This is a dataset for instruction fine-tuning, containing 150k image-text pairs. \\
    \midrule
    \multicolumn{4}{c}{\textbf{3 Complex Modalities  \S \ref{subsec:Complex Modalities}}} \\
    \midrule
    \textbf{PointCLIPV2} \cite{zhu2023pointclip}  & P     & Text, Point Cloud Image & ModelNet40 and ModelNet10: These two datasets are commonly used for 3D object classification tasks. They contain multiple categories of 3D models and are widely used in 3D computer vision research.\newline{}ScanObjectNN: This dataset is used for zero-shot 3D classification tasks and contains 3D objects from real-world scenes.\newline{}ShapeNetPart: Used for 3D part segmentation tasks, this dataset includes various categories of 3D models and their annotated parts.\newline{}ScanNet V2: Used for 3D object detection tasks, this dataset contains 3D reconstructions and annotations of indoor scenes. \\
    \hline
    \textbf{RemoteSensing ChatGPT} \cite{guo2024RemoteSensingChatGPT} & R     & Text, Remote Sensing Image & AID: Used for scene classification tasks. LoveDA: Used for land use classification tasks. DOTA: Used for object detection tasks. BLIP Dataset: Used for image captioning tasks. \\
    \hline
    \textbf{VELMA} \cite{schumann2024velma} & IT    & Text, Navigation Image & Touchdown: contains approximately 10,000 navigation instances. These instances are based on Google's Street View and cover the dense urban street network of downtown Manhattan. The navigation instructions in the dataset were written by annotators while following the routes in Street View.\newline{}Map2seq: also contains approximately 10,000 navigation instances. Unlike Touchdown, the navigation instructions in Map2seq were written by annotators while viewing route maps and later verified as correct in Street View. \\
    \hline
    \textbf{RT-2} \cite{brohan2023rt2}  & P+F   & Text, Image, Robots Actions & WebLI Dataset: This is a large-scale visual-language dataset containing approximately 10 billion image-text pairs, covering 109 languages. These pairs were filtered to ensure that the top 10\% of examples with high cross-modal similarity scores were used for training.\newline{}Robotic Dataset: Based on the dataset by \citet{brohan2022rt1}, it contains demonstration clips collected using mobile manipulator robots. Each demonstration is accompanied by natural language instructions describing the task being performed. \\
    \midrule
    \multicolumn{4}{c}{\textbf{4 Any-to-any \S \ref{sec:Any Modalities} }} \\
    \midrule
    \textbf{ImageBind} \cite{Imagebind} & P     & Images, text, audio, depth, thermal, and IMU data & Audioset (AS): Used for training and evaluation. Contains 10-second videos from YouTube, categorized into 527 classes. The dataset includes approximately 20K balanced segments, 18K test segments, and about 2M unbalanced training segments.\newline{}ESC-50: Used for zero-shot evaluation. Contains 2000 5-second audio clips, divided into 50 categories. The dataset has predefined 5-fold evaluations.\newline{}Clotho: Contains audio and text descriptions from the Freesound platform. Includes 2893 audio clips in the development set and 1045 in the test set, each clip having 5 descriptions.\newline{}AudioCaps: Contains audio-video clips and text descriptions from YouTube. Uses clips from the Audioset dataset.\newline{}VGGSound: Contains about 200K 10-second video clips, annotated with 309 sound categories.\newline{}SUN RGB-D: Uses registered RGB and depth images for model training.\newline{}SUN Depth-only: Uses depth images from the SUN RGB-D dataset for evaluation.\newline{}NYU-v2 Depth-only (NYU-D): Uses depth images from the NYU-v2 dataset for evaluation.\newline{}LLVIP: Contains pairs of RGB images and thermal (infrared low-light) images. Collected in outdoor environments, it includes street-view RGB and infrared images.\newline{}Ego4D: Used for scene classification tasks. Contains 9645 videos of 108 unique scenes. \\
    \hline
    \textbf{PandaGPT} \cite{PandaGPT} & IT    & Text, Image, Video, Audio, depth, thermal, and IMU & 160k image-language instruction-following data released by Llava \cite{llava} and MiniGPT4 \cite{Minigpt-4} \\
    \midrule
    \multicolumn{4}{c}{\textbf{5 Domain-Specific \S \ref{sec:Domain specific}}} \\
    \midrule
    \textbf{Llava-Med} \cite{llavamed} & P+F   & CXR, CT, MRI, histopathology, and gross pathology & PubMed Central (PMC): Used for obtaining captions of medical images, for fine-tuning medical visual instruction sets.\newline{}MedMNIST: Used for image classification tasks, containing various medical image modalities.\newline{}MIMIC-CXR: Used for radiology report generation and summarization tasks, containing chest X-ray images and associated clinical texts.\newline{}CBIS-DDSM: Used for breast lesion classification tasks, containing mammography images.\newline{}CheXpert: Used for classification tasks of chest X-ray images. \\
    \hline
    \textbf{CancerGPT} \cite{CancerGPT} & R     & Text, BioMedicine Image & DrugComb Portal: This is a publicly accessible database containing results from drug combination screening studies on various cancer cell lines. The database includes drug sensitivity data and drug pair combination data. In the paper, the Loewe synergy scores from this database were used to evaluate the synergistic effects of drug combinations. The database contains 4,226 unique drugs, 288 cell lines, and a total of 718,002 records of drug pair synergy. \\
    \hline
    \textbf{BiomedGPT} \cite{zhang2023biomedgpt} & R & Text, BioMedicine Image & MedMNIST: This dataset is for image classification tasks and includes various medical image modalities and views. CBIS-DDSM: This dataset is for breast lesion classification tasks, including mass and calcification classification. VQA-RAD, SLAKE, PathVQA: These datasets are for visual question answering tasks and cover radiology and pathology data. MIMIC-CXR, MIMIC-III: These datasets are for text understanding and generation tasks, containing clinical notes and reports. PubMed Abstracts, NCBI BioNLP: These datasets are for pre-training language models and provide a large amount of text data. \\
    \bottomrule
    \end{tabular}%
}
\label{tab:ExistedDataset}
\end{table*}

\begin{enumerate}
\item {\bf ShareGPT4V \cite{ShareGPT4V}} 
\label{App:ShareGPT4V}
    \item[] \textit{\textbf{Collection}}: 
        This dataset collects approximately 100,000 images from multiple sources, ensuring diversity. Using GPT-4 Vision and specific prompts, it generates detailed captions for each image. The quality of the generated captions is verified by replacing the captions of existing models and reassessing performance.
    \item[] \textit{\textbf{Description}}: 
        This dataset has the following characteristics:
        \begin{itemize}
            \item  Rich Descriptiveness: Each caption averages 942 characters, covering a wide range of information such as world knowledge, object attributes, spatial relationships, and aesthetic evaluations.
            \item  High-Quality Captions: Compared to the simple captions commonly found in mainstream image-text datasets, the captions generated by ShareGPT4V are more detailed and accurate. The images in the dataset come from various sources, including detection and segmentation images, images containing complex text, as well as images from the web featuring artworks, landmarks, celebrities, and more.
        \end{itemize}

\item {\bf ALLaVA \cite{chen2024allava}}
\label{App:Allava}
    \item[] \textit{\textbf{Collection}}: This dataset uses the capabilities of GPT-4V to generate captions, questions, and answers. The images come from LAION and Vision-FLAN.
    \item[] \textit{\textbf{Description}}: The ALLaVA dataset uses the capabilities of GPT-4V to generate detailed captions, complex instructions, and detailed answers, creating a high-quality synthetic dataset. This dataset contains 1.4 million data points, including fine captions, complex instructions, and detailed answers generated by GPT-4V.

\item {\bf Evol-Intruct-GPT4-Turbo-143K \cite{chen2024allava} }
\label{App:Evol-Intruct-GPT4-Turbo-143K}
    \item[] \textit{\textbf{Collection}}: This dataset selects WizardLM\_evol\_instruct\_V2 as the question set and uses GPT-4 Turbo to regenerate the answers.
    \item[] \textit{\textbf{Description}}: The captions and answers in Evol-Instruct-GPT4-Turbo-143K, generated by GPT-4V, are of high quality, providing detailed visual descriptions and complex instructions. The images come from diverse sources, including natural images scraped from the internet and datasets integrated from multiple VQA tasks, ensuring data diversity. The generated questions and answers involve complex reasoning, helping to enhance the model ability to handle complex visual tasks.

\item {\bf Mini-Gemini \cite{li2024minigemini}}
\label{App:Mini-Gemini}
    \item[] \textit{\textbf{Collection}}: This dataset collects and creates data from various public sources, including high-quality responses, task-oriented instructions, and generation-related data.
    \item[] \textit{\textbf{Description}}: This dataset aims to enhance model performance through precise image understanding, reasoning-based generation, and the extension of the operational scope of current Vision-Language Models (VLMs).

\item {\bf LLaVA-Instruct-158K \cite{llava}}
\label{App:LLaVA-Instruct-158K}
    \item[] \textit{\textbf{Collection}}: This dataset uses GPT-4 to generate image-based questions and answers. This process includes three components: Conversation, Detailed Description, and Complex Reasoning.
    \item[] \textit{\textbf{Description}}: 
    \begin{itemize}
        \item Diversity: This dataset contains various types of questions and answers, covering conversation, detailed description, and complex reasoning. This helps the model learn to handle multiple types of language instructions.
        \item Challenge: This dataset aims to challenge the model limits, including diverse images such as indoor and outdoor scenes, memes, paintings, sketches, and associated detailed annotations and questions.
        \item Alignment: The dataset construction focuses on the alignment between images and text, ensuring the model can understand and respond to visual content.
    \end{itemize}

\item {\bf DetGPT Dataset \cite{pi2023detgpt}}
\label{App: DetGPT Dataset}
    \item[] \textit{\textbf{Collection}}: This dataset utilizes existing image captions and object detection datasets, combining two types of text annotations: image captions and object categories within the images. Using system prompts, ChatGPT generates detailed image descriptions and query-answer pairs for target objects. Through the generated descriptions and query-answer pairs, annotations are reorganized to associate each image with its corresponding instruction-answer pair.
    \item[] \textit{\textbf{Description}}: Using ChatGPT to generate image-related queries and answers reduces the workload of manual annotation. This dataset supports open vocabulary object detection, allowing the model to identify and locate a broader range of object categories during inference.

\item {\bf Large-scale Robust Visual (LRV)-Instruction \cite{GAVIE}}
\label{App: Large-scale Robust Visual (LRV)-Instruction}
    \item[] \textit{\textbf{Collection}}: This dataset is generated based on GPT-4 and the Visual Genome dataset.
    \item[] \textit{\textbf{Description}}: Unlike existing research that mainly focuses on positive instruction samples, LRV-Instruction is designed to include both positive and negative instructions to achieve more robust visual instruction alignment. Negative instructions are designed at three semantic levels: (i) non-existent object operations, (ii) existent object operations, and (iii) knowledge operations. The instructions in the dataset include both statements and questions, adding to the diversity of the dataset and allowing the model to better understand and respond to different types of instructions. By fine-tuning on the LRV-Instruction dataset, the model can reduce instances of hallucination and improve performance across multiple public datasets, showing improvements over existing methods.

\item {\bf Osprey-724K \cite{yuan2024osprey}}
\label{App: Osprey-724K}
    \item[] \textit{\textbf{Collection}}: This dataset is generated using meticulously designed prompt templates based on publicly available datasets such as COCO, RefCOCO, RefCOCO+, RefCOCOg, PACO-LVIS, and others. The language model GPT-4 is used to generate high-quality mask-text pairs to produce data that follows instructions. The robustness and flexibility of the dataset are enhanced by designing negative samples and short-form responses.
    \item[] \textit{\textbf{Description}}: This dataset contains 724,000 samples, each including a fine mask region and the corresponding text description. The precise alignment of mask regions and text descriptions promotes a deeper understanding of specific areas within images and fine-grained pixel-level alignment by the model.
    \begin{itemize}
        \item Negative Sample Mining: A method for mining negative samples is introduced to enhance the model ability to recognize different categories.
        \item Short-Form Response Instructions: To improve the model response flexibility, the dataset also includes short-form response instructions.
    \end{itemize}

\item {\bf MagnifierBench \cite{OtterHD}}
\label{App: MagnifierBench}
    \item[] \textit{\textbf{Collection}}: The images originate from the Panoptic Scene Graph Generation (PVSG) dataset, which contains video data of numerous complex scenes. The annotation team is first instructed to carefully examine the videos and identify unique complex frames containing numerous small objects. A small square, equivalent to 1\% of the image size, is placed in each video to help annotators judge the scale of the small objects. Once suitable frames are identified, the next task for annotators is to develop question-answer pairs for those tiny objects. In the subsequent annotation phase, the author team carefully reviews each question-answer entry in the dataset, excluding those involving overly large objects or questions that can be easily answered using common sense knowledge.
    \item[] \textit{\textbf{Description}}: 
    \begin{itemize}
        \item High-Resolution Image Understanding: MagnifierBench is designed to evaluate the capability of large multimodal models (LMMs) in recognizing tiny details and spatial relationships in high-resolution images.
        \item Complex Scenes: The images in the dataset depict complex scenes densely packed with small objects, primarily sourced from first-person videos of household activities.
        \item Fine Annotation: During the dataset compilation, annotators are required to carefully zoom in and focus on these small objects, which occupy approximately 1
        \item Multiple Choice and Free Response: The dataset includes two types of question-answer pairs: multiple choice and free response. This requires the model to handle different types of questions and generate accurate answers.
    \end{itemize}

\item {\bf High-Quality Bilingual Dataset. \cite{chen2024InternVL15}}
\label{App: High-Quality Bilingual Dataset}
    \item[] \textit{\textbf{Collection}}: A data translation pipeline is developed using open-source large language models (LLMs) to translate English datasets into other languages (e.g., Chinese) while maintaining the naturalness and consistency of labels. This pipeline is used to collect bilingual datasets.
    \item[] \textit{\textbf{Description}}: This dataset covers common scenes and document images, annotated with question-answer pairs in both English and Chinese. This bilingual feature significantly enhances the performance in OCR (Optical Character Recognition) and Chinese-related tasks.

\item {\bf CCS400K (Common Crawl Screenshot 400K) \cite{cogagent}}
\label{App: CCS400K (Common Crawl Screenshot 400K)}
    \item[] \textit{\textbf{Collection}}: By extracting URLs from the latest Common Crawl data, web page screenshots are captured. The Playwright tool is used to capture all visible DOM elements and their corresponding rendering boxes.
    \item[] \textit{\textbf{Description}}: This dataset comprises 400,000 web page screenshots, including all visible DOM elements and their corresponding rendering boxes. It provides 140 million REC (Referring Expression Comprehension) and REG (Referring Expression Generation) question-answer pairs, aimed at enhancing the understanding of GUI elements.

\item {\bf Free-form Text-Image Composition  \cite{dong2024InterLM-XComposer2}}
\label{App: Free-form Text-Image Composition}
    \item[] \textit{\textbf{Collection}}: This dataset is collected from internal resources, encompassing various writing styles, text editing examples, adherence to complex instructions, and the use of diverse materials for personalized content creation.
    \item[] \textit{\textbf{Description}}: This dataset focuses on free-form text-image combinations, aimed at generating intertwined text and image content consistent with user-specified text requirements. It includes various writing styles, flexible text editing, adherence to complex instructions, and the use of materials for personalized content creation.

\item {\bf Noisy Instruction-following Data \cite{LLaVAR}}
\label{App: Noisy Instruction-following Data}
    \item[] \textit{\textbf{Collection}}: This dataset uses publicly available OCR tools (PaddleOCR) to collect 422,000 text-rich images (e.g., movie posters, book covers) from the LAION dataset.
    \item[] \textit{\textbf{Description}}: This study collects 422,000 text-rich images from the LAION dataset and processes them using OCR tools. Due to the performance limitations of OCR technology when handling diverse fonts and colorful backgrounds, the collected data contains noise, meaning OCR results may include missing words and grammatical errors. To ensure diversity in instructions, the directive "recognize all visible text in the image" is rewritten in ten different ways. Each image is paired with a randomly sampled input instruction, and the recognized text is used as the desired output response, creating single-turn instruction-response pairs in a conversational format.

\item {\bf GPT-4-based Instruction-following Data \cite{LLaVAR}}
\label{App: GPT-4-based Instruction-following Data}
    \item[] \textit{\textbf{Collection}}: Using the OCR-recognized text and image captions, these are used to prompt GPT-4, generating 16,000 conversations as high-quality instruction-following examples.
    \item[] \textit{\textbf{Description}}: Compared to Noisy Instruction-following Data, these data are generated using the GPT-4 model to provide more organized sentences and more accurate question-answer pairs. The GPT-4 model is prompted with OCR results and image captions to generate instruction-following data, requiring GPT-4 to denoise the OCR results and develop specific questions based on the inputs to create complex instructions. The generated data contain multiple question-answer pairs, forming multi-turn conversations that mimic human interactions. The questions and answers generated by GPT-4 go beyond simple text recognition, including instructions that require reasoning, writing, and elaboration skills.
    This study selects images from specific clusters containing sufficiently visible and coherent sentences to generate higher-quality instruction-following data. To prevent image captions from overly focusing on textual content, OCR bounding boxes are used to mask the text, and image inpainting techniques fill the masked areas to generate captions that do not rely on the image text.

\item {\bf Wukong \cite{VisCPM}}
\label{App: Wukong}
    \item[] \textit{\textbf{Collection}}: The data are collected from images and text through specific filtering criteria, including image size, text language, length, and frequency.
    \item[] \textit{\textbf{Description}}: This dataset comprises 100 million image-text pairs in Chinese. The images and text data are filtered based on image size, language, length, and frequency. This dataset is specifically tailored for Chinese content, providing support for Chinese image-text tasks.

\item {\bf Laion-COCO \cite{VisCPM}}
\label{App: Laion-COCO}
    \item[] \textit{\textbf{Collection}}: This dataset is generated using the BLIP tool from the Laion-2B dataset, mimicking the MSC COCO style.
    \item[] \textit{\textbf{Description}}: The dataset is substantial, containing 600 million entries, and provides detailed descriptions of the images.

\item {\bf Zero \cite{VisCPM}}
\label{App: Zero}
    \item[] \textit{\textbf{Collection}}: The dataset is filtered from a pool of 5 billion image-text pairs based on user click-through rates.
    \item[] \textit{\textbf{Description}}: These image-text pairs are filtered from a pool of 5 billion pairs. The Zero dataset is distinguished by its user behavior-based filtering mechanism, which may help the model learn user preferences.

\item {\bf UniMM-Chat \cite{muffin}}
\label{App: UniMM-Chat}
    \item[] \textit{\textbf{Collection}}: 
    Using images from the COCO dataset as a foundation, annotations from five commonly used vision-language (VL) datasets—VQAv2, OKVQA, AOKVQA, Visual Dialog, and COCO Caption—are combined. ChatGPT is then used to generate multi-turn conversations based on the merged annotations, providing richer contextual information for the images.
    \item[] \textit{\textbf{Description}}: UniMM-Chat merges information describing the same image from different datasets, transforming it into knowledge-rich dialogue data. The dataset contains dialogues that average 9.89 turns each, with each dialogue linked to a unique image. The dataset is diverse and can be quantified by analyzing question types and their direct nouns or verbs using the Berkeley Neural Parser tool.

\item {\bf AS-1B \cite{wang2024asmv2}}
\label{App: AS-1B}
    \item[] \textit{\textbf{Collection}}: Created through a scalable semi-automated data engine, this dataset combines human feedback with efficient model loops. The data engine iteratively improves data quality in a "data-human-model" cycle.
    \item[] \textit{\textbf{Description}}: The AS-1B dataset contains annotations for over 1 billion regions, including semantic labels, question-answer pairs, and detailed descriptions. It covers 3.5 million common and rare real-world concepts, ranging from general categories (e.g., human, backpack) to fine-grained or rare categories with attributes (e.g., old metal latch). Specifically, AS-1B includes 132.2 billion tagged descriptions of concepts and their attributes, providing extensive textual information for images and regions. 
    This dataset offers broad coverage of open-world semantics, aiding models in learning to generalize and understand various scenes and objects encountered in the open world. It supports various vision and vision-language tasks, including region text retrieval, region recognition, caption generation, and question answering. Unlike datasets that mainly focus on holistic image understanding, AS-1B focuses on the recognition and understanding of individual instances within scenes, aligning more closely with the characteristics of the human visual system.
    In constructing the dataset, a "data-human-model" cycle was employed by the data engine to iteratively improve data quality. Various models, including large language models (LLMs), detection models, captioning models, and visual question answering (VQA) models, were used as "annotators" to add semantic annotations to dense region proposals generated by state-of-the-art object detectors.

\item {\bf Shikra-RD \cite{shikra}}
\label{App: Shikra-RD}
    \item[] \textit{\textbf{Collection}}: This dataset uses images and corresponding descriptions from the Flickr30K Entities dataset. Each image has five descriptions, and the objects appearing in the images are marked with bounding boxes. Researchers explain the format of the bounding boxes to GPT-4 and require GPT-4 to understand the images based on these five sentences and the bounding boxes, then design question-answer pairs.
    \item[] \textit{\textbf{Description}}: The Shikra-RD dataset is specifically created for training the Shikra model to enhance its performance in Referential Dialogue (RD) tasks. This dataset is generated by using the Flickr30K Entities dataset in combination with GPT-4 and includes naturally occurring communication data with location annotations.

\item {\bf GRIT (Grounded Image-Text pairs) \cite{peng2023kosmos2}}
\label{App: GRIT (Grounded Image-Text pairs)}
    \item[] \textit{\textbf{Collection}}: This dataset is constructed through a pipeline that includes two main steps: generating noun phrase-bounding box pairs and creating referential expression-bounding box pairs. First, the spaCy library is used to parse noun phrases from the titles, then a pre-trained localization model (such as GLIP) is employed to obtain bounding boxes related to the noun phrases. Next, dependency relationships are used to expand the noun phrases into referential expressions, retaining only those expressions or noun phrases that are not included in other expressions. Finally, the bounding boxes of the noun phrases are assigned to the corresponding generated referential expressions.
    \item[] \textit{\textbf{Description}}: GRIT is a large-scale web-based dataset composed of image-text pairs. These image-text pairs link noun phrases and referential expressions in the text to objects or regions in the images. This data format acts as a "hyperlink," connecting objects or regions in the images with the captions.

\item {\bf Grounding-anything Dataset (GranD) \cite{GLaMM}}
\label{App: Grounding-anything Dataset (GranD)}
    \item[] \textit{\textbf{Collection}}: The GranD dataset is generated through an automated multi-level annotation process, which includes object localization and attributes (Level-1), relationships and landmarks (Level-2), scene graphs and dense captions (Level-3), and additional contextual insights (Level-4).
    \item[] \textit{\textbf{Description}}: GranD is a large-scale, densely annotated dataset designed to support pixel-level object localization and rich language descriptions. It is generated through an automated annotation process aimed at enhancing the capabilities of Large Multimodal Models (LMMs) in visual instruction tuning. The GranD dataset includes detailed contextual information, ranging from fine-grained details to high-level backgrounds, such as historical information and landmark details.

\item {\bf ViP-Bench \cite{cai2024vipllava}}
\label{App: ViP-Bench}
    \item[] \textit{\textbf{Collection}}: These images and question pairs are collected from datasets such as MM-Vet, MMBench, and Visual Genome. Each image is paired with a question designed to test the ability to understand and interpret at the region level.
    \item[] \textit{\textbf{Description}}: ViP-Bench is a comprehensive benchmark suite designed to evaluate the ability of multimodal models to understand arbitrary visual prompts. It covers key aspects of visual understanding, including recognition, optical character recognition (OCR), knowledge, mathematics, object relationship reasoning, and language generation.

\item {\bf ReasonDet \cite{wei2023lenna}}
\label{App: ReasonDet}
    \item[] \textit{\textbf{Collection}}: Derived from processing the ReasonSeg dataset.
    \item[] \textit{\textbf{Description}}: The ReasonDet dataset is specifically constructed to evaluate models' performance on inference-based detection tasks. It is derived from processing the ReasonSeg dataset and is used to measure models' reasoning detection capabilities.

\item {\bf Visual Grounding Dataset \cite{CogVLM}}
\label{App: Visual Grounding Dataset}
    \item[] \textit{\textbf{Collection}}: This dataset is created by sampling from the LAION-115M dataset, ensuring that each image has at least two bounding boxes. Nouns are extracted using spaCy, and bounding boxes are predicted using GLIPv2.
    \item[] \textit{\textbf{Description}}: In this dataset, each noun in the image description is associated with a bounding box to indicate its position in the image. This dataset is sampled from the LAION-115M dataset, a subset of LAION-400M. It ensures that over 75\% of the images contain at least two bounding boxes.

\item {\bf InDET \cite{dang2023instructdet}}
\label{App: InDET}
    \item[] \textit{\textbf{Collection}}: Two pipelines (global prompts and local prompts) are used to generate detection expressions from the base model. CLIP is employed for visual-text matching to filter out unsuitable expressions. Semantic clustering is used to discover common attributes among multiple objects, and an LLM generates a summary of commonalities for each combination. The final instruction set is produced through post-processing, which includes removing duplicate expressions and using LLaMA for synonym replacement.
    \item[] \textit{\textbf{Description}}: By using existing vision-language models (VLM) and large language models (LLM), such as LLaVA and LLaMA, instructions guided by text prompts and object bounding boxes (bbxs) are generated to achieve broad coverage of user intentions. These instructions are designed to be more human-like, capable of describing the attributes, categories, and relationships of objects.

\item {\bf Query-answer instruction dataset \cite{pi2023detgpt}}
\label{App: Query-answer instruction dataset}
    \item[] \textit{\textbf{Collection}}: The dataset is constructed by utilizing existing image description and object detection datasets, combined with two types of text annotations: (1) image descriptions, providing visual content descriptions from different perspectives; and (2) object categories appearing in the images. Based on the given descriptions and objects, query-answer prompts are designed to guide ChatGPT in generating more detailed scene descriptions and query-answer pairs.
    \item[] \textit{\textbf{Description}}: The generated query-answer pairs in the dataset are goal-oriented, meaning that the user input reflects the goal they wish to achieve, and the dataset is responsible for identifying objects in the images that help achieve these goals. The dataset includes diverse query types, such as abstract queries based on user instructions and queries that require the model to reason about which objects in the image meet the user objectives. The generated query-answer pairs rely on common-sense knowledge, enabling the model to reason based on information stored in large language models. Each image is accompanied by a detailed description, and multiple instruction-answer pairs are generated based on these descriptions and the objects in the images. Additionally, to ensure consistent model output, the query-answer pairs in the dataset follow a specific format, such as ending the answers with "Therefore the answer is: [object\_names]."

\item {\bf Language-Assisted Multimodal instruction tuning dataset \cite{lamm}}
\label{App: Language-Assisted Multi-Modal instruction tuning dataset}
    \item[] \textit{\textbf{Collection}}: Images and point clouds were collected from publicly available datasets, and instructions and responses were generated based on the original labels of these datasets using the GPT-API through a self-instruction method.
    \item[] \textit{\textbf{Description}}: This dataset encompasses a wide range of 2D and 3D vision tasks, emphasizing fine-grained information and factual knowledge. It includes 186,098 image-language instruction-response pairs and 10,262 point cloud-language instruction-response pairs. Images and point clouds were collected from publicly available datasets, and instructions and responses were generated based on the original labels of these datasets using the GPT-API through a self-instruction method. The dataset construction process involves multiple rounds of everyday conversation, multiple rounds of factual knowledge dialogue, single-round detailed descriptions, and single-round visual task dialogues.

\item {\bf Point-Text Instruction Following Data \cite{pointllm}}
\label{App: Point-Text Instruction Following Data}
    \item[] \textit{\textbf{Collection}}: The Cap3D dataset and GPT-4 model were used. Cap3D is a large-scale 3D object description dataset built on Objaverse. Leveraging GPT-4 reasoning capabilities and world model, the authors prompted GPT-4 to generate diverse instruction-following data based on the context provided by the descriptions.
    \item[] \textit{\textbf{Description}}: The dataset includes point clouds and textual instructions to train and guide models in understanding 3D object point clouds and generating corresponding responses based on user instructions. It supports a two-stage training method: first, a feature alignment stage, aligning point cloud features with text feature space; second, an instruction tuning stage, using complex instructions to enhance the understanding and response capabilities to user commands. The dataset comprises brief descriptive instructions and complex instructions, aiding the model in learning to comprehend objects from different perspectives and accurately responding to diverse human commands. Quality is ensured through manual filtering and validation, improving the effectiveness of model training. The instructions cover various aspects of 3D objects, including type, appearance, and functionality, enhancing the comprehensive understanding of 3D structures.

\item {\bf RSICap \cite{hu2023rsgpt}}
\label{App: RSICap}
    \item[] \textit{\textbf{Collection}}: The dataset is constructed based on the DOTA (Detection of Transportation Aircraft) object detection dataset. The DOTA dataset is chosen for its rich image diversity, including images from various satellite and aerial sensors, as well as different resolutions and image types. The images are divided into 512×512 pixel blocks, and 2,585 blocks are randomly selected for annotation. Five remote sensing experts annotated the images, following principles of describing image attributes, object attributes, and scenes.
    \item[] \textit{\textbf{Description}}: This is a high-quality remote sensing image captioning dataset containing 2,585 manually annotated captions, providing rich and detailed information. These captions offer comprehensive descriptions for each image, including scene descriptions (such as residential areas, airports, or farmland) and object information (such as color, shape, quantity, and absolute location).

\item {\bf HqDC-1.4M, RSSA, HqDC-Instruct and RS-Specialized-Instruct \cite{pang2024h2rsvlm}}
\label{App: H2RSVLM}
    \item[] \textit{\textbf{Collection}}: Images from publicly available remote sensing datasets are used. Effective prompts are designed to leverage Google commercial Gemini-Vision model to generate captions for remote sensing images. Carefully crafted prompts and a few examples guide the Gemini language model to generate multi-turn dialogue and reasoning data.
    \item[] \textit{\textbf{Description}}: 
    \begin{itemize}
        \item HqDC-1.4M: This is a large-scale, high-quality remote sensing image captioning dataset containing 1.4 million image-caption pairs. These captions, generated by high-performance multimodal language models, include rich information such as image attributes and object properties, covering type, scene, and object details. This significantly enhances the VLM understanding and recognition capabilities for remote sensing images. The scale and diversity of the dataset ensure the model broad applicability and excellent generalization ability.
        \item RSSA: This dataset aims to enhance model self-awareness in the field of remote sensing. It includes a series of answerable and unanswerable tasks, effectively increasing the honesty of Remote Sensing Vision-Language Models (RSVLMs) and reducing hallucinations.
        \item HqDC-Instruct and RS-Specialized-Instruct: These two datasets are used to enhance the multi-turn dialogue and complex reasoning capabilities of Remote Sensing Vision-Language Models (RSVLMs) while also teaching the model professional knowledge and skills in remote sensing image processing.
    \end{itemize}

\item {\bf LHRS-Align \cite{muhtar2024LHRS-Bot}}
\label{App: LHRS-Align}
    \item[] \textit{\textbf{Collection}}: The dataset construction process involves obtaining remote sensing images from Google Earth and collecting geographic features from the OSM database. Large language models (LLMs) are then used to generate image captions. The data cleaning process includes deduplication, pruning, and semantic balancing.
    \item[] \textit{\textbf{Description}}: This is a large-scale remote sensing image-text dataset, created by georeferencing remote sensing images with geographic vector features and using the rich attribute information from the OpenStreetMap (OSM) database to generate image captions.

\item {\bf LHRS-Instruct \cite{muhtar2024LHRS-Bot}}
\label{App: LHRS-Instruct}
    \item[] \textit{\textbf{Collection}}: The dataset is constructed by utilizing two public image captioning datasets and the LHRS-Align dataset. The public datasets undergo rigorous data cleaning, and GPT-4 is used to generate complex instruction data.
    \item[] \textit{\textbf{Description}}: This is a multimodal instruction-following dataset specifically designed for remote sensing image understanding. It includes not only instruction data for various remote sensing image understanding tasks but also complex visual reasoning data generated by GPT-4.

\item {\bf SkyEye-968k \cite{zhan2024skyeyegpt}}
\label{App: SkyEye-968k}
    \item[] \textit{\textbf{Collection}}: The dataset is created from the integration of existing public datasets and some event-based data generated from aerial videos (e.g., ERA-VQA). Data is manually verified and selected to ensure high quality.
    \item[] \textit{\textbf{Description}}: This dataset is specifically designed for training SkyEyeGPT, including single-task image-text instructions and multi-task dialogue instructions. It aims to improve the model performance on various specific tasks and enhance its multi-task dialogue capabilities.

\item {\bf M-Paper \cite{hu2023mPLUG-PaperOwl}}
\label{App: M-Paper}
    \item[] \textit{\textbf{Collection}}: LaTeX source files are collected from arXiv and PapersWithCode, and these files are parsed to extract charts and related text. The image format of the charts is extracted using PDF cropping tools such as GROBID, while LaTeX code is directly extracted from the source files. Additionally, summaries (outlines) are generated using GPT-3.5 to better align with user intent.
    \item[] \textit{\textbf{Description}}: The first dataset that supports joint understanding of multiple scientific charts, including charts and tables in image formats or LaTeX code. It is constructed by parsing LaTeX source files from high-quality papers, aligning charts with relevant paragraphs, and creating professional chart analysis samples for training and evaluation.

\item {\bf A comprehensive multimodal instruction dataset \cite{OneLLM}}
\label{App: A comprehensive multimodal instruction dataset}
    \item[] \textit{\textbf{Collection}}: The image-text pairs data includes LAION-400M and LAION-COCO. Training data for videos, audio, and point clouds come from WebVid-2.5M, WavCaps, and Cap3D, respectively. Due to the lack of large-scale depth/normal map-text data, depth/normal maps are generated using a pre-trained DPT model. IMU-text pairs use IMU sensor data from Ego4D and corresponding video narratives (i.e., text annotations). fMRI-text pairs use the subj01 imaging sessions from the NSD dataset, with image titles related to visual stimuli as text annotations.
    \item[] \textit{\textbf{Description}}: The dataset includes 2 million items, covering eight modalities: images, audio, video, point clouds, depth/normal maps, inertial measurement units (IMU), and functional magnetic resonance imaging (fMRI). It is constructed by collecting and organizing instruction-tuning (IT) datasets for different modalities, encompassing a range of tasks from detailed descriptions/reasoning, dialogues, and short question-answering to captions. During the dataset design, the authors carefully considered prompt design across different modalities and tasks to avoid conflicts and ensure that the model can effectively handle multimodal inputs.

\item {\bf AnyInstruct-108k \cite{zhan2024anygpt}}
\label{App: AnyInstruct-108k}
    \item[] \textit{\textbf{Collection}}: The dataset is synthesized using generative models, involving two main stages: generating text-based dialogues with multimodal elements and converting text to multimodal outputs. Text dialogues are generated using GPT-4, with non-textual modalities embedded within these dialogues. Advanced generative models, such as OpenAI DALL-E-3, MusicGen, and Microsoft Azure Text-to-Speech API, are used to convert text descriptions into multimodal elements.
    \item[] \textit{\textbf{Description}}: The dataset contains 108,000 multi-turn dialogue samples that intricately interweave various modalities, such as text, speech, images, and music. It is a large-scale any-to-any multimodal instruction dataset, enabling the model to handle arbitrary combinations of multimodal inputs and outputs.

\item {\bf Modality-switching Instruction Tuning (MosIT) \cite{NExT-GPT}}
\label{App: Modality-switching Instruction Tuning (MosIT)}
    \item[] \textit{\textbf{Collection}}: The authors design template dialogue examples, and based on these templates, GPT-4 generates additional dialogues across various scenarios, covering over 100 topics or keywords. The interactive content is diverse, including both direct and implicit requests made by the "human" role and tasks such as perception, reasoning, suggestions, and planning performed by the "machine" role. Each dialogue includes 3-7 turns (i.e., question-answer pairs), with "human"-"machine" interactions involving multiple modalities on either the input or output side and alternating between modalities.
    \item[] \textit{\textbf{Description}}: The dataset is manually curated and annotated for modality switching instruction tuning (MosIT), containing 5,000 samples. It covers a wide range of multimodal inputs and outputs, providing the necessary complexity and variability to train multimodal large language models (MM-LLMs) that can handle diverse user interactions and accurately deliver the desired responses.

\item {\bf MACAW-LLM Instruction Dataset \cite{Macaw-LLM}}
\label{App: MACAW-LLM Instruction Dataset}
    \item[] \textit{\textbf{Collection}}: Instruction-response pairs are generated using GPT-3.5-TURBO, based on COCO image captions and Charades and AVSD video captions.
    \begin{itemize}
        \item A dataset is created by randomly selecting 10,000 images and their captions from the COCO dataset.
        \item The dataset is further developed by integrating 9,848 video captions from the Charades and AVSD datasets.
    \end{itemize}
    \item[] \textit{\textbf{Description}}: 
    This is a large-scale multimodal instruction dataset that includes a variety of instruction tasks utilizing image and video modalities.
    \begin{itemize}
        \item The dataset covers multi-turn dialogues, including 69,000 image instances and 50,000 video instances.
        \item The dataset construction leverages the generative capabilities of current LLMs (such as GPT-3.5-TURBO) to ensure that the target text aligns correctly with human instructions.
        \item The dataset currently focuses primarily on single-turn dialogues, but the authors are actively incorporating multi-turn dialogues and diverse multimodal content to enhance its richness.
    \end{itemize}

\item {\bf MedC-K \cite{wu2024PMC-LLaMA}}
\label{App: MedC-K}
    \item[] \textit{\textbf{Collection}}: 
    The data is collected from below and cleaned to remove irrelevant elements such as URLs, author lists, and duplicate content:
    \begin{itemize}
        \item Academic Papers: Biomedical-related papers are selected from the S2ORC dataset based on their PubMed Central (PMC) IDs.  
        \item Textbooks: Collected from Open Library, university libraries, and renowned publishers, covering a wide range of medical specialties.  
    \end{itemize}

    \item[] \textit{\textbf{Description}}: MedC-K includes: 
    \begin{itemize}
        \item 4.8 million biomedical academic papers providing high-quality, cutting-edge medical knowledge. 
        \item 30,000 medical textbooks covering a broad range of medical specialties, offering fundamental and core medical knowledge.
    \end{itemize}
    The collected content undergoes thorough cleaning, including the removal of irrelevant elements such as URLs, author lists, and duplicate information to ensure data quality. By injecting text content from these papers and books into the language model, the goal is to enable the model to accumulate sufficient domain-specific medical knowledge and establish better embedding spaces for complex terminology. In the training batches, a 15:4:1 ratio is used to mix samples from books, papers, and general text corpora to emphasize the fundamental medical knowledge found in textbooks.

\item {\bf MedC-I \cite{wu2024PMC-LLaMA}}
\label{App: MedC-I}
    \item[] \textit{\textbf{Collection}}: 
    \begin{itemize}
        \item Medical Dialogue: Instructions are expanded based on data from Med-Alpaca and ChatDoctor to enhance the model robustness to diverse instructions.
        \item Medical Reasoning QA: Open-source medical multiple-choice question-answering datasets (such as USMLE, PubMedQA, and MedMCQA) are used, and reasoning outputs are generated with ChatGPT.
        \item Medical Knowledge Graph Prompts: QA pairs are constructed using UMLS to link medical terms with their knowledge descriptions or relationships.
    \end{itemize}
    \item[] \textit{\textbf{Description}}: MedC-I is a large-scale, high-quality, domain-specific instruction dataset designed for the instruction fine-tuning phase, containing 202 million tokens. The dataset consists of three parts:

    \begin{itemize}
        \item Medical Consultation Dialogues: Using patient-doctor dialogue data, patient questions are treated as instructions and doctor responses as ground truth, enhancing the model robustness to diverse instructions.
        \item Medical Reasoning QA: Utilizing open-source medical multiple-choice question datasets and causal analysis generated by ChatGPT to provide detailed reasoning guidance.
        \item Medical Knowledge Graph Prompts: Constructing QA pairs using medical knowledge graph resources such as UMLS, linking medical terms with their knowledge descriptions or corresponding relationships to align with clinicians' experience.
    \end{itemize}

\item {\bf ChiMed-VL \cite{liu2023Qilin-Med-VL}}
\label{App: ChiMed-VL}
    \item[] \textit{\textbf{Collection}}: The dataset is constructed by utilizing several open-source English medical multimodal datasets and translating these datasets into Chinese using GPT-3.5, followed by expert quality control.
    \item[] \textit{\textbf{Description}}: ChiMed-VL is a large-scale Chinese vision-language dataset designed for general healthcare. It contains over 1 million image-text pairs and is carefully curated to use various types of images to provide detailed and comprehensive explanations of medical data.
 
\end{enumerate}
\subsection{Summary of Evaluation Metrics}
\label{sec:Summary of Evaluation Metrics}
We summarize the evaluation metrics of MLLMs from the perspective of different multimodal tasks.
\begin{enumerate}
    \item \textbf{Metrics for Text-Image Tasks:}
    \begin{itemize}
        \item \textbf{BLEU, ROUGE, METEOR, CIDEr, SPICE:} Common metrics for image captioning tasks, used to evaluate the quality of generated textual descriptions.
        \item \textbf{Accuracy} and \textbf{VQA score:} Used in Visual Question Answering (VQA) tasks to assess the correctness of model answers.
        \item \textbf{Recall@K, MRR, Median Rank:} Evaluate retrieval performance for image-text alignment.
        \item \textbf{FID (Fréchet Inception Distance), IS (Inception Score), CLIPScore:} Measure image generation quality and alignment with text prompts.
    \end{itemize}

    \item \textbf{Metrics for Text-Audio Tasks:}
    \begin{itemize}
        \item \textbf{WER (Word Error Rate), CER (Character Error Rate):} Measure transcription errors in Automatic Speech Recognition (ASR) tasks.
        \item \textbf{MOS (Mean Opinion Score), PESQ (Perceptual Evaluation of Speech Quality):} Evaluate speech quality in Text-to-Speech (TTS) tasks.
        \item \textbf{Recall@K, MRR:} Assess retrieval performance for audio-text alignment.
    \end{itemize}

    \item \textbf{Metrics for Text-Video Tasks:}
    \begin{itemize}
        \item \textbf{BLEU, METEOR, CIDEr, ROUGE:} Used in video captioning tasks to evaluate the quality of generated textual descriptions.
        \item \textbf{Accuracy, BLEU:} Evaluate correctness in video question answering tasks.
        \item \textbf{Recall@K, MRR:} Measure retrieval performance for video-text alignment.
    \end{itemize}

    \item \textbf{Metrics for Multimodal Classification:}
    \begin{itemize}
        \item \textbf{Accuracy, Precision, Recall, F1 score, ROC-AUC:} Used for tasks like sentiment analysis to evaluate model performance.
    \end{itemize}
\end{enumerate}

\subsection{Overview of Future Directions}\label{apdx:Overview of Future Directions}
We summarize the future directions according to the five lifecycle stages of MLLMs, as shown in Figure. \ref{fig:future_directions}.
\begin{figure}[t]
    \centering
\begin{tikzpicture}[node distance=2cm]
\tikzstyle{block} = [rectangle, draw, fill=blue!10, text centered, minimum height=1cm]
\tikzstyle{arrow} = [thick,->,>=stealth]

\node (dataProcessing) [block, minimum width=4cm, minimum height=3cm, align=center] {
    \textbf{Multimodal Data}\\[0.5em] 
    \small 1) Construction of high-quality multimodal data (Sec. \ref{sec:Construction of High-quality Multimodal Data})\\
    \small 2) Exploring the effectiveness of synthetic data (Sec. \ref{sec:Exploring the Effectiveness of synthetic data})\\
    \small 3) Utilizing unaligned and incomplete multimodal data (Sec. \ref{sec:Unaligned and Incomplete Multimodal Data})
};
\node (modelDesign) [block, below of=dataProcessing, yshift=-1.5cm,minimum width=5cm, minimum height=3cm, align=center] {\textbf{Structure Design of MLLMs}\\[0.5em]
 \small 1) More sophisticated multimodal perceiver (Sec. \ref{sec:More Sophisticated Multimodal Perceiver}) \\
\small 2) Perceiver adaptive to LLMs (Sec. \ref{sec:Perceiver Adaptive to LLMs})\\
\small 3) Unifying multimodalities (Sec. \ref{sec:Unifying Multimodalities})
};

\node (modelTrain) [block, below of=modelDesign, yshift=-1cm,minimum width=5cm, minimum height=2cm, align=center] {\textbf{Training of MLLMs}\\[0.5em]
 \small 1) Energy-efficient training approaches (Sec. \ref{sec:Green MLLMs}) \\
};

\node (modelEvaluation) [block, below of=modelTrain, yshift=-1cm,minimum width=6cm, minimum height=3cm, align=center] {\textbf{Evaluation of MLLMs}\\[0.5em]
 \small 1) More comprehensive benchmarks for MLLMs (Sec. \ref{sec:More Comprehensive Benchmarks}) \\
 \small 2) Safety of MLLMs (Sec. \ref{sec:Safety of MLLMs}) \\
};

\node (modelDeployment) [block, below of=modelEvaluation, yshift=-1.5cm,minimum width=6cm, minimum height=3cm, align=center] {\textbf{Deployment of MLLMs}\\[0.5em]
 \small 1) Energy-efficient inference approaches (Sec. \ref{sec:Green MLLMs}) \\
  \small 2) Multimodal agents (Sec. \ref{sec:Multimodal Agents}) \\
  \small 2) Domain-specific MLLMs (Sec. \ref{sec:Domain-specific MLLMs}) \\
};

\draw [arrow] (dataProcessing) -- (modelDesign);
\draw [arrow] (modelDesign) -- (modelTrain);
\draw [arrow] (modelTrain) -- (modelEvaluation);
\draw [arrow] (modelEvaluation) -- (modelDeployment);
\end{tikzpicture}
    \caption{Future directions on five lifecycles of MLLMs.}
    \label{fig:future_directions}
\end{figure}

\clearpage

\begin{IEEEbiography}[{\includegraphics[width=0.8in,height=1in,clip,keepaspectratio]{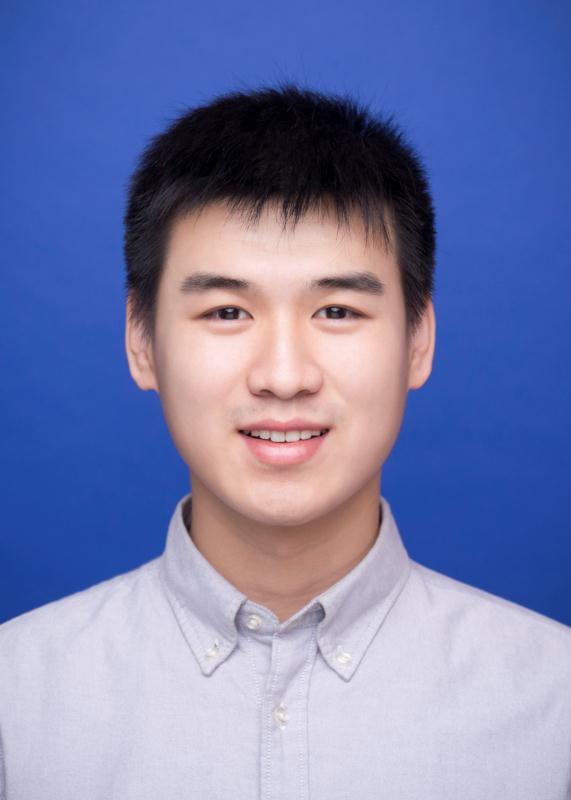}}]{Shezheng Song}
is pursuing the PhD degree in College of Computer, National University of Defense Technology, Changsha, China from 2022. His research interests include Natural language processing and multimodal information processing. He has published several papers in conferences, such as AAAI, TKDE, TNNLS.
\end{IEEEbiography}

\begin{IEEEbiography}[{\includegraphics[width=0.8in,height=1in,clip,keepaspectratio]{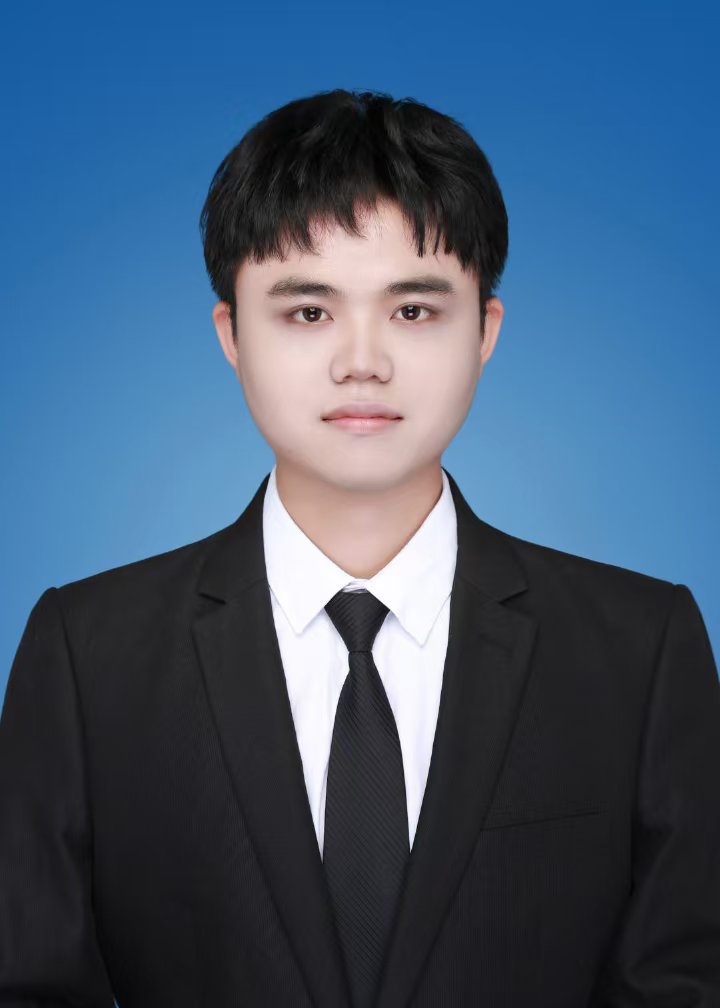}}]{Xiaopeng Li}
is currently pursuing a Ph.D. in the College of Computer Science and Technology at the National University of Defense Technology, Changsha, China, starting in 2023. His research focuses on large language models. He has published several papers in top international conferences, including AAAI and ICSE.
\end{IEEEbiography}

\begin{IEEEbiography}[{\includegraphics[width=0.8in,height=1in,clip,keepaspectratio]{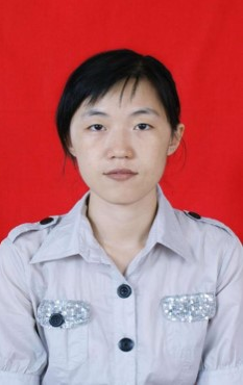}}]{Shasha Li}
    obtained BS and PHD from National University of Defense Technology(NUDT) in 2005 and 2011 supervised by Prof. Huowang Chen and Prof. Zhoujun Li. She was a Post-graduate Internships in MSRA supervised by Prof. Chin-Yew Lin during 2008-2011. Currently, She is a associate professor of School of Computer Science in NUDT. She focus on natural language processing and knowledge graph. She has published more than 40 papers, including ACL, COLING, TKDE, IPM etc.
\end{IEEEbiography}

\begin{IEEEbiography}[{\includegraphics[width=0.8in,height=1in,clip,keepaspectratio]{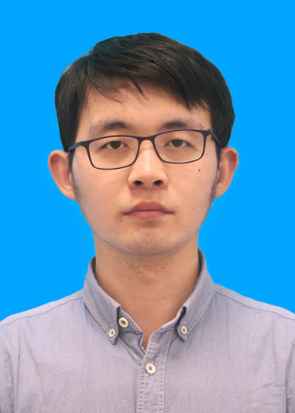}}]{Shan Zhao}
received the PhD degree in College of Computer, National University of Defense Technology, Changsha, China, in 2021.  He is currently an associate professor with the Hefei University of Technology (HFUT), China. His research interests include Natural language processing, Multimodal information extraction. He has published several papers in refereed journals and conferences, such as TNNLS, AAAI, IJCAI.
\end{IEEEbiography}

\begin{IEEEbiography}[{\includegraphics[width=0.8in,height=1in,clip,keepaspectratio]{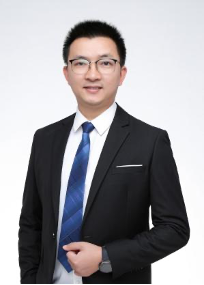}}]{Jie Yu}
    obtained BS and PHD from National University of Defense Technology(NUDT) in 2005 and 2010 supervised by Prof. Huowang Chen and Prof. Zhoujun Li. He was a Post-graduate Internships in National University of Singapore supervised by Prof. Chang Ee-Chien during 2008-2009. Currently, he is a professor and a vice director of the center of system software in NUDT. 
\end{IEEEbiography}

\begin{IEEEbiography}[{\includegraphics[width=0.8in,height=1in,clip,keepaspectratio]{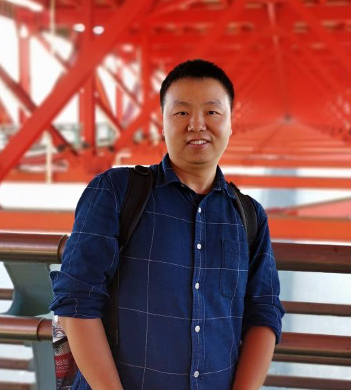}}]{Jun Ma}
obtained his BS and PHD from National University of Defense Technology(NUDT) China in 2005 and 2011 supervised by Prof. Zhiying Wang. He is a associate research fellow in Software Engineering Research Center of Computer School, NUDT. His research focuses on operating system, information security, software engineering. 	
\end{IEEEbiography}

\begin{IEEEbiography}[{\includegraphics[width=0.8in,height=1in,clip,keepaspectratio]{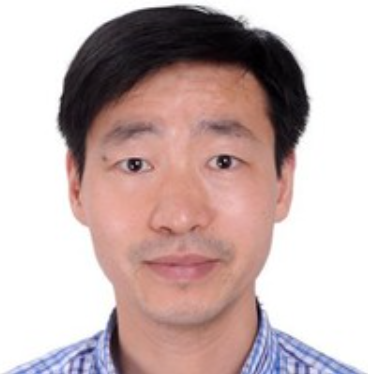}}]{Xiaoguang Mao}
Xiaoguang Mao, a professor at the National University of Defense Technology, PhD supervisor, deputy head of the Department of Computer Science and Technology, academic leader in software engineering, distinguished member of the China Computer Federation (CCF), recipient of the New Century Excellent Talent Award from the Ministry of Education.   	
\end{IEEEbiography}

\begin{IEEEbiography}[{\includegraphics[width=0.8in,height=1in,clip,keepaspectratio]{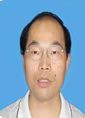}}]{Weimin Zhang}
is currently a Professor in the College of Meteorology and Oceanography at the National University of Defense Technology. His current research interests include numerical weather prediction, high-performance computing, big data, and their interdisciplinary applications in ocean science and meteorology areas.
\end{IEEEbiography}

\begin{IEEEbiography}[{\includegraphics[width=0.8in,height=1in,clip,keepaspectratio]{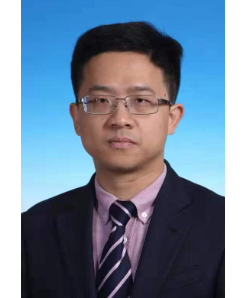}}]{Meng Wang}
is Professor/Doctoral Supervisor at Hefei University of Technology. His main research direction is multimedia content analysis, retrieval, recommendation, and large-scale computing. He has published over 100 papers in his research field, including more than 20 papers on ACM/IEEE Transactions and over 30 papers on top international conferences such as ACM Multimedia, SIGIR, and WWW.
\end{IEEEbiography}

\end{document}